\documentclass{article} 
\usepackage{iclr2026_conference,times}

\usepackage{amsmath,amsfonts,bm}

\def\eqref#1{equation~\ref{#1}}

\def\1{\bm{1}}

\def\ba{{\mathbf{a}}}

\DeclareMathAlphabet{\mathsfit}{\encodingdefault}{\sfdefault}{m}{sl}
\SetMathAlphabet{\mathsfit}{bold}{\encodingdefault}{\sfdefault}{bx}{n}

\newcommand{\pmv}[2]{#1$_{\scriptsize\pm\,#2}$}

\definecolor{softpink}{HTML}{FCE4EC} 
\definecolor{softgreen}{HTML}{86EFAC} 

\definecolor{checkgreen}{RGB}{34,139,34}
\definecolor{crossred}{RGB}{200,45,45}

\newcommand{\cmark}{\textcolor{checkgreen}{\ding{51}}}
\newcommand{\xmark}{\textcolor{crossred}{\ding{55}}}

\makeatletter
\newcommand\footnoteref[1]{\protected@xdef\@thefnmark{\ref{#1}}\@footnotemark}
\makeatother

\newcommand{\cbit}{\begin{compactitem}}
\newcommand{\ceit}{\end{compactitem}}
\newcommand{\cben}{\begin{compactenum}}
\newcommand{\ceen}{\end{compactenum}}

\newcommand{\beq}{\begin{equation}}
	\newcommand{\eeq}{\end{equation}}

\definecolor{darkgreen}{RGB}{41,166,41}

\newcommand{\bit}{\begin{itemize}}
	\newcommand{\eit}{\end{itemize}}
\newcommand{\ben}{\begin{enumerate}}
	\newcommand{\een}{\end{enumerate}}

\newcounter{x}

\newcommand{\bx}{\mathbf{x}}

\newcommand{\br}{\mathbf{r}}

\definecolor{celadon}{rgb}{0.67, 0.88, 0.69}
\definecolor{carolinablue}{rgb}{0.6, 0.73, 0.89}

\newcommand{\bmu}{\boldsymbol{\mu}}
\newcommand{\bomega}{\boldsymbol{\omega}}

\newcommand{\bSigma}{\mathbf{\Sigma}}

\newcommand{\Din}{\mathcal{D}_{\rm in}}
\newcommand{\Dgen}{\mathcal{D}_{\rm gen}}

\definecolor{aliceblue}{rgb}{0.867, 0.917, 0.964}
\definecolor{aliceyellow}{rgb}{0.999, 0.945, 0.796}
\definecolor{alicegray}{rgb}{0.844, 0.867, 0.898}

\usepackage{hyperref}
\usepackage{url}

\usepackage{tcolorbox}
\usepackage{graphicx}
\usepackage{booktabs}
\usepackage{amssymb}
\usepackage{pifont}
\usepackage{multirow}
\usepackage{wrapfig}
\usepackage{threeparttable}
\usepackage{subcaption}
\usepackage{tabularx}
\usepackage[table]{xcolor}
\usepackage{bm}

\setcitestyle{numbers}
\title{From Data to Program: Fast \& Direct Generative Program Inference from Empirical Data}

\author{Simon Klüttermann \\
Carnegie Mellon University\\
\And
Xueying Ding \\
Carnegie Mellon University\\
\And
Leman Akoglu \\
Carnegie Mellon University\\
}

\usepackage{amssymb,amsmath,amsthm,enumitem}
\usepackage{amsfonts} 
\usepackage{xspace}
\usepackage{bbm}
\usepackage[normalem]{ulem}
\usepackage{nicefrac}
\newcommand{\method}{{\sf PRODiGI}\xspace}
\newcommand{\methodft}{{\sf PRODiGI-FT}\xspace}

\iclrarxivcopy

\begin{document}

\maketitle
 
\begin{abstract}
Estimating probability densities from a finite set of samples typically requires dataset-specific model fitting. We introduce \method\footnote{We open-source all artifacts at  \url{https://github.com/psorus/PRODiGI}}, a pretrained data-to-program model that infers an explicit, executable generative program in a single forward pass. 
Pretrained on synthetic datasets paired with their ground-truth programs, \method accommodates diverse generative families and data dimensionalities through template prediction and non-autoregressive program parameter decoding. Its inferred programs support direct sampling, density and score evaluation, and inspection independently of the pretrained model. 
We further introduce program-space fine-tuning, which refines differentiable program parameters by maximizing the likelihood of empirical data while keeping model parameters intact. 
 Experiments show that 
 \methodft achieves better density and score metrics than existing pretrained models, {while offering multi-fold speedups over its closest competitors}. Program-space fine-tuning further reduces generation
MMD by 90\%. By turning empirical data into explicit, reusable programs, \method introduces a new direction for fast, interpretable tabular generative modeling.
\end{abstract}

\vspace{-0.15in}
\section{Introduction}
\vspace{-0.05in}
\label{sec:intro}

Probability density estimation from finite samples is a fundamental problem in statistical inference that underpins generative modeling. Classical approaches range from parametric estimation, which assumes a distributional family (e.g., Gaussian mixtures), to nonparametric methods such as kernel density estimation (KDE), which make fewer assumptions but may deteriorate in high dimensions. 

Several classical generative families are highly expressive: Gaussian Mixture Models (GMMs) are universal density approximators \cite{carreira2002mode}; Structural Causal Models (SCMs) capture a broad class of data-generating processes \cite{bongers2021foundations}; and by Sklar's theorem, Copulas can represent any multivariate distribution through its marginals and dependence structure \cite{sklar1959fonctions}. They offer interpretable, often analytically tractable representations, however at a computational cost: Fitting such models from data may require iterative optimization, combinatorial search, or model selection; e.g., EM \cite{dempster1977maximum} and component selection for GMMs, causal graph search for SCMs, and dependence estimation for Copulas.

In this work, we take a radically different approach to the estimation of density functions. Capitalizing on the prowess of modern tabular foundation models (TFMs) \cite{kurashkin2026current,hollmann2023tabpfn,hollmann2025accurate,shen2025fomod,zhang2025mitra,qu2026tabiclv2,ding2026zero}, we pretrain a single encoder-decoder architecture that maps a given dataset (i.e., $\mathcal{D} = \{\bx_i\}_{i=1}^n$ with i.i.d. samples from a continuous target distribution) directly to its generative program. 

\begin{figure}
 \vspace{-0.15in}
    \centering
    \includegraphics[width=1.0\linewidth]{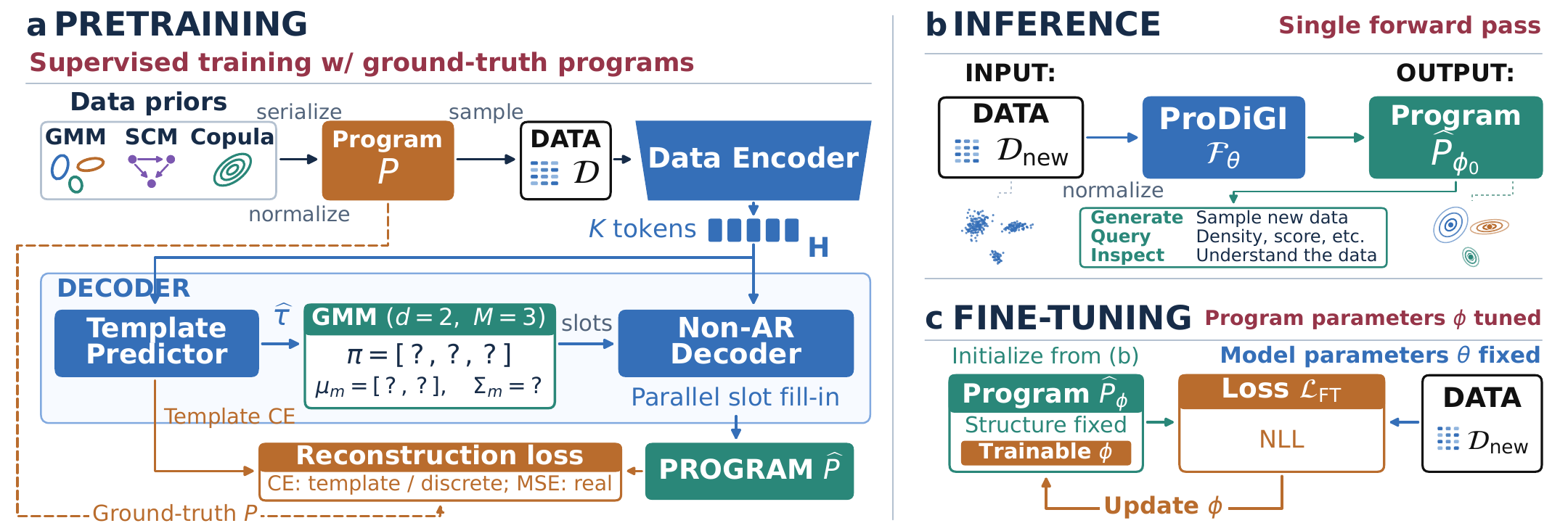}
    \vspace{-0.2in}
    \caption{\method overview:
\textbf{(a)} \textbf{Pretraining}: Data sampled from normalized, serialized programs across diverse priors is encoded 
for template prediction and non-autoregressive slot filling; 
\textbf{(b)} \textbf{Inference}: A single forward pass maps input data to a standalone generative program;  
\textbf{(c)} \textbf{Fine-tuning}: Inferred program's differentiable parameters are refined, model parameters remain fixed.}
    \label{fig:architecture}
    \vspace{-0.2in}
\end{figure}

Our key inspiration is that TFMs are pretrained on synthetic datasets sampled from diverse data priors, where each dataset originates from a \textit{known} generative process---what we call a \emph{program}. Existing TFMs discard this program after sampling the dataset and train on downstream tasks such as classification, regression, or outlier detection. In contrast, we train the model to recover the associated serialized program given a sampled dataset.
Our proposed model, \method (\textsf{PRO}gram \textsf{Di}scovery for \textsf{G}enerative \textsf{I}nference), thus amortizes program inference into pretraining: once pretrained, it infers a generative program from empirical data in a single forward pass (See Figure~\ref{fig:architecture}).

 \method introduces key technical innovations, including serialized canonical program templates for GMM, SCM and Copula based data priors,  a novel non-autoregressive program decoder, and a program-parameter (rather than model-parameter) based fine-tuning, besides being (to the best of our knowledge) the \textit{first} pretrained model that maps (empirical) data directly to a general (generative) program \textit{without} explicit model fitting, search, optimization or model selection. 

Classical parametric density estimators fit prespecified distribution families, while neural density estimators (NDEs) use neural networks to represent more flexible densities \cite{germain2015made,papamakarios2017masked,durkan2019neural,papamakarios2021normalizing}.
Despite this difference, both classes of estimators are \textit{dataset-specific}: their parameters must be fitted separately to each dataset, typically by optimizing a likelihood-based objective.

Closest to our work  is DiScoFormer \cite{ilin2026discoformer}, 
a pretrained  Transformer that outputs
point-wise density and score vector estimates given input samples, but is \textit{not} natively {generative}: it requires an external iterative sampler that repeatedly queries its score estimates.
In addition, TabPFN's \cite{hollmann2025accurate}   pretrained classification and regression models can be  repurposed for both density estimation and generation through an autoregressive factorization of the joint density into univariate conditionals which, however,  requires repeated model evaluations across features.
In contrast to these TFMs, \method is a plug-in density \textit{function} estimator; it outputs an \textit{explicit generative program} in a single forward pass, which supports density evaluation and direct sampling. (See Table~\ref{tab:method-comparison}.)


Together, these distinctions position \method as a new paradigm for tabular generative modeling:\\
As a zero-shot framework that rapidly infers a reusable program that can be directly sampled, queried, and inspected, it bridges modern density estimation with interpretable generative programming.
Our main contributions are summarized as follows.

\begin{itemize}
[leftmargin=10pt, itemsep=-0.01in, topsep=-0.025in]
    \item \textbf{TFM for Generative Program Inference -- without the Fit:} We introduce \method, the \textbf{first} data-to-program pretrained model. 
    Trained to {map empirical samples} from diverse 
    data priors {to underlying generative programs},
    \method requires no fitting or optimization at inference time: it infers a generative program from new empirical data in a single forward pass.

    \item \textbf{Program Serialization \& Normalization:}
    Programs have structure; e.g. GMMs have components,  each with mean and covariance of size related to dimensionality. We represent programs with {canonically serialized templates}, accommodating ($i$) {heterogeneous priors} 
 and ($ii$) {varying dimensions}. We {carefully normalize programs} to remove sensitivity to feature location and scale.

    \item \textbf{Novel Decoder Architecture:}  \method consists of a data encoder and a {Non-AutoRegressive (N-AR) program decoder}.
    We depart from next-token-at-a-time decoding for ($i$)
   program tokens are not always sequentially dependent (e.g., mean vector entries are independent), and ($ii$) AR is prone to  cascading errors yielding invalid programs.
   Instead, \method employs (1) a {template prediction head}, and (2) {a N-AR reconstruction head} to fill-in the template slots in parallel. 

\item \textbf{Fine-tuning the Program -- not the Model:~} 
Empirical data can be used for fine-tuning, however, as inference-time data lacks the ground-truth program,  instead of traditional fine-tuning of the {model's} own parameters, \method fine-tunes the {inferred program's} differentiable parameters,  to align its generated samples with the input data using a likelihood loss.

\item \textbf{Fast, Effective Generation and Estimation:}
\method offers fast generative modeling {at an order-of-magnitude} speedup over existing approaches. Beyond direct sampling, its inferred programs support inspection and statistical queries, including analytical evaluation for GMM-, SCM- and Copula-based programs. Experiments demonstrate effective  distribution matching in two-sample tests, as well as accurate point-wise and population-wide statistical estimation.

\end{itemize}


    

    

\section{Related Work}
\vspace{-0.1in}
\label{sec:prelim}

\begin{table*}[t]
\vspace{-0.1in}
\centering
\small
\setlength{\tabcolsep}{3.25pt}
\renewcommand{\arraystretch}{1.15}
\caption{
Comparing classical, neural, and pretrained
approaches to density and generative modeling.
NDEs: class of Neural Density Estimators;
DDE: data-driven Deep Density Estimation \cite{puchert2021data};
SNO: Score Neural Operator \cite{liao2024score};
NS: Neural Statistician \cite{edwards2016neural};
NNKDE: pretrained adaptive KDE \cite{zhang2026adaptive}.
Last three rows describe capabilities after estimator construction; where direct implies no additional estimation procedure, such as an external sampler,  numerical marginalization, etc.
}
\vspace{-0.1in}
\label{tab:method-comparison}


\begin{tabularx}{\textwidth}{
    @{}>{\raggedright\arraybackslash}X
    cc|c|*{7}{c}|c@{}
}
\toprule
\textbf{Property}
& \textbf{KDE}
& \textbf{GMM}
& \textbf{NDEs}
& \textbf{TPFN}
& \textbf{DiSco}
& \textbf{DDE}
& \textbf{SNO}
& \textbf{NS}
& \textbf{NNKDE}
& \textbf{ST}
& \textbf{Ours} \\
\midrule

Zero-shot/No per-dataset fit
& \xmark & \xmark & \xmark
& \cmark & \cmark & \cmark & \cmark & \cmark & \cmark & \cmark
& \cmark \\

Explicit structure and param.s
& \cmark & \cmark & \xmark
& \xmark & \xmark & \xmark & \xmark & \xmark & \cmark & \cmark
& \cmark \\

Input-free upon construction
& \xmark & \cmark & \cmark
& \xmark & \xmark & \xmark & \cmark & \cmark & \xmark & \cmark
& \cmark \\

\mbox{Pretrn.-model-free upon constr.}
& \cmark & \cmark & \cmark
& \xmark & \xmark & \xmark & \xmark & \xmark & \cmark & \cmark
& \cmark \\

Multiple output families
& \xmark & \xmark & \xmark
& \xmark & \xmark & \xmark & \xmark & \xmark & \xmark & \xmark
& \cmark \\

Variable dimensionality
& \cmark & \cmark & $\triangle$
& \cmark & \xmark & \xmark & \xmark & \xmark & \xmark & \xmark
& \cmark \\

\midrule

Direct generation
& \cmark & \cmark & $\triangle$
& \cmark\textsuperscript{$^\ast$} & \xmark & \xmark
& \xmark & \cmark & \cmark & \cmark
& \cmark \\

Direct density estimation
& \cmark & \cmark & $\triangle$
& \cmark\textsuperscript{$\ast$} & \cmark & \cmark
& \xmark & \xmark & \cmark & \cmark
& \cmark \\

Direct score estimation
& \cmark & \cmark & $\triangle$
& \xmark & \cmark & \xmark
& \cmark & \xmark & \cmark & \cmark
& \cmark \\

\bottomrule
\end{tabularx}

\vspace{3pt}
\begin{minipage}{\textwidth}
\footnotesize
Explicit structure refers to kernel, mixture, SCM, or Copula
representations, rather than neural weights or latent contexts.
Discarding input samples permits retaining a distribution, embedding
or latent context.
Multiple output families refers to the inferred representation,
not the diversity of pretraining distributions.
Variable dimensionality denotes support for different input
feature counts without retraining or switching pretrained models.
$\triangle$: architecture-dependent; 
\textsuperscript{$\ast$}TabPFN uses autoregressive conditional evaluations,
requiring repeated pretrained-model calls.
\end{minipage}
\vspace{-0.15in}
\end{table*}

Density estimation and generative modeling have a broad literature, spanning classical statistical estimators and modern neural approaches. 
Here we review pretrained statistical models and their comparison to classical and neural estimators, while Appdx. \ref{app:related}
gives detailed related work.



 \textbf{Pretrained Models for Generation and Estimation.~}
Pretrained models differ substantially in their formulations of what they infer and which tasks they support directly. 
Set Transformer \cite{SetTransformer} is a predecessor to our work, directly predicting GMM parameters from an input dataset and subsequently refining them using EM. While this demonstrates amortized parameter inference for a fixed GMM configuration, \method extends this approach to heterogeneous generative families, varying dimensionalities, and variable program structures through template prediction and non-autoregressive parameter decoding.
DiScoFormer \cite{ilin2026discoformer} is pretrained to predict point-wise densities and score vectors through cross-attention to an input dataset, while generation requires mak{ing} iterative model calls for score estimates. TabPFN \cite{hollmann2025accurate} instead repurposes its pretrained classification and regression models through an autoregressive factorization, $\hat{p}(\bx|\mathcal{D})$$=$$\prod_{j=1}^{d}\hat{p}(x_j| x_{<j},\mathcal{D})$. Density estimation combines the conditional predictions, while generation samples features sequentially, with repeated model  calls.

Other pretrained approaches include DDE \cite{puchert2021data} and {NN}KDE \cite{zhang2026adaptive}, which train separate models for each dimensionality. DDE predicts pointwise densities, whereas {NN}KDE predicts bandwidth matrices for a sample-centered kernel mixture supporting density evaluation, score evaluation, and direct sampling. Score Neural Operator \cite{liao2024score} maps distribution embeddings to time-dependent score functions and generates through iterative sampling. In contrast, \method maps an input dataset to an explicit generative program directly; subsequent sampling and statistical queries execute this program without further calls to the pretrained network. 

\textbf{Comparing Classical, Neural, and Pretrained Statistical Approaches.~}
Table~\ref{tab:method-comparison} highlights complementary properties of classical, neural, and pretrained approaches. GMMs and conventional NDEs yield input-free estimators but require dataset-specific fitting. KDE and {NN}KDE provide explicit statistical representations while retaining observations as kernel centers. Among pretrained methods, TabPFN, DiScoFormer, and DDE require the sample context; Neural Statistician and Score Neural Operator can retain summaries instead, but still rely on shared pretrained networks. \method combines inference without per-dataset optimization, explicit statistical structure, and independence from both the original samples and the pretrained network after construction. It additionally predicts programs across multiple prior families, namely GMMs, SCMs, and Copulas.

These properties support fast generation, density estimation, and score-vector estimation: a single forward pass constructs a reusable program that natively supports all three tasks without further pretrained-model evaluations. The distinction concerns both construction and subsequent execution. Fitted GMMs and KDE already support direct sampling and density/score evaluation; \method combines these capabilities with pretrained inference and an input-free representation. TabPFN requires repeated conditional-model evaluations, while DiScoFormer directly predicts densities and scores but requires an external sampler for generation.  Thus, \method's computational advantage lies in avoiding per-dataset optimization and repeated pretrained-network calls, with its construction cost amortized over subsequent samples and statistical queries.

\vspace{-0.225in}
\section{\method for Zero-shot Data to Program}
\vspace{-0.05in}
\label{sec:method}

\textbf{Overview.}
\method maps an empirical tabular dataset directly to an executable generative
program. Figure~\ref{fig:architecture} outlines pretraining, inference, and fine-tuning stages. Given a dataset
$\mathcal{D}=\{\bx_i\}_{i=1}^{n}$, $\bx_i\in\mathbb{R}^{d}$,
\method first encodes $\mathcal{D}$ into a fixed-size representation, predicts
the structural template of the underlying program, and then reconstructs its
parameters using a non-autoregressive decoder:

\vspace{-0.175in}
\begin{equation}
    \widehat{P}
    =
    \mathcal{F}_{\theta}(\mathcal{D})
    =
    D_{\theta_D}\!\left(
        \widehat{\tau},
        E_{\theta_E}(\mathcal{D})
    \right).
    \label{eq:data2program}
\end{equation}
\vspace{-0.2in}

Here, $\widehat{\tau}$ specifies the program family and structural attributes,
while $\widehat{P}$ is the resulting concrete generative program. Once inferred,
$\widehat{P}$ can be executed independently of $\mathcal{D}$.

\vspace{-0.05in}
\subsection{Data Encoder}
\label{sec:data_encoder}
\vspace{-0.05in}

As datasets are unordered collections of samples, we use a Set Transformer~\cite{SetTransformer} for data encoder: 

\vspace{-0.175in}
\begin{equation}
    \mathbf{H}
    =
    E_{\theta_E}(\mathcal{D})
    \in \mathbb{R}^{K\times h},
    \label{eq:encoder}
\end{equation}
where $K$ is the number of representation tokens with dimension $h$. 
No positional encoding is applied to the input samples, so that the encoder
remains permutation-invariant with respect to their ordering.
The resulting $K$ tokens  provide a fixed-size
representation of a dataset with arbitrary sample size. 

\vspace{-0.05in}
\subsection{Program Template Prediction} 
\vspace{-0.05in}
\label{sec:templates} 

A \emph{program} specifies a complete generative process, including its family, structure, and distributional parameters. Because different program families have different structures and numbers of parameters, we represent program structure separately from its concrete values. 

We formalize a template as 
$\tau=(\text{fam},\ba) 
$, 
where $\text{fam}\in\{\mathrm{GMM},\mathrm{SCM},\mathrm{Copula}\}$ denotes the data prior family and $\ba$ represents the discrete attributes that specify its structure. The corresponding concrete program is obtained by instantiating the template with its parameter values. 

\vspace{-0.025in}
\textbf{Heterogeneous Programs.~}
Unlike previous work \cite{hollmann2025accurate, ilin2026discoformer}, \method is trained with programs from three distinct priors, and tackles the challenge of unifying them through structured serialized programs. 
GMMs contain varying number of components, each with mean and diagonal covariance, in varying dimensions. SCMs are specified by a directed acyclic graph (DAG) among variables with structural equations specifying causal relationships. We materialize the DAG by 
generating a random acyclic graph,  
 sample the inputs, edge weights, and noise variables from the  standard Normal { and Uniform} distributions, and set the structural equations 
 as the weighted sum followed by a random activation function. 
Finally, Copulas consist of marginals from a pool of parametric probability functions such as Uniform, Exponential, Student-t,  etc. alongside a Gaussian  copula.
Appdx. \ref{app:priors} gives detailed descriptions and configurations for the data priors.

\vspace{-0.025in}
\textbf{Normalization.~}
To remove sensitivity to feature-wise shifts and scaling, we train \method in standardized coordinates. Crucially, we \textit{normalize the generative programs themselves}, ensuring consistency between training data and program targets. To this end, we reparameterize each sampled program to target zero mean and unit variance per feature, and then draw training data from it. This uses analytical moments for GMMs and Copula marginals and sample estimates for SCMs. Appdx.~\ref{app:normalization} details the transformations to normalize programs from each prior family.

During inference, we standardize each input feature using its empirical mean and standard deviation, infer a normalized program with a single forward pass, and use the same statistics to transform the program back to the original feature coordinates.

\vspace{-0.025in}
\textbf{Serialization, Templates, and Prediction.~}
For the priors considered here, a GMM template is identified by the dimensionality and the number of mixture components; an SCM template is identified by the dimensionality and maximum number of parents per node; and a Copula template is identified by the dimensionality alone. For example, in a configuration of $50$ dimensions using up to {10} components in GMMs and up to 5 parents per variable in SCMs, these choices produce $|\mathcal{T}|=786$ templates in total: $500$ GMM, $236$ SCM, and $50$ Copula templates. 

To ensure a unique structural representation, we canonically serialize each program by imposing a deterministic ordering of its entities and parameters. As a result, equivalent programs differing only in arbitrary ordering (e.g., GMM components or SCM parent variable indices) map to the same template and slot layout.
Appdx.~\ref{app:programs} provides details on program templates, canonical serialization, and tokenization, along with concrete examples.

Prior to decoding, a 
template prediction head operates on the encoded dataset representation to infer which template to instantiate, denoted as

\vspace{-0.15in}
\begin{equation} \widehat{\tau} = T_{\theta_T}(\mathbf{H})\;, \quad \text{ where } \tau=(\text{fam},\ba) \;, \text{ and } \;\text{fam}\in\{\mathrm{GMM},\mathrm{SCM},\mathrm{Copula}\}\;.  \label{eq:template-pred} \end{equation}
\vspace{-0.15in}

\vspace{-0.1in}
\subsection{Non-Autoregressive (N-AR) Program Decoder} 
\vspace{-0.05in}
\label{sec:program_decoder} Given a template $\tau$, its serialized representation defines an ordered sequence of typed fill-in slots, \begin{equation} S_{\tau}=(s_1,\ldots,s_{L_\tau}), \end{equation} where each slot specifies the semantic role and type of a value to be reconstructed. For instance, slots may represent mixture weights and mean entries in GMMs, parent indices and coefficients in SCMs, or marginal families and dependence parameters in Copulas. Each slot is represented by a learned embedding that encodes its semantic identity and canonical position. The slot representations are processed jointly by the decoder $D_{\theta_D}(\cdot,\cdot)$ through cross-attending the dataset embedding: \begin{equation} \mathbf{R} = D_{\theta_D}(S_\tau ; \mathbf{H}) \in\mathbb{R}^{L_\tau\times h} \;.
\label{eq:decoder} \end{equation}

\vspace{-0.05in}
\textbf{Why N-AR Decoding.~} \method radically abandons autoregressive decoding; its Transformer-based decoder removes the causal mask and predicts all template slots
\textit{in parallel}. This is appropriate because unlike natural-language sequences, program parameters do
not generally have a left-to-right dependency: for example, entries of a GMM
mean vector or the marginal choices in a Copula are independently sampled.
Nevertheless, self-attention allows the decoder to capture dependencies among
slots when needed, such as the structural relationships within an SCM. 

In addition, our N-AR decoder  avoids the error accumulation of autoregressive decoding, where an early
mistake can propagate to subsequent tokens and ultimately produce \textit{invalid} programs.
Because \method's decoding is performed against a structurally valid, syntactically typed
template, the output remains consistent with the selected program structure.

Specifically, for each slot $s$, a type-specific prediction head maps its decoder
representation $\br_s$ to the corresponding program value:
\begin{equation}
\widehat{z}_s = \phi_t(W_t\br_s)\;,
\qquad  \text{ where } s\in\mathcal{S}_t, \text{ and } \;
t\in\{\mathrm{cat},\mathrm{int},\mathrm{cont}\}\;,
\label{eq:slot-heads}
\end{equation}
where $\mathcal{S}_{\mathrm{cat}}$, $\mathcal{S}_{\mathrm{int}}$, and
$\mathcal{S}_{\mathrm{cont}}$ denote categorical, integer, and continuous typed 
slots, respectively, $\phi_{\mathrm{cat}}=\phi_{\mathrm{int}}=\operatorname{softmax}$
and $\phi_{\mathrm{cont}}$ is the identity. This parallel template-based decoding not only eliminates
autoregressive error propagation but also substantially reduces program decoding
latency.



 \vspace{-0.05in}
 \subsection{Training} 
 \vspace{-0.05in}
 \label{sec:training} We train all three components of \method, data encoder, template predictor and program decoder, end-to-end to recover the ground-truth program $P$ from empirical data
$\mathcal{D}$ drawn from $P$. For each training task, we first sample a
generative program uniformly at random from a synthetic prior and then draw
$i.i.d.$ samples from that program:
\begin{equation}
    P \sim \operatorname{Unif}(\mathcal{P})\;,
    \qquad
    \mathcal{D}=\{\bx_i\}_{i=1}^{n}\;,
    \quad
    \bx_i \overset{\mathrm{i.i.d.}}{\sim} P\;\;.
    \label{eq:training-generation}
\end{equation}

 \vspace{-0.15in}
 \paragraph{Oracle-template training.} The template predictor and parameter decoder are coupled through the template: different templates can imply different types and number of parameters and thus number of slots to decode. Consequently, supervising the decoder through a predicted template would be ill-defined whenever the template prediction is incorrect. Therefore,  we  provide the ground-truth template $\tau^\star=\tau(P)$ to the decoder during training: \begin{equation} \mathbf{R} = D_{\theta_D} \left( S_{\tau^\star}, E_{\theta_E}(\mathcal{D}) \right). \label{eq:oracle-template} \end{equation} The template prediction head is trained simultaneously to recover $\tau^\star$. During inference, the ground-truth template is not available, for which the predicted $\widehat{\tau}$ is used. 

 \vspace{-0.1in}
 \paragraph{Training objective.} Letting $\mathcal{S}_{\mathrm{cat}}$, $\mathcal{S}_{\mathrm{int}}$, and $\mathcal{S}_{\mathrm{real}}$ denote categorical, integer, and continuous valued slots in a template, respectively, we  
 optimize
\begin{align}
\mathcal{L}
&=
\mathcal{L}_{\mathrm{temp}}
+\mathcal{L}_{\mathrm{int}}
+\mathcal{L}_{\mathrm{cat}}
+\mathcal{L}_{\mathrm{cont}} \;.
\label{eq:loss}
\end{align}
The first term is the cross-entropy loss for the template classification head, which directly predicts the index of the ground-truth template among the $|\mathcal{T}|$ plausible templates. The remaining terms supervise the parameters of the ground-truth template, using cross-entropy for categorical and  integer slots and mean-squared error for continuous-valued slots.

\begin{figure}
\vspace{-.1in}
    \centering
    \includegraphics[width=\linewidth]{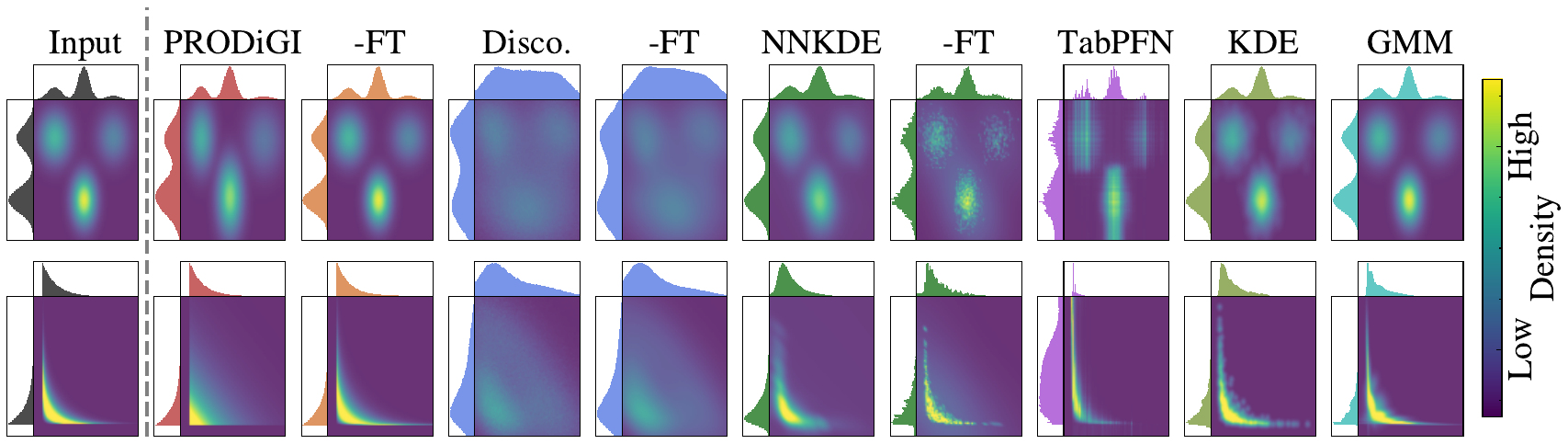}
   \vspace{-0.2in}
    \caption{Example input data vs. baseline generations, with corresponding marginals. {Appdx. \ref{sec:other} presents further examples and case studies.} 
    }
    \label{fig:gen_all_priors}
    \vspace{-0.15in}
\end{figure}

\subsection{Fine-tuning} 
\vspace{-0.05in}
\label{sec:finetune}

Conventional supervised fine-tuning adapts pretrained tabular foundation models by updating their weights on labeled downstream data. For unsupervised tasks like density estimation and data generation, however, only empirical samples are available, without labels or ground-truth programs. We introduce \textbf{program-space fine-tuning}, which \textit{adapts the parameters of the inferred generative program rather than those of the pretrained model itself}.

Starting from \method's zero-shot prediction, we refine the inferred program's differentiable parameters $\phi$ via gradient descent, while keeping the pretrained model weights $\theta$ and the inferred  program structure fixed. Formally,

\vspace{-0.075in}


\begin{equation}
    \widehat{P}_{\phi_0} := \text{\method}(X_{\text{new}};\theta),
    \quad
    \min_{\phi}\;
        -\sum_{x \in X_{\text{new}}} \log \widehat{P}_{\phi}(x) 
\end{equation}

\vspace{-0.15in}

Starting from the inferred program parameters $\phi_0$, we minimize the analytical negative log-likelihood (NLL) of the empirical data. 
Through this optimization, fine-tuning achieves dataset-specific adaptation using exclusively empirical samples.

Importantly, this fine-tuning paradigm supports refinement across all three program families: component {weights}, means and covariances in GMMs, functional mechanism parameters in SCMs, and parametric marginals and dependence parameters in Copulas.  

Program-space fine-tuning decouples adaptation from the pretrained model. By confining optimization to the substantially smaller parameter space of the inferred generative program, it reduces gradient and optimizer memory and limits the degrees of freedom for dataset-specific tuning. Unlike model-weight fine-tuning, it requires no backpropagation through the pretrained model, while subsequent generation and estimation tasks also operate independently of the model.

\vspace{-0.05in}
\section{Experiments}
\vspace{-0.1in}
\label{sec:exp}




\textbf{Baselines.~}
We compare \method against existing pretrained models \textbf{DiScoFormer} \cite{ilin2026discoformer},  \textbf{TabPFN} \cite{hollmann2025accurate}, and \textbf{NNKDE} \cite{zhang2026adaptive}.
DiScoFormer natively supports zero-shot density and score estimation but not generation for which we employ {Langevin dynamics}  \cite{RobertsTweedie1996}.   TabPFN is natively a zero-shot regressor/classifier; density estimation and sampling are executed auto-regressively, while score vectors are obtained by finite differentiation. 
NNKDE employs KDE using its zero-shot estimated  sample-specific bandwidth matrices.
We also fine-tune \method, DiscoFormer, and NNKDE as prescribed, variants denoted as \textbf{-FT}.
 Details on task execution  mechanisms for pretrained baselines are given in Appdx. \ref{app:baseline_details}.
In addition, we compare these zero-shot models to the classical, also non-training baseline \textbf{KDE} \cite{parzen1962estimation} {and to the fit-per-dataset \textbf{GMM} \cite{dempster1977maximum}.

For GMM, we select the number of components (up to 10) using BIC; for KDE, we tune the RBF kernel bandwidth by maximizing held-out data likelihood. Both DiScoFormer and NNKDE are trained on one \textit{fixed} input dimensionality at a time, therefore, we use a separate model  for each $d\in\{1,2,5,10,20,30,40,50\}$ for these baselines and restrict our evaluation to these dimensions.


\textbf{Tasks and Metrics.~}  
We evaluate \method against baselines on three main tasks: \textbf{(1) Generation} quality via distribution match, \textbf{(2) Density estimation}, and \textbf{(3) Score (vector) estimation}. 

For  \textit{generation},  
 we use \textbf{(1.i)} the MMD statistic \cite{gretton2012kernel} {with  bandwidth $\sigma$$=$$\sqrt{d}$ scaling with the dimension} as a \textit{distribution match} metric, measuring the discrepancy between i.i.d. original samples from $\Din {\sim} P$ and generated samples $\Dgen 
 {\sim} \widehat{P}$. We also  report the \textbf{(1.ii)} rejection rate of a two-sample test as the  proportion of datasets where the
null hypothesis is rejected.
For \textit{density estimation},  we compute   \textbf{(2.i)} $\mathrm{MAE}_{\mathrm{dense}}
=
\frac{1}{n}
\sum_{i=1}^{n}
|
\log{\widehat{f}(\bx_i)}-\log{f(\bx_i)}
|$, \textbf{(2.ii)} the negative log likelihood (NLL), and also report \textbf{(2.iii)} the Spearman correlation between the ground truth ranking of test samples by density vs. the ranking based on the estimates. 
Finally for \textit{score estimation}, we report
\textbf{(3.i)} $\mathrm{MAE}_{\mathrm{score}}
=
\frac{1}{n}
\sum_{i=1}^{n}
\|
\nabla \log(\widehat{f})(\bx_i)-\nabla \log(f)(\bx_i)
\|_1$, and \textbf{(3.ii)} the cosine similarity to the ground truth vector for directional accuracy.
In addition, we report \textit{running time} for each task  on 1000 examples each, as different baselines employ varying procedures per task. 



\vspace{-0.025in}
\textbf{Datasets.~}
We evaluate on $96$ datasets from GMM, SCM, and Copula priors following our priors, $96$ OOD datasets (Appendix \ref{app:ood})  and $96$ real-world datasets from OddBench \cite{ding2026macrodata}, balanced across dimensions and priors.
We discard class labels and labeled anomalies for unsupervised evaluation, and randomly subsample features exceeding pretraining dimensions.
For real data, we derive pseudo ground truth from Oracle GMMs with up to $100$ components selected by BIC, retaining $37$ datasets with indistinguishable GMM-generated and real samples under a two-sample test.

\vspace{-0.025in}
\textbf{Implementation details.}
We train \method using simulated datasets that contain $n$$\in$$ [1024,8192]$ observations and dimensions $d$$\in$$ [1,50]$, with $1500$ batches per epoch and  $48$ program-reconstruction tasks per batch. Tasks are balanced across the GMM, SCM, and Copula priors and across their corresponding templates.
Encoder $E_{\theta_E}(\cdot)$  produces $K$$=$$20$ tokens in $h$$=$$512$ 
dimensions, consists of four Set Transformer blocks with four attention heads each, and contains $12.6$M parameters.
Template prediction head $T_{\theta_T}(\cdot)$ is a one-layer MLP with {7.9M} parameters. 
Decoder $D_{\theta_D}(\cdot,\cdot)$ consists of three Transformer blocks with four attention heads and contains $14.0$M parameters.
Overall, \method has {$|\theta|$$=${$32.5$M}} trainable parameters.
{We train for $750$ Epochs on $54$M distinct programs, which takes $14$ days on a H100 GPU,  and evaluate each baseline on a l40s-48 GPU.}

\subsection{Main Results}


\begin{figure}[!t]
\vspace{-0.1in}
    \centering
\includegraphics[width=1.0\linewidth]{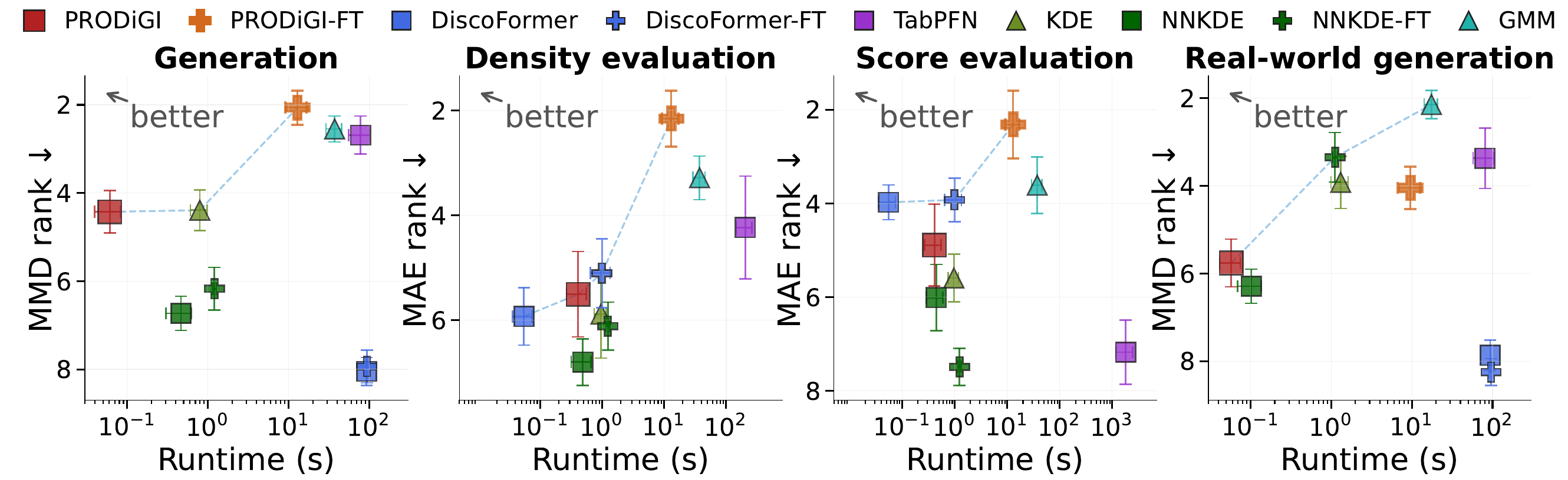}
\vspace{-0.225in}
    \caption{Performance-Inference time trade-off on Generation, Density, and Score estimation tasks on synthetic (all priors) and real-world data. \textbf{\methodft performs best on density and scoring, while \method offers the fastest generation, remaining competitive on real-world datasets}.}
    \label{fig:rank_comparison_all_priors_realgen}
    
\end{figure}


\textbf{Mixed-prior data.~} 
Figure~\ref{fig:gen_all_priors} compares generated samples on two distinct prior families. Fine-tuning substantially improves \method's recovery of clusters and nonlinear structure. DiScoFormer struggles to reproduce the input distributions despite fine-tuning, while NNKDE variants generate overly diffuse samples. TabPFN and KDE closely match the inputs; although KDE simply perturbs observed samples using its tuned bandwidth, while TabPFN incurs substantially higher runtime.  GMM struggles to capture skewed marginals smoothly. 


\begin{table}[!th]
\vspace{-0.1in}
    \centering
    \caption{Performance comparison across tasks and metrics over all prior families$^\text{a}$: (1) \textbf{Generation distribution match} via MMD ($\downarrow$) where $\% p$$<$$0.05$ ($\downarrow$) denotes the proportion of datasets where the null hypothesis ($H_0$: generated and real distributions are identical) is rejected; 
    (2) \textbf{Density estimation} via point-wise MAE ($\downarrow$), population-wide Spearman correlation ($\uparrow$), and NLL of predicted densities ($\downarrow$); 
    (3) \textbf{Score estimation} via MAE ($\downarrow$) and cosine similarity ($\uparrow$). Average rank ($\downarrow$) per metric is across methods over all datasets. Time per task (sec.) is averaged over datasets.}
    \label{tab:all_metrics_priors}
    \vspace{-0.1in}

    \setlength{\tabcolsep}{1.5pt} 
    \renewcommand{\arraystretch}{1.05}

    \resizebox{\columnwidth}{!}{%
    \begin{threeparttable}

\begin{tabular}{c@{\hspace{2pt}}l|@{\hspace{2pt}}*{2}{c}|@{\hspace{2pt}}*{7}{c}}
        \toprule
        & \textbf{Metric}  &\textbf{PRODiGI} & \textbf{-FT} & \textbf{DiscoFormer} & \textbf{-FT} & \textbf{NNKDE} & \textbf{-FT} & \textbf{TabPFN} & \textbf{KDE} & \textbf{GMM} \\
        \midrule
\multirow{5}{*}{\rotatebox[origin=c]{90}{\textbf{Generation}}}
                & MMD ($\downarrow$) {\tiny$\times 10^3$} & \pmv{5.46}{1.45} & \pmv{\underline{0.56}}{0.20} & \pmv{41.37}{8.20} & \pmv{67.58}{16.51} & \pmv{19.51}{2.77} & \pmv{10.98}{1.96} & \pmv{\textbf{0.33}}{0.04} & \pmv{5.24}{0.95} & \pmv{0.78}{0.30} \\

                & Rank ($\downarrow$) & \pmv{4.43}{0.16} & \pmv{\textbf{2.06}}{0.13} & \pmv{8.05}{0.11} & \pmv{7.94}{0.12} & \pmv{6.73}{0.13} & \pmv{6.17}{0.16} & \pmv{2.69}{0.14} & \pmv{4.39}{0.15} & \pmv{\underline{2.55}}{0.10} \\

        \noalign{\smallskip}\cline{2-11}\noalign{\smallskip}
                & \% $p$$<$$0.05$ ($\downarrow$) & 57.3\% & \textbf{14.6\%} & 99.0\% & 97.9\% & 92.7\% & 80.2\% & \underline{29.2\%} & 61.5\% & 31.3\% \\

        \noalign{\smallskip}\cline{2-11}\noalign{\smallskip}
                & Time ($\downarrow$) & \pmv{\textbf{0.06}}{0.01} & \pmv{12.90}{1.29} & \pmv{92.70}{0.14} & \pmv{93.27}{0.14} & \pmv{\underline{0.46}}{0.26} & \pmv{1.20}{0.02} & \pmv{78.09}{7.52} & \pmv{0.79}{0.06} & \pmv{37.24}{2.74} \\

                & Rank ($\downarrow$) & \pmv{\textbf{1.04}}{0.02} & \pmv{4.98}{0.03} & \pmv{7.60}{0.05} & \pmv{8.60}{0.05} & \pmv{\underline{2.15}}{0.06} & \pmv{3.86}{0.05} & \pmv{7.41}{0.14} & \pmv{3.02}{0.06} & \pmv{6.33}{0.05} \\

        \midrule
        \midrule
\multirow{8}{*}{\rotatebox[origin=c]{90}{\textbf{Density}}}
                & MAE ($\downarrow$) & \pmv{6.39}{0.80} & \pmv{\textbf{2.26}}{0.35} & \pmv{4.74}{0.44} & \pmv{3.64}{0.35} & \pmv{6.29}{0.63} & \pmv{6.09}{0.62} & \pmv{\underline{2.27}}{0.12} & \pmv{6.89}{0.81} & \pmv{3.43}{0.44} \\

                & Rank ($\downarrow$) & \pmv{5.51}{0.27} & \pmv{\textbf{2.15}}{0.18} & \pmv{5.93}{0.18} & \pmv{5.10}{0.22} & \pmv{6.80}{0.15} & \pmv{6.11}{0.15} & \pmv{4.23}{0.33} & \pmv{5.89}{0.28} & \pmv{\underline{3.28}}{0.14} \\

        \noalign{\smallskip}\cline{2-11}\noalign{\smallskip}
                & Spear. ($\uparrow$) & \pmv{0.72}{0.02} & \pmv{\textbf{0.90}}{0.01} & \pmv{0.67}{0.02} & \pmv{0.70}{0.02} & \pmv{0.78}{0.01} & \pmv{0.77}{0.01} & \pmv{\underline{0.84}}{0.01} & \pmv{0.77}{0.02} & \pmv{0.82}{0.02} \\

                & Rank ($\downarrow$) & \pmv{5.35}{0.30} & \pmv{\textbf{1.89}}{0.15} & \pmv{7.68}{0.12} & \pmv{6.42}{0.17} & \pmv{5.21}{0.17} & \pmv{5.16}{0.15} & \pmv{\underline{3.62}}{0.25} & \pmv{5.89}{0.27} & \pmv{3.78}{0.19} \\

        \noalign{\smallskip}\cline{2-11}\noalign{\smallskip}
                & NLL ($\downarrow$) & \pmv{4.84}{3.20} & \pmv{\textbf{0.77}}{3.21} & -\tnote{b} & -\tnote{b} & \pmv{4.67}{3.21} & \pmv{4.43}{3.21} & -\tnote{b} & \pmv{5.23}{3.32} & \pmv{\underline{1.68}}{3.20} \\

                & Rank ($\downarrow$) & \pmv{3.84}{0.17} & \pmv{\textbf{1.65}}{0.12} & -\tnote{b} & -\tnote{b} & \pmv{5.02}{0.09} & \pmv{4.15}{0.09} & -\tnote{b} & \pmv{4.19}{0.16} & \pmv{\underline{2.16}}{0.09} \\

        \noalign{\smallskip}\cline{2-11}\noalign{\smallskip}
                & Time ($\downarrow$) & \pmv{\underline{0.41}}{0.08} & \pmv{13.05}{1.27} & \pmv{\textbf{0.05}}{0.02} & \pmv{0.99}{0.38} & \pmv{0.49}{0.24} & \pmv{1.24}{0.03} & \pmv{203.19}{19.18} & \pmv{0.96}{0.07} & \pmv{37.24}{2.74} \\

                & Rank ($\downarrow$) & \pmv{2.99}{0.09} & \pmv{7.00}{0.03} & \pmv{\textbf{1.08}}{0.04} & \pmv{4.36}{0.06} & \pmv{\underline{2.32}}{0.07} & \pmv{5.77}{0.05} & \pmv{8.69}{0.07} & \pmv{4.55}{0.13} & \pmv{8.23}{0.04} \\

        \midrule
        \midrule
\multirow{6}{*}{\rotatebox[origin=c]{90}{\textbf{Score}}}
                & MAE ($\downarrow$) & \pmv{\textbf{94.80}}{33.33} & \pmv{232.00}{118.62} & \pmv{159.63}{78.69} & \pmv{\underline{158.60}}{78.18} & \pmv{162.72}{77.21} & \pmv{219.13}{101.65} & \pmv{196.07}{87.51} & \pmv{165.08}{80.45} & \pmv{159.08}{78.35} \\

                & Rank ($\downarrow$) & \pmv{4.89}{0.29} & \pmv{\textbf{2.32}}{0.24} & \pmv{3.98}{0.12} & \pmv{3.93}{0.16} & \pmv{6.01}{0.24} & \pmv{7.49}{0.13} & \pmv{7.18}{0.23} & \pmv{5.59}{0.17} & \pmv{\underline{3.61}}{0.20} \\

        \noalign{\smallskip}\cline{2-11}\noalign{\smallskip}
                & Cos. ($\uparrow$) & \pmv{0.59}{0.03} & \pmv{\textbf{0.78}}{0.03} & \pmv{0.52}{0.03} & \pmv{0.52}{0.03} & \pmv{0.48}{0.02} & \pmv{0.38}{0.02} & \pmv{0.43}{0.03} & \pmv{0.47}{0.03} & \pmv{\underline{0.59}}{0.03} \\

                & Rank ($\downarrow$) & \pmv{4.99}{0.31} & \pmv{\textbf{2.17}}{0.19} & \pmv{4.99}{0.18} & \pmv{4.79}{0.20} & \pmv{5.85}{0.19} & \pmv{6.84}{0.15} & \pmv{5.81}{0.27} & \pmv{6.05}{0.23} & \pmv{\underline{3.51}}{0.22} \\

        \noalign{\smallskip}\cline{2-11}\noalign{\smallskip}
                & Time ($\downarrow$) & \pmv{\underline{0.41}}{0.08} & \pmv{13.04}{1.27} & \pmv{\textbf{0.05}}{0.02} & \pmv{0.99}{0.38} & \pmv{0.44}{0.21} & \pmv{1.24}{0.03} & \pmv{1806.43}{246.25} & \pmv{0.97}{0.07} & \pmv{37.29}{2.74} \\

                & Rank ($\downarrow$) & \pmv{3.04}{0.08} & \pmv{7.04}{0.04} & \pmv{\textbf{1.08}}{0.04} & \pmv{4.35}{0.06} & \pmv{\underline{2.26}}{0.07} & \pmv{5.78}{0.05} & \pmv{8.64}{0.07} & \pmv{4.55}{0.12} & \pmv{8.25}{0.04} \\

        \bottomrule
    \end{tabular}%

\begin{tablenotes}
\item[a] Detailed results on individual priors and dimensions are given in Appdx. \ref{app:detailed_tables}. 
\item[b] NLL is not reported for DiscoFormer and TabPFN because their local density estimates are not normalized to integrate to one.
\end{tablenotes}
\end{threeparttable}
    }
    \vspace{-0.225in}
\end{table}


Figure \ref{fig:rank_comparison_all_priors_realgen} and Table \ref{tab:all_metrics_priors} show competitive accuracy--runtime tradeoffs for \method and \methodft across generation, density, and score estimation. For generation, \method achieves lower MMD than DiScoFormer and NNKDE in just $0.06$ seconds, approximately $8\times$ faster than the next-fastest method. Fine-tuning reduces MMD by $90\%$ to $0.56$ ($\times 10^{-3}$), outperforming the per-dataset-fit GMM's $0.78$ at one-third of its runtime. TabPFN achieves a slightly lower mean MMD compared to \methodft, but has a higher average rank and is substantially slower than either \method variant. 
Two-sample test rejection rate ($\% p$$<$$0.05$) 
closely mirrors lower MMD scores, with TabPFN (29.2\%), GMM (31.3\%), and especially \methodft (14.6\%)  {with the two-sample test failing to reject equality for a majority of datasets.
DiscoFormer variants rank lowest despite taking over 80 seconds per generation.


In density estimation, \methodft leads most reported metrics and average ranks, where fine-tuning reduces MAE from $6.39$ to $2.26$ and increases Spearman correlation from $0.72$ to $0.90$. 
While TabPFN records a very similar mean MAE ($2.27$ vs. $2.26$), its higher average rank ($4.23$ vs. $2.15$) indicates lower relative performance across datasets.
\methodft is also approximately $16\times$ faster than TabPFN and $3\times$ faster than GMM. DiScoFormer offers the fastest density queries but lower accuracy; moreover, its unnormalized local estimates, like TabPFN's, preclude NLL comparison.


For score estimation, \methodft achieves the best MAE rank ($2.32$) and highest mean cosine similarity ($0.78$) and lowest rank ($2.17$), with a runtime at roughly one-third of the closest competitors.

NNKDE-FT improves generation MMD but worsens both score metrics, while DiScoFormer-FT provides little improvement in score estimation. TabPFN's strong generation performance does not extend to scores; it has the second lowest Cosine Similarities ($0.43$) and requires more than 30 minutes, about $138\times$ longer than \methodft.

In summary, \method offers the fastest generation, while \methodft achieves the lowest rank across every task and quality metric considered here. 

\textbf{Program Inference and Refinement.}
Beyond downstream performance, we investigate whether \method
recovers meaningful program structures and whether its predictions
provide benefits beyond conventional optimization.
Figure \ref{fig:opti_detail} (left) shows accurate reconstruction
of key GMM, Copula, and SCM properties.
In particular, SCM graph reconstruction achieves a directed-edge
F1 of $0.59$, compared with $0.39$ for a random baseline,
despite the ambiguity of recovering programs from observational
data (Appendix \ref{app:progrecon}).

Figure \ref{fig:opti_detail} (right) examines the effect of
fine-tuning budgets. \methodft consistently benefits from
additional optimization, whereas NNKDE-FT largely saturates and
DiScoFormer-FT shows limited improvements.
Balancing accuracy and runtime, we use $300$ optimization steps
for \methodft, $1000$ for NNKDE-FT, and $10$ epochs for
DiScoFormer-FT.

Importantly, these improvements cannot be attributed to
optimization alone. Replacing \method's predicted parameters
with random initializations, while retaining the same template
and optimization budget, increases generation MMD from $0.56$
to $3.07$ ($\times 10^{-3}$) on in-distribution data and from
$37.29$ to $48.11$ on real-world data. Learned initialization
performs better on $70/96$ and $72/96$ datasets, respectively,
demonstrating the benefit of amortized program inference in both settings
(Appendix \ref{app:ftnoprodigi}).

\begin{figure}
\vspace{-.2in}
    \centering

    \includegraphics[width=1\linewidth]{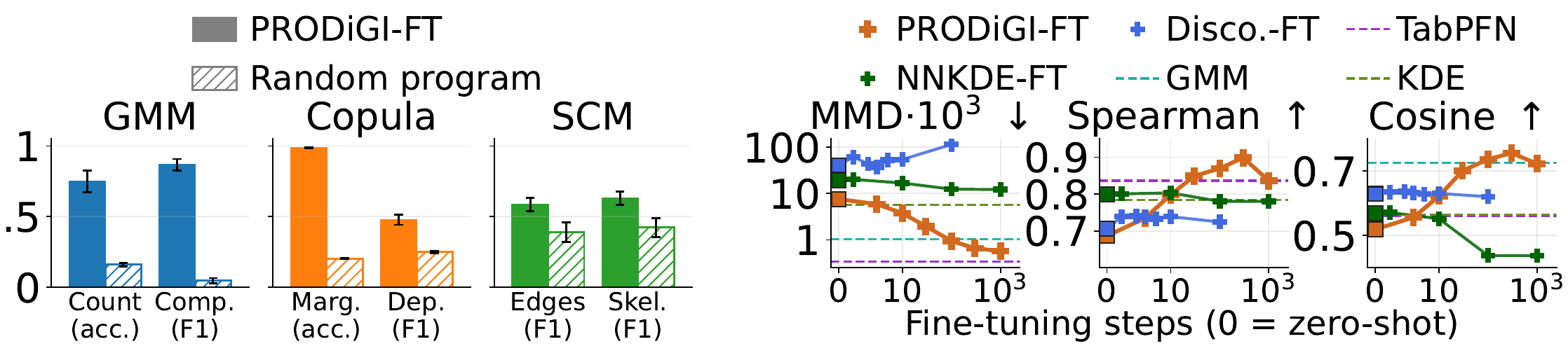}
        
    \caption{Left: parameter reconstruction performance for each family. A component is considered correct for the F1 when its true mean is within one standard deviation, and dependencies measure whether a correlation of $0.2$ or larger is correctly found. Right: Performance of fine-tuning methods with more or less fine-tuning. \methodft shows the best performance and largest improvement. 
    }
    \label{fig:opti_detail}
    \vspace{-0.15in}
\end{figure}

\vspace{-0.025in}
\textbf{Generalization Beyond Training Priors.}
To probe the boundaries of our  data priors, we  extend our evaluation to real-world datasets as a stress test.   
Figure \ref{fig:gen_rw_heatmap} illustrates two examples (additional examples in Fig.   \ref{fig:rw_examples}): \method and \methodft capture the main density structure and the marginals, inferring copula-based programs with Uniform and Kumaraswamy marginals (see Appdx. \ref{app:prodigioutput} for the inferred and fine-tuned programs.)
NNKDE and KDE also recover the broad structure but blur sharp boundaries, while NNKDE-FT yields noisier estimates.
DiScoFormer variants distort the marginals, TabPFN produces spurious density spikes, and GMM introduces artificial modes.



Figure \ref{fig:rank_comparison_all_priors_realgen} (right) shows that 
\method remains the fastest generator at $0.06$ seconds, while NNKDE-FT and KDE offer competitive intermediate tradeoffs between \method's speed and GMM's generation quality. Fine-tuning reduces \method's MMD by approximately $46\%$ and improves recovery of the underlying geometry, achieving lower mean MMD than TabPFN at over $8\times$ faster runtime. Despite fine-tuning gains, real-world distributions with complex manifolds (Figure   \ref{fig:rw_examples})  substantially challenge all methods (Appdx.  Table \ref{tab:all_metrics_rw}). 
 Even per-dataset-fit GMM, 
 which achieves the best mean MMD and average rank, exhibits a two-sample rejection rate of $77.1\%$ on real-world data, up from $31.3\%$ on mixed-prior data.

To isolate the effects of distribution shift, we additionally
evaluate 96 synthetic OOD datasets with unseen GMM component
counts, SCM activation functions, and Copula marginals
(Appendix \ref{app:ood}).
For \methodft, MMD increases from $0.56$ to $1.82$
($\times 10^{-3}$), density Spearman decreases from $0.90$
to $0.71$, and score cosine from $0.78$ to $0.58$.
While fine-tuning substantially improves OOD performance,
it cannot introduce new program structures or mechanisms.
These findings motivate extending the program vocabulary
to improve generalization. In particular, incorporating
additional distribution families, functional mechanisms,
and structural templates would allow \method to represent
a broader range of real-world distributions.


\begin{figure}
\vspace{-.2in}
    \centering
    \includegraphics[width=\linewidth]{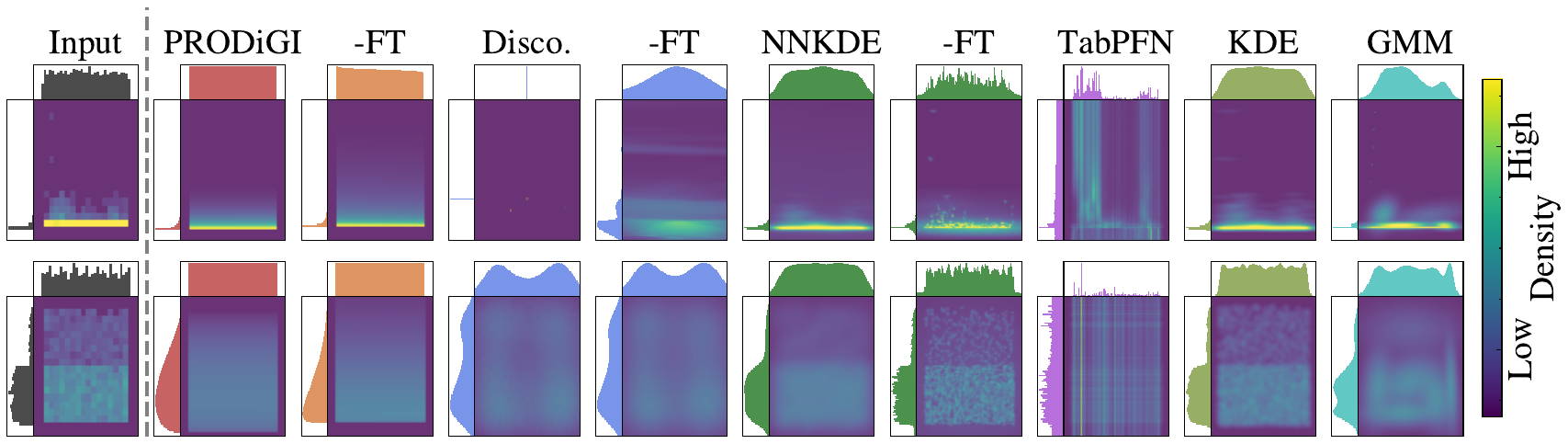}
   \vspace{-0.25in}
    \caption{Example real-world data vs. baseline generations, with corresponding marginals for (top) \textsf{FinancialBeneficiary} and (bottom) \textsf{FinancialTransaction} datasets (also see Fig.   \ref{fig:rw_examples}). 
    }
    \label{fig:gen_rw_heatmap}
    \vspace{-0.15in}
\end{figure}









\subsection{Ablation Results and Analyses}

\method features a non-autoregressive (N-AR) decoder, and normalized and canonically serialized programs for standardized training and deterministic decoding, respectively.


\begin{wrapfigure}{r}
{0.6\textwidth}
    \centering
    \vspace{-0.2in}
    \begin{minipage}[b]{0.25\linewidth}
        \centering
        \includegraphics[width=1\linewidth]{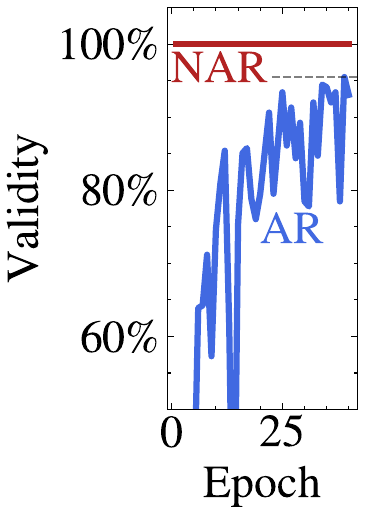}
    \end{minipage}\hfill
    \begin{minipage}[b]{0.25\linewidth}
        \centering
        \includegraphics[width=1\linewidth]{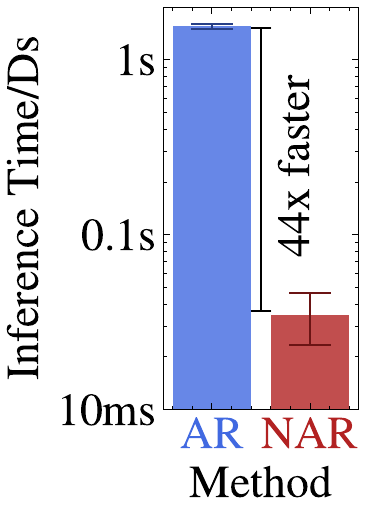}
    \end{minipage}\hfill
    \begin{minipage}[b]{0.25\linewidth}
        \centering
        \includegraphics[width=1\linewidth]{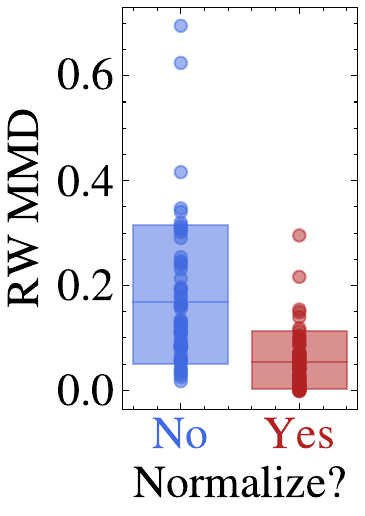}
    \end{minipage}\hfill
    \begin{minipage}[b]{0.25\linewidth}
        \centering
        \includegraphics[width=1\linewidth]{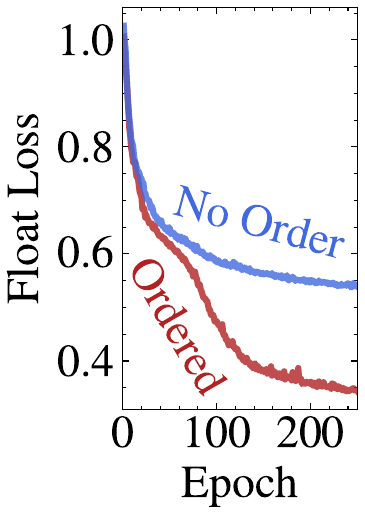}
    \end{minipage}
    \vspace{-0.2in}
    \caption{Ablations of \method (on 20-dimensional data); left to right: (1) N-AR vs AR decoding validity and speed, (2) generation MMD w/ and w/out normalization, (3) decoding loss w/ and w/out canonical program serialization.
    }
    \label{fig:ablations}
    \vspace{-0.15in}
\end{wrapfigure}
Figure \ref{fig:ablations} shows that {N-AR decoding ensures  syntactically valid programs, as well as  faster inference.}  
AR decoding, in contrast, cannot guarantee  validity and is $44\times$ slower.
{Program normalization during training enables varying input data scales and lowers  generation MMD by $68\%$ on real-world (RW) data.} Finally, 
{canonical serialization resolves reconstruction ambiguity; the loss plateaus prematurely without it, as the model struggles to determine module sequence.}



\begin{wrapfigure}{r}{0.6\linewidth}
    \centering
    \begin{minipage}[b]{0.66\linewidth}
        \centering
        \includegraphics[width=\linewidth]{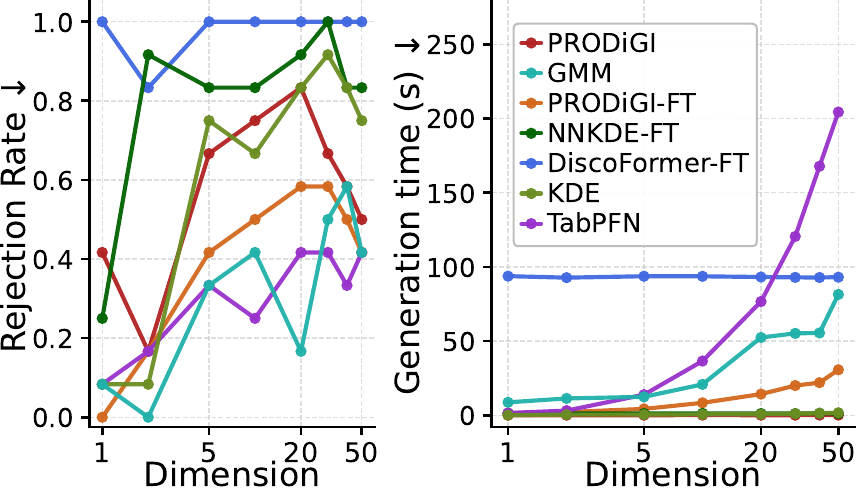}
    \end{minipage}\hfill
    \begin{minipage}[b]{0.32\linewidth}
        \centering
        \includegraphics[width=\linewidth]{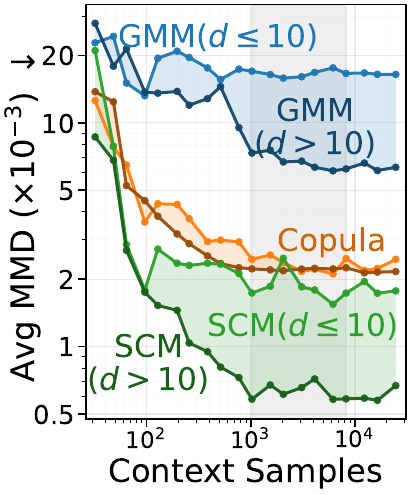}
    \end{minipage}
    \vspace{-0.05in}
     \caption{
   (left) Two-sample rejection rate vs. input dimensionality $d$; (middle) Running time vs. $d$; (right) Average MMD per prior under $d\leq 10$ vs. $d>10$ as {the} number of context  samples is increased. 
    }
    \label{fig:variations}
    \vspace{-1.0em}
\end{wrapfigure}
Finally, we analyze performance and runtime across input dimensionality $d$. Figure \ref{fig:variations} (left) shows that the two-sample rejection rate increases with $d$ for all methods, reflecting harder distribution matching in higher dimensions. Regarding runtime (Figure \ref{fig:variations}, middle), \method and its fine-tuned variant scale gracefully with $d$. In contrast, TabPFN’s runtime grows significantly due to its autoregressive mechanism, while DiScoFormer is dominated by its iterative sampling algorithm.
 In addition, Figure \ref{fig:variations} (right) shows that \method's inference improves with larger context sample sizes, particularly for high-dimensional inputs. This behavior aligns with standard sample complexity theory, in which the required sample size scales with input dimensionality.








\vspace{-0.05in}
\section{Conclusion}
\vspace{-0.075in}
\label{sec:conclusion}
We introduced a data-to-program paradigm through \method, a model pretrained on diverse data priors to infer diverse executable generative programs from empirical data in one forward pass. At its core, canonical serialization, program normalization, and non-autoregressive template decoding enable effective and efficient training and inference. 
Experiments show that \method achieves lower average density and score ranks than existing pretrained models while offering multi-fold speedups. Further, its novel program-space fine-tuning adapts only the inferred program parameters with pretrained backbone fixed, improving distribution match by 90\%. By mapping data directly to explicit statistical representations, \method presents a new direction for tabular generative modeling.

\clearpage
\newpage

\subsection*{AI use statement}


In this work, we used generative AI tools to interpret results, implement and debug program code, and edit our research paper to improve readability. The remaining tasks requiring disclosure do not apply to our paper. We have reviewed all AI-assisted work: LLM-generated code was manually verified and tested for correctness. We have made sure that every AI interpretation is accurate and that readability changes do not alter the meaning of our text. We take responsibility for the final content of this work,
including text, claims or artifacts produced with the aid of generative AI.




\subsection*{Reproducibility statement}

To support reproducibility, we publish both training and inference code, all used datasets, and the trained \method checkpoint at \url{https://github.com/psorus/PRODiGI}. We further provide detailed results for each baseline in Appendix \ref{app:detailed_tables}.



\bibliographystyle{plain}
\bibliography{refs}

\appendix
\section*{Appendix}

\section{Extended Related Work}
\label{app:related}
\textbf{Density Estimation.~}
Density estimation has a long history, spanning classical statistical estimators and modern neural approaches. Gaussian mixture models (GMMs) date back to Pearson \cite{pearson1894gaussianmixture} and represent densities as weighted combinations of Gaussian components. Kernel density estimation (KDE) instead constructs a nonparametric estimate by placing smoothing kernels around observed samples \cite{rosenblatt1956remarks,parzen1962estimation}. Neural density estimators (NDEs) learn flexible density representations through neural networks, including autoregressive models that factorize the joint density into conditional distributions \cite{germain2015made}, and normalizing flows that transform a simple base distribution through invertible mappings \cite{papamakarios2021normalizing}. Conventionally, these approaches construct an estimator separately for each dataset, through parameter fitting or data-dependent smoothing. In contrast, \method amortizes this construction by mapping empirical samples directly to an explicit generative program.


\textbf{Tabular Generative Modeling.~}
Tabular generative modeling aims to synthesize records that preserve feature distributions and dependencies across heterogeneous numerical and categorical variables. Representative neural approaches include CTGAN and TVAE \cite{xu2019modeling}, which adapt generative adversarial networks and variational autoencoders to tabular data. More recently, TabDDPM \cite{kotelnikov2023tabddpm} and TabDiff \cite{shi2025tabdiff} develop diffusion processes for numerical and categorical features, while TabSyn \cite{zhang2024mixed} performs diffusion in a learned VAE latent space. Language-model-based approaches such as GReaT \cite{borisov2023language} instead fine-tune pretrained language models on textual representations of table rows and generate records autoregressively. These methods require dataset-specific training or fine-tuning of a neural generator. 
In comparison, \method infers an explicit generative program from a new dataset in a single forward pass, at the same time, its current formulation is restricted to continuous distributions and does not support categorical variables.

\textbf{Pretrained Models for Generation and Estimation.~}
Pretrained tabular models differ substantially in their formulations of what they infer and which operations they support directly. 
DiScoFormer \cite{ilin2026discoformer}  is trained to predict density values and score vectors at query points through cross-attention to an input dataset; generation requires coupling these predictions with a sampling procedure. TabPFN \cite{hollmann2025accurate} instead repurposes its pretrained classification and regression models through an autoregressive factorization, $\hat{p}(\bx\mid\mathcal{D})=\prod_{j=1}^{d}\hat{p}(x_j\mid x_{<j},\mathcal{D})$. Density evaluation combines the conditional predictions, while generation samples features sequentially, requiring repeated model evaluations.

Other approaches explicitly transfer knowledge across datasets include Neural Statistician \cite{edwards2016neural}, which infers a posterior over a dataset-level latent context that conditions a shared neural generator, enabling direct sampling but requiring latent-variable marginalization for density and score evaluation. Few-shot autoregressive image models \cite{reed2018fewshot} learn densities conditioned on small support sets and generate samples sequentially. Data-driven deep density estimation (DDE) \cite{puchert2021data} pretrains on synthetic distributions to predict point-wise densities, using separate models for each dimensionality; sampling and score estimation require additional procedures. Score Neural Operator \cite{liao2024score} maps distribution embeddings to time-dependent score functions and supports generation through iterative sampling, without directly predicting density values. Pretrained adaptive KDE \cite{zhang2026adaptive} instead predicts sample-specific bandwidth matrices, while retaining KDE as the nonparametric estimator, yielding an explicit kernel mixture that supports density and score evaluation and direct sampling, while retaining the observed samples as kernel centers. Like DDE, adaptive KDE requires training a separate model for each dimensionality.

Table~\ref{tab:pretrained-capabilities} summarizes the capabilities of pretrained models relevant to three statistical tasks; data generation, density estimation and score vector estimation.

\begin{table*}[t]
    \centering
    \small
    \setlength{\tabcolsep}{4pt}
    \renewcommand{\arraystretch}{1.15}
    \caption{
        Capabilities relevant to generation/samling, density
        estimation, and score-vector estimation.
        Additional procedures are distinguished from direct/native model outputs.
      }
      \vspace{-0.1in}
    \label{tab:pretrained-capabilities}
    \begin{tabularx}{\textwidth}{@{}c*{3}{>{\centering\arraybackslash}X}@{}}
        \toprule
        Method / \textbf{Task}
        & \textbf{Generation}
        & \textbf{Density}
        & \textbf{Score vector} \\
        \midrule
  Neural Statistician \cite{edwards2016neural}
        & Direct conditional generation
        & Numerical latent marginalization
        & Numerical marginal-score est. \\

        DDE \cite{puchert2021data}
        & External sampler
        & Direct prediction
        & Additional estimation \\

        Score Neural Operator \cite{liao2024score}
        & Iterative sampling
        & Additional likelihood computation
        & Time-conditioned score prediction\textsuperscript{b} \\

        Adaptive NNKDE \cite{zhang2026adaptive}
        & Direct mixture sampling
        & Explicit kernel density
        & Analytic score \\

        TabPFN \cite{hollmann2025accurate}
        & Autoregressive sampling
        & Autoregressive product of conditional densities
        & Not a native output\textsuperscript{a} \\

          DiScoFormer \cite{ilin2026discoformer}
        & External  sampler
        & Direct prediction
        & Direct prediction \\
 
        \midrule
        \method
        & Direct program sampling
        & Program evaluation
        & Program evaluation \\
        \bottomrule
    \end{tabularx}

    \vspace{2pt}
    \begin{minipage}{\textwidth}
        \footnotesize
        \textsuperscript{a}TabPFN provides conditional probability
        predictions, not native score vectors; score evaluation requires
        a separately specified estimation procedure.
        \textsuperscript{b}Score comparisons must match the target density's
        noise level and input space; latent-space scores are not directly
        comparable to input-space scores.
    \end{minipage}
\end{table*}

\section{Details on Data Priors}
\label{app:priors}
\method supports three different prior families, Gaussian Mixtures, Structural Causal Models, and Copulas, which we detail as follows.

\subsection{Gaussian Mixtures}

We simulate data by drawing from multivariate GMMs with
varying number of components and dimensions. 
$m$-components in $d$ dimensions; with centroids $\bmu_{j}^{(k)} \sim \mathcal{N}(0,5)$ and diagonal covariance $\bSigma_{jj}^{(k)} \sim \mathcal{U}(0.5,2)$ for $k\in [m]$ and $j \in [d]$.Each component is assigned a random weight $\bomega^{(k)}$. The weights are sampled such that each weight is at least $1\%$, and the pairwise difference between any two component weights is also at least $1\%$. These constraints become increasingly restrictive as the number of components grows. The remaining degrees of freedom are sampled from a Dirichlet distribution with a randomly selected concentration parameter $\alpha \in [0.1, 10.0]$.
 We vary $m \in [1,M]$ and $d \in [1,D]$ uniformly as we draw new datasets, with $M=10$ and $D=50$.

Table \ref{tab:gmmhps} provides hyperparameter configurations for sampling datasets from the GMM prior.

\begin{table}[h!]
\centering
\caption{Hyperparameters for Gaussian Mixture Model (GMM) Data Prior}
\begin{tabular}{lll}
\toprule
\textbf{Hyperparameter} & \textbf{Values} & \textbf{Description} \\
\midrule
$m$ & $[1,10]$ & Number of mixture components \\
$d$ & $[1,50]$ & Dimensionality of data \\
$\bmu_{j}^{(k)}$ & $\sim\mathcal{N}(0,5)$ & Component means \\
$\bSigma_{jj}^{(k)}$ & $(0.5,2]$ & Diagonal variances \\ 
$\bomega$ & $(0.01,1.0-0.01\cdot \tfrac{m\cdot(m+1)}{2}]$ & Component weights\\
\bottomrule
\end{tabular}
\label{tab:gmmhps}
\end{table}

\subsection{Structural Causal Models}
\label{app:scms}
 Structural causal models (SCMs) represent the generative dependencies among variables through causal graphs and structural equations. An SCM is defined by a directed acyclic graph
\( G = (V, E) \), where each node \( j \in V \) corresponds to a variable, and causal mechanisms linking them.

For SCM data, we first instantiate a causal graph $G$ as a growing random network. We iteratively add new nodes and connect them to $k$ existing nodes. We draw $k$ uniformly from $k\in [0,\mathrm{MD}]$ and limit the depth of such generated graphs to MP, where MP is drawn uniformly from $\mathrm{MP} \in [1,5]$. Each edge is assigned a random weight $\bomega_{i,j}\in [0.5,1.5] \cup [-1.5,-0.5]$ and each node is assigned an activation function $a_j$, a random noise factor $\log(\eta_j)\sim \mathcal{U}(-1,0)$, two bias variables $b_j^0, b_j^1 \sim \mathcal{N}(0,1)$ and an output scale $\delta_j \sim \mathcal{U}(0.5,1.5)$.
This allows us to construct a structural equation for each node as
\begin{equation}
    X_j
= \delta_j a_j\!\left(
\sum_{i\in\mathrm{Pa}(X_j;G)} \bomega_{i,j}X_i + s_j^0
\right)
+ \epsilon_j + s_j^1\;,
\qquad
\epsilon_j\sim\mathcal{N}(0,\eta_j).
\end{equation}
where \( \mathrm{Pa}(X_j; G) \) denotes the set of parent (direct cause) variables of \( X_j \) in \( G \).

 Table \ref{tab:scmhps} provides the hyperparameter settings of the SCM priors. We visualize each activation function in Figure~\ref{fig:app_scmfunc}.

\begin{table}[h!]
\centering
\caption{Hyperparameters for Structural Causal Model (SCM) based Data Priors}
\begin{tabular}{lll}
\toprule
\textbf{Hyperparameter} & \textbf{Values} & \textbf{Description} \\
\midrule
MD & $[1,5]$ & Max. In Degree, Max. number of nodes that \\ & &  directly influence any node\\
MP & $[1,5]$ & Max. Path Length, Depth of the graph \\
$d$ & $[1,50]$ & Dimensionality of data/Number of nodes \\
$w_{i,j}$ & $[0.5,1.5]\cup[-1.5,0.5]$ & Edge factors\\
$\eta_i$ & $[0.1,1.0]$ & Gaussian noise factor besides causal relation.\\
$s_j^0$ & $\sim\mathcal{N}(0,1)$ & Shift inside the activation function\\
$s_j^1$ & $\sim\mathcal{N}(0,1)$ & Shift outside the activation function\\
$\delta_j$ & $[0.5,1.5]$ & Node scaling factor\\
$a(\cdot)$ & $\{\text{tanh, odd, even}\}$ & Activation function for structural equations  \\
\bottomrule
\end{tabular}
\label{tab:scmhps}
\end{table}

\begin{figure}
    \centering
    \includegraphics[width=0.7\linewidth]{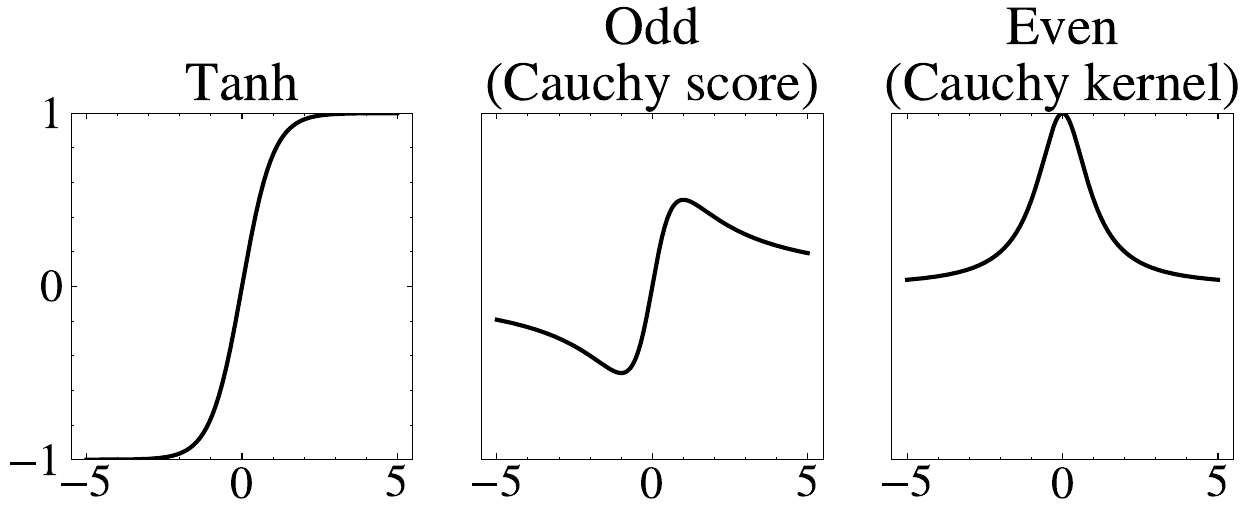}
    \vspace{-0.1in}
    \caption{SCM activation functions used in this paper. We limit our choice to three functions to make sure their behavior is seperable. Odd represents the Cauchy score functions $\tfrac{x}{1+x^2}$ and Even represents the Cauchy kernel ($\tfrac{1}{1+x^2}$).}
    \label{fig:app_scmfunc}
\end{figure}

\subsection{Copulas}

Real-world univariate features often exhibit skewed, heavy-tailed behavior \cite{clauset2009power}. While SCMs induce non-Gaussian marginals, they do not provide explicit control over feature-wise distributions; we therefore adopt copula models \cite{nelsen2006introduction,houssou2022generation}. By Sklar’s theorem \citep{sklar1959fonctions}, any joint CDF of continuous variables $\{X_j\}_{j=1}^d$ with marginals $F_j$ can be expressed as
\[
F(x_1,\ldots,x_d) = C\!\left(F_1(x_1),\ldots,F_d(x_d)\right),
\]
enabling independent specification of marginals and dependence structure. We sample each feature’s marginal from a diverse pool of parametric distributions (Gaussian, Uniform, Kumaraswamy, Exponential, Student’s $t$), and use a Gaussian copula for $C$. Examples of each marginal function are shown in Figure~\ref{fig:app_marginals}. We further sample a correlation matrix $R\in\mathbb{R}^{d\times d}$ from the Lewandowski–Kurowicka–Joe (LKJ) distribution~\cite{lkj_dist_Lewandowski09} to guarantee nontrivial correlation structures even in high dimensions.

Given $C(R)$ and $\{F_j\}_{j=1}^d$, data samples are generated by drawing $\mathbf{u} \sim C(R)$ on $[0,1]^d$ and transforming $x_j = F_j^{-1}(u_j)$ for all $j \in [d]$. We widely vary the marginal families, their parameters, and the copula to synthesize diverse tabular  datasets.

Table \ref{tab:copulahps} provides the hyperparameter settings of the Copula priors.

\begin{table}[h!]
\caption{Hyperparameters for Copula Data Prior}
\centering
\scalebox{0.9}{
\begin{tabular}{l c l}
\hline
\textbf{Hyperparameter} & \textbf{Values} & \textbf{Description} \\
\toprule
$C$  & $\text{Gaussian}$ & Copula parameter family \\
$d$ & $[1,50]$ & Dimensionality of data/Number of marginals \\

Marginal & $\{\text{Uniform, Normal,  Exponential, }$ & Marginal family for each feature \\
 & $ \text{Student t, Kumaraswamy}\}$ &  \\
$\mu$/$\sigma$ & $[-5, 5]$/$[0.5, 2.0]$ & Mean/Variance for each marginal\\
$\nu$ & $[2.5, 10]$ & Degrees of freedom for Student’s $t$ distribution\\
$a$,$b$ & $[0.5,5.0]$ & Shape parameters of the Kumaraswamy distribution.\\
\bottomrule
\end{tabular}
}
\label{tab:copulahps}
\end{table}

\begin{figure}
    \centering
    \includegraphics[width=0.7\linewidth]{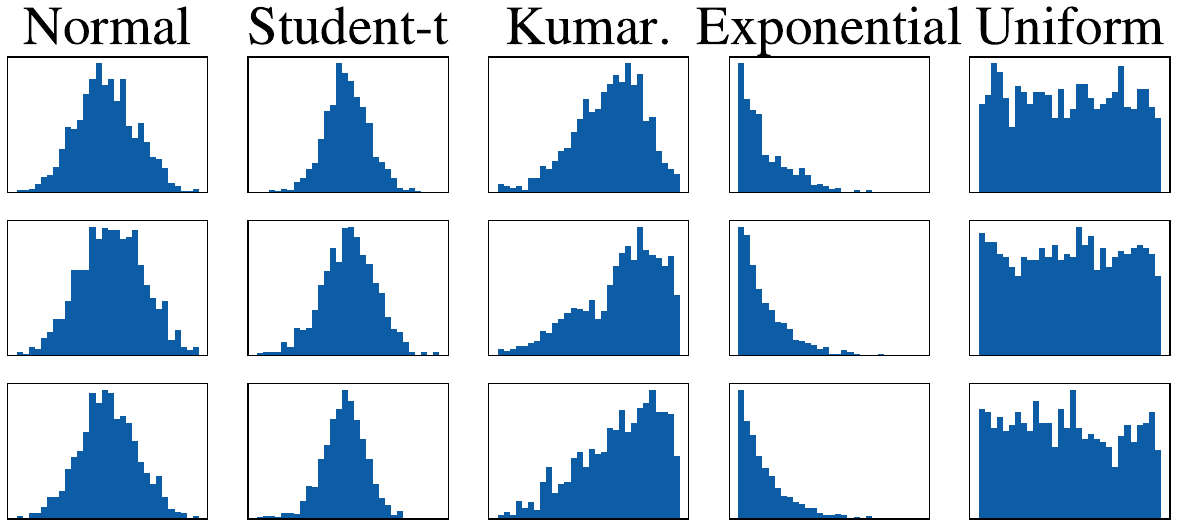}
    \vspace{-0.1in}
    \caption{Examples of each copula marginal used in this paper.}
    \label{fig:app_marginals}
\end{figure}


\section{Normalization}
\label{app:normalization}



To remove variation in feature location and scale during training, we \textbf{normalize each sampled generative program}. Specifically, we transform its parameters to target zero mean and unit variance for each generated feature. This transformation uses analytical moments for GMMs and Copula marginals, and sample estimates for SCMs.
Normalizing the program itself ensures that the generated data and the target program share the same coordinate system.

\textbf{During training}, we  sample training datasets from the normalized programs, which readily generate normalized datasets, and train the model to recover the corresponding normalized program.

\textbf{During inference}, we first standardize each input feature using its empirical mean and standard deviation, then infer a normalized program with a single forward pass. Finally, we apply the inverse transformation to the inferred program
using the input dataset's empirical feature means and standard deviations,
restoring the original feature locations and scales.

We describe the transformations for normalizing each prior family below, assuming finite, strictly positive marginal variances.

\subsection{Gaussian Mixtures}

We normalize a GMM using its global mean and feature-wise standard deviations. For component weights $\omega^{(k)}$, means $\boldsymbol{\mu}^{(k)}$, and diagonal covariance matrices $\boldsymbol{\Sigma}^{(k)}$, these are
\begin{equation}
\begin{aligned}
\boldsymbol{\mu}_{\mathrm{glob}}
&= \sum_k \omega^{(k)} \boldsymbol{\mu}^{(k)},\\
\boldsymbol{\sigma}_{\mathrm{glob}}^2
&= \sum_k \omega^{(k)}
\left[
\operatorname{diag}\bigl(\boldsymbol{\Sigma}^{(k)}\bigr)
+ \bigl(\boldsymbol{\mu}^{(k)}
- \boldsymbol{\mu}_{\mathrm{glob}}\bigr)^2
\right],
\end{aligned}
\end{equation}
where $\operatorname{diag}(\cdot)$ extracts the diagonal as a vector, and vector squares are element-wise. Let $D$ be the diagonal matrix with diagonal $\boldsymbol{\sigma}_{\mathrm{glob}}$. The normalized component parameters are
\begin{equation}
\begin{aligned}
\boldsymbol{\mu}^{(k)}_{\mathrm{norm}}
&= D^{-1}
\bigl(\boldsymbol{\mu}^{(k)}-\boldsymbol{\mu}_{\mathrm{glob}}\bigr),\\
\boldsymbol{\Sigma}^{(k)}_{\mathrm{norm}}
&= D^{-1}\boldsymbol{\Sigma}^{(k)}D^{-1}.
\end{aligned}
\end{equation}
The mixture weights remain unchanged.

\subsection{Structural Causal Models}

For general nonlinear SCMs, feature means and standard deviations are not readily available analytically. We therefore estimate them using $10{,}000$ samples from each sampled SCM, obtaining $\hat{\mu}_i$ and $\hat{\sigma}_i$ for feature $i$.

For structural equations of the form (See Appendix~\ref{app:scms} for notation.)
\begin{equation}
X_j
= \delta_j a_j\!\left(
\sum_{i\in\mathrm{Pa}(X_j;G)} \omega_{j,i}X_i + s_j^0
\right)
+ \epsilon_j + s_j^1,
\qquad
\epsilon_j\sim\mathcal{N}(0,\eta_j)\;, 
\end{equation}
Substituting
$X_j=\hat{\mu}_j+\hat{\sigma}_j X_{j,\mathrm{norm}}$
for the child and
$X_i=\hat{\mu}_i+\hat{\sigma}_i X_{i,\mathrm{norm}}$
for each parent $i\in\mathrm{Pa}(X_j;G)$
yields the normalized parameters:
\begin{equation}
s^1_{j,\mathrm{norm}}
= \frac{s_j^1-\hat{\mu}_j}{\hat{\sigma}_j},
\qquad
\delta_{j,\mathrm{norm}}
= \frac{\delta_j}{\hat{\sigma}_j},
\qquad
\eta_{j,\mathrm{norm}}
= \frac{\eta_j}{\hat{\sigma}_j^2},
\end{equation}
and
\begin{equation}
\omega_{j,i,\mathrm{norm}}
= \omega_{j,i}\hat{\sigma}_i,
\qquad
s^0_{j,\mathrm{norm}}
= s_j^0
+ \sum_{i\in\mathrm{Pa}(X_j;G)}
\omega_{j,i}\hat{\mu}_i.
\end{equation}
These updates preserve the causal graph and express the structural equations in the normalized coordinates. Because the moments are estimated, the normalized SCM has approximately zero marginal means and unit marginal variances.

\subsection{Copulas}

For a Copula-based joint distribution, we standardize each marginal. Specifically, for each marginal $F_i$, we analytically compute its mean
$\mu_i$ and standard deviation $\sigma_i$ from its sampled
parameters, and standardize its quantile function as follows:
\begin{equation}
F_{i,\mathrm{norm}}^{-1}(u)
= \frac{F_i^{-1}(u)-\mu_i}{\sigma_i}\;.
\end{equation}
These strictly increasing affine transformations preserve the Copula.

Uniform, Normal, and Exponential marginals reduce to unique standardized distributions:
$\mathcal{U}(-\sqrt{3},\sqrt{3})$,
$\mathcal{N}(0,1)$, and $E-1$ with
$E\sim\mathrm{Exp}(1)$, respectively.
Standardized Student-$t$ marginals retain their degrees of freedom $\nu>2.5$, while affinely transformed Kumaraswamy marginals retain their two shape parameters.

\section{Program Serialization and Tokenization}
\label{app:programs}

\subsection{What is a Generative Program?}

A \emph{generative program} is a complete specification of a data-generating
process. It consists of a program family, its structural configuration, and
the distributional or functional parameters required to instantiate the
generator. We consider three heterogeneous program families: Gaussian Mixture
Models (GMMs), Structural Causal Models (SCMs), and Copulas.

We represent each program as a structured sequence of tokens, analogous to a
domain-specific language (DSL). This representation preserves the semantic
role of each parameter while allowing programs with different structures and
dimensionalities to be represented in a common format.

\subsection{Program Serialization}

Each concrete program $P$ is serialized into an ordered sequence
\[
    S(P)=(z_1,\ldots,z_L),
\]
where each element corresponds to a structural field or program parameter.
The serialization is designed to be human-readable while providing a
canonical representation for the program decoder.

\begin{tcolorbox}[ colback=blue!5!white, colframe=blue!75!black, title=\textbf{Example of a GMM Program.}, fonttitle=\bfseries, fontupper=\small ]
\begin{verbatim}
[GMM][DIMENSIONALITY][2][NUM_COMPONENTS][4]
[START_COMPONENT][0][WEIGHT][0.6647]
[INDEX][0][MEAN][-0.3763][DIAG_STD][0.6549]
[INDEX][1][MEAN][-0.5755][DIAG_STD][0.3401][END_COMPONENT]
[START_COMPONENT][1][WEIGHT][0.2407]
[INDEX][0][MEAN][0.6825][DIAG_STD][0.7333]
[INDEX][1][MEAN][0.9778][DIAG_STD][0.3844]
[END_COMPONENT]
[START_COMPONENT][2][WEIGHT][0.0752]
[INDEX][0][MEAN][1.6824][DIAG_STD][0.7456]
[INDEX][1][MEAN][2.2755][DIAG_STD][0.4177]
[END_COMPONENT]
[START_COMPONENT][3][WEIGHT][0.0194]
[INDEX][0][MEAN][1.6918][DIAG_STD][0.7129]
[INDEX][1][MEAN][2.4417][DIAG_STD][0.4465]
[END_COMPONENT][END_GMM]
\end{verbatim}
\end{tcolorbox}

The example specifies a $2$-dimensional GMM with four components. Each
component has its own mean vector, covariance representation, and mixture
weight. Thus, the serialization describes a
specific generative model, rather than merely identifying the GMM family.


\begin{tcolorbox}[
  colback=blue!5!white,
  colframe=blue!75!black,
  title=\textbf{Example of an SCM Program.},
  fonttitle=\bfseries,
  fontupper=\small
]
\begin{verbatim}
[SCM][DIMENSIONALITY][6][MAX_PARENTS][3]
[START_COMPONENT][INDEX][0][FUNCTION][ODD]
[PARENTS][-1][0][ -1][0][ -1][0]
[NOISE][0.1571][SHIFT][0.2789][-0.241][SCALE][2.5983][END_COMPONENT]
[START_COMPONENT][INDEX][1][FUNCTION][EVEN]
[PARENTS][-1][0][ -1][0][ -1][0]
[NOISE][0.1355][SHIFT][0.3286][-2.3577][SCALE][3.7092][END_COMPONENT]
[START_COMPONENT][INDEX][2][FUNCTION][TANH]
[PARENTS][0][-0.5697][4][0.1237][5][0.8932]
[NOISE][0.1316][SHIFT][-0.5354][0.3304][SCALE][1.2897][END_COMPONENT]
[START_COMPONENT][INDEX][3][FUNCTION][ODD]
[PARENTS][1][-0.1624][-1][0][ -1][0]
[NOISE][0.9306][SHIFT][0.0881][-0.085][SCALE][1.0584][END_COMPONENT]
[START_COMPONENT][INDEX][4][FUNCTION][EVEN]
[PARENTS][-1][0][ -1][0][ -1][0]
[NOISE][0.1013][SHIFT][0.7026][-2.051][SCALE][3.5272][END_COMPONENT]
[START_COMPONENT][INDEX][5][FUNCTION][ODD]
[PARENTS][0][-0.4509][1][-0.2481][-1][0]
[NOISE][0.219][SHIFT][-0.076][0.1409][SCALE][2.4045][END_COMPONENT]
[END_SCM]

\end{verbatim}
\end{tcolorbox}

Here, the variables form a DAG: variables $0$ and $1$ and $4$ are root nodes,
variable $3$ depends on variable $1$, variable $5$ depends on variables $0$ and $1$ and variable $2$ depends on variables $0$ ,$4$ and $5$. The serialization, therefore, specifies both the causal structure
and the structural equations.


\begin{tcolorbox}[
  colback=blue!5!white,
  colframe=blue!75!black,
  title=\textbf{Example of a Copula Program.},
  fonttitle=\bfseries,
  fontupper=\small
]
\begin{verbatim}
[COPULA][DIMENSIONALITY][3]
[START_COMPONENT][INDEX][0][MARGINAL][EXPONENTIAL]
[PARAMS][0][1][0][0][END_COMPONENT]
[START_COMPONENT][INDEX][1][MARGINAL][NORMAL]
[PARAMS][0][1][0][0][END_COMPONENT]
[START_COMPONENT][INDEX][2][MARGINAL][NORMAL]
[PARAMS][0][1][0][0][END_COMPONENT]
[CORRELATIONS]
[0.0748][0.3018][-0.5349]
[END_CORRELATIONS][END_COPULA]
\end{verbatim}
\end{tcolorbox}

This program specifies a $3$-dimensional Copula through its dependence
structure and three feature-wise marginal distributions. In particular, the
marginal family and its parameters can differ across dimensions.

\subsection{Templates}

A \emph{template} captures the structure of a program while leaving its
dataset-specific values unspecified. Formally, for a concrete program $P$,
we write
\[
    \tau(P)=(f,\ba),
\]
where $f$ is the program family and $\ba$ contains the discrete structural
attributes that determine the program layout.

The template induces a sequence of typed fill-in slots,
\[
    S_{\tau}=(s_1,\ldots,s_{L_\tau}),
\]
where each slot specifies the semantic role and type of a value to be
predicted. A concrete program is obtained by filling these slots with the
corresponding parameter values.

For example, the GMM program above gives rise to a template of the form

\begin{tcolorbox}[
  colback=blue!5!white,
  colframe=blue!75!black,
  title=\textbf{Example of a GMM Template.},
  fonttitle=\bfseries,
  fontupper=\small
]
\begin{verbatim}
[GMM][DIMENSIONALITY][2][NUM_COMPONENTS][4]
[START_COMPONENT][0][WEIGHT][<float>]
[INDEX][<int>][MEAN][<float>][DIAG_STD][<float>]
[INDEX][<int>][MEAN][<float>][DIAG_STD][<float>][END_COMPONENT]
[START_COMPONENT][1][WEIGHT][<float>]
[INDEX][<int>][MEAN][<float>][DIAG_STD][<float>]
[INDEX][<int>][MEAN][<float>][DIAG_STD][<float>]
[END_COMPONENT]
[START_COMPONENT][2][WEIGHT][<float>]
[INDEX][<int>][MEAN][<float>][DIAG_STD][<float>]
[INDEX][<int>][MEAN][<float>][DIAG_STD][<float>]
[END_COMPONENT]
[START_COMPONENT][3][WEIGHT][<float>]
[INDEX][<int>][MEAN][<float>][DIAG_STD][<float>]
[INDEX][<int>][MEAN][<float>][DIAG_STD][<float>]
[END_COMPONENT][END_GMM]
\end{verbatim}
\end{tcolorbox}

The template determines the number, ordering, semantic role, and type of the
decoder outputs, while the corresponding concrete program supplies their
values.

\subsection{Tokenization}

The serialized program is decomposed into typed tokens. Categorical fields,
such as \texttt{GMM},\texttt{DIMENSIONALITY},  and \texttt{WEIGHT}, are represented
as symbolic tokens. Discrete quantities, such as \texttt{DIM}, \texttt{N\_COMP},
and \texttt{ID}, are represented as integer slots. Vector-valued quantities,
such as \texttt{MEAN}, are decomposed into one real-valued slot per dimension.
Likewise, SCM parent references are represented as integer-valued slots.
We also define special Fill-in tokens to encode templates with missing values.




\subsection{Canonical Ordering}

Each program family is serialized according to a canonical ordering of its
fields and entities. 
This removes permutation ambiguity, providing a consistent reconstruction target during training  and is required to prevent the decoder from regressing to the mean.

We order GMM components by their component weights, while ensuring that two component weights differ by at least $1\%$, and order the parents of an SCM node. For an example of a model without such a canonical order, see Figure \ref{fig:ablations} right.

Positional encoding is used for program-decoder slots, but not for the
observations in the input dataset. The latter are treated as an unordered set
and are encoded by the permutation-invariant Set Transformer.

\subsection{Vocabulary Construction}

We initialize the symbolic vocabulary through a bootstrap procedure. We sample
programs from the GMM, SCM, and Copula priors, serialize and tokenize them, and
collect the symbols, fields, family names, and entity types that occur. The
resulting vocabulary is fixed for training and inference, and each such token is translated into a learnable $512$ dimensional vector.





\newpage
\section{Baseline Details}
\label{app:baseline_details}

\subsection{DiscoFormer}

DiscoFormer \cite{ilin2026discoformer} is a pretrained model for density and score estimation, designed to leverage the structure of analytically tractable distributions. It is trained on a prior over Gaussian Mixture Models (GMMs), using their analytically known log densities and score functions as supervision.

We follow the official implementation of DiscoFormer available at \url{https://github.com/Vilin97/ensemble-score-matching}. Since the authors do not provide pretrained checkpoints, we train a separate DiscoFormer model for each dimensionality considered in our evaluation, $d \in \{1,2,5,10,20,30,40,50\}$, using the default hyperparameters.

DiscoFormer natively provides both log-density estimates and score vectors. However, it does not enforce the normalization of the estimated densities, i.e., there is no guarantee that they integrate to one. In our experiments, we observe that this normalization condition is not satisfied. Consequently, the negative log-likelihood (NLL) is not a meaningful metric for the resulting density estimates.

Since DiscoFormer does not directly support sample generation, we use a Langevin chain \cite{RobertsTweedie1996} to generate samples from the provided density and score estimates. We initialize each chain using context points and run $32$ parallel chains with a step size of $\eta=0.01$. We discard the first $20{,}000$ iterations as burn-in and subsequently retain every $10$-th generated sample.

Furthermore, the original paper describes a fine-tuning procedure for DiscoFormer on a given dataset, which optimizes the model by matching sample scores to those inferred from the corresponding density estimates. While the original paper recommends fine-tuning for up to $3$ epochs per dataset, we observe modest improvements with additional training, with gains continuing up to approximately $10$ epochs. We therefore report results after $10$ epochs of fine-tuning for each dataset.

\subsection{TabPFN}

TabPFN \cite{hollmann2023tabpfn} is a pretrained transformer model for tabular prediction that performs in-context learning by conditioning on a set of labeled examples. Rather than fitting a separate predictive model for each dataset, TabPFN is pretrained on a large prior over synthetic tabular datasets and uses the resulting prior to make predictions for previously unseen datasets. While TabPFN is designed primarily for supervised tasks such as regression and classification, several unsupervised extensions, including generation and density estimation, have also been proposed.

We follow the official implementation of TabPFN available at \url{https://github.com/PriorLabs/TabPFN} (version 8.1.0) and the TabPFN extensions available at \url{https://github.com/PriorLabs/tabpfn-extensions} (version 0.4.2). Both generation and density estimation operate by treating a subset of features as inputs and predicting either the conditional density or the value of another feature. By iteratively repeating this procedure until all features have been predicted, TabPFN can generate new samples and estimate densities.

To obtain score vectors, we rely on the density estimates. Since the score is defined as the gradient of the log-likelihood, we estimate it using finite differences between $2\cdot d$ density predictions:
\[
(\nabla \log(f)(x))_i
\approx
\frac{\log(f)(x+\epsilon\cdot e_i)-\log(f)(x-\epsilon\cdot e_i)}
{2\cdot\epsilon},
\]
where $e_i$ denotes the $i$-th unit vector. This introduces the hyperparameter $\epsilon$, which we set to $\epsilon=0.1$. While this is a relatively large value for numerical differentiation, we found in preliminary testing that smaller values of $\epsilon$ result in score estimates dominated by noise in the density predictions.

The densities produced by TabPFN also do not integrate to one. Consequently, we do not report negative log-likelihood (NLL) values.

TabPFN also supports fine-tuning on individual datasets. However, we found the computational cost of fine-tuning prohibitively high for the scope of our evaluation and therefore use the pretrained model without further fine-tuning.

\subsection{GMM}\label{app:baseline:gmm}

Gaussian Mixture Models (GMMs) provide a flexible parametric family of density models for which the density, score function, and sampling procedure are all available analytically. We fit a GMM to each dataset and use the resulting model to obtain both density and score estimates. In contrast to the pretrained approaches considered above, the GMM is fitted directly to the data using the EM algorithm \cite{dempster1977maximum} and does not involve any pretraining or fine-tuning.

To select the number of mixture components, we consider GMMs with between $1$ and $10$ components and select the model with the lowest Bayesian Information Criterion (BIC). For each candidate number of components, we use five random initializations and select the resulting model according to the BIC criterion. The selected GMM provides the density and score estimates used in our evaluation. Since both quantities are available analytically for the fitted mixture model, no numerical approximation is required.

For sample generation, we first sample a mixture component according to its fitted mixing weights and then draw a sample from the corresponding Gaussian component.

In contrast to the GMM components predicted by our model, this baseline GMM estimates a full covariance matrix rather than restricting the components to diagonal covariance matrices. This increases the number of parameters for a $50$-dimensional GMM with $10$ components from $1010$ to $13260$, making our baseline GMMs more expressive.

\subsection{KDE}

Kernel Density Estimation (KDE) provides a non-parametric approach to density estimation by representing the density as a mixture of kernel functions centered at the observed data points. 

We select the KDE bandwidth using $5$-fold cross-validation. Specifically, we consider bandwidth scaling factors $\alpha \in \{0.1,0.2,0.3,0.5,0.75,1.0,2.0,3.0,5.0,7.5,10.0\}$ and select the value that maximizes the average log-likelihood on the held-out folds. 

The KDE density and score are computed analytically according to the corresponding Gaussian mixture formulation, treating each observed data point as the center of a Gaussian component. Consequently, the density and score expressions follow directly from those of a Gaussian mixture model, with the KDE bandwidth determining the covariance of each component.

To generate new samples, we first select a data point from the observed data uniformly at random and then add Gaussian noise with the selected KDE bandwidth. Thus, each generated sample corresponds to a draw from one of the Gaussian kernels comprising the KDE.

\subsection{NNKDE}

Neural Network Guided Kernel Density Estimation (NNKDE) \cite{zhang2026adaptive} is a non-parametric density estimation method that uses a neural network to predict a local bandwidth for each observed data point. We use the official code and pretrained NNKDE models corresponding to each dimensionality considered in our evaluation, except for a $40$-dimensional model, which we train from scratch using the provided code and hyperparameters. Before applying NNKDE, we whiten each dataset using a ZCA whitening transformation and transform the resulting density and score estimates back to the original data space. While the original paper does not propose such a whitening transformation, we found it necessary to achieve competitive performance with NNKDE.
 
For each training point, the neural network predicts a local covariance matrix, which defines the corresponding Gaussian kernel. The NNKDE density is then obtained as the mixture of these Gaussian kernels, with each training point serving as a kernel center. The density and score are computed analytically from the resulting Gaussian mixture, following the same formulation as for a GMM with a Gaussian component centered at each training point, but with a distinct, locally adapted covariance matrix for each component.

To generate new samples, we first select a training point uniformly at random and then draw Gaussian noise using the corresponding local covariance matrix. The generated sample is given by the selected training point plus this noise, followed by the inverse whitening transformation to map the sample back to the original data space. No fine-tuning is performed.

The NNKDE model also supports dataset-specific fine-tuning through a global bandwidth-scale calibration procedure (Eq.~9--10 in the original paper). Specifically, a single global scale parameter is optimized by minimizing the self-exclusive leave-one-out (LOO) negative log-likelihood (NLL) of the target dataset, thereby rescaling the local bandwidths predicted by the pretrained model. We perform this optimization for $1000$ gradient descent steps. This remains computationally inexpensive in our experiments and provides sufficient iterations for the optimization to converge reliably.

\section{Additional Experiments}
\label{sec:other}

\subsection{Case Studies}\label{app:casestudy}

We investigate whether the quality of the reconstructed datasets extends beyond matching the observed data distribution to recovering the underlying distribution parameters. Figure \ref{fig:prior_examples} presents several examples of prior dataset reconstructions. In addition to closely approximating the overall dataset distribution, our method accurately recovers the individual GMM components, Copula marginals, and SCM structure.




\begin{figure}[htbp]
    \centering
    \begin{subfigure}[b]{0.32\linewidth}
        \centering
        \includegraphics[width=\linewidth]{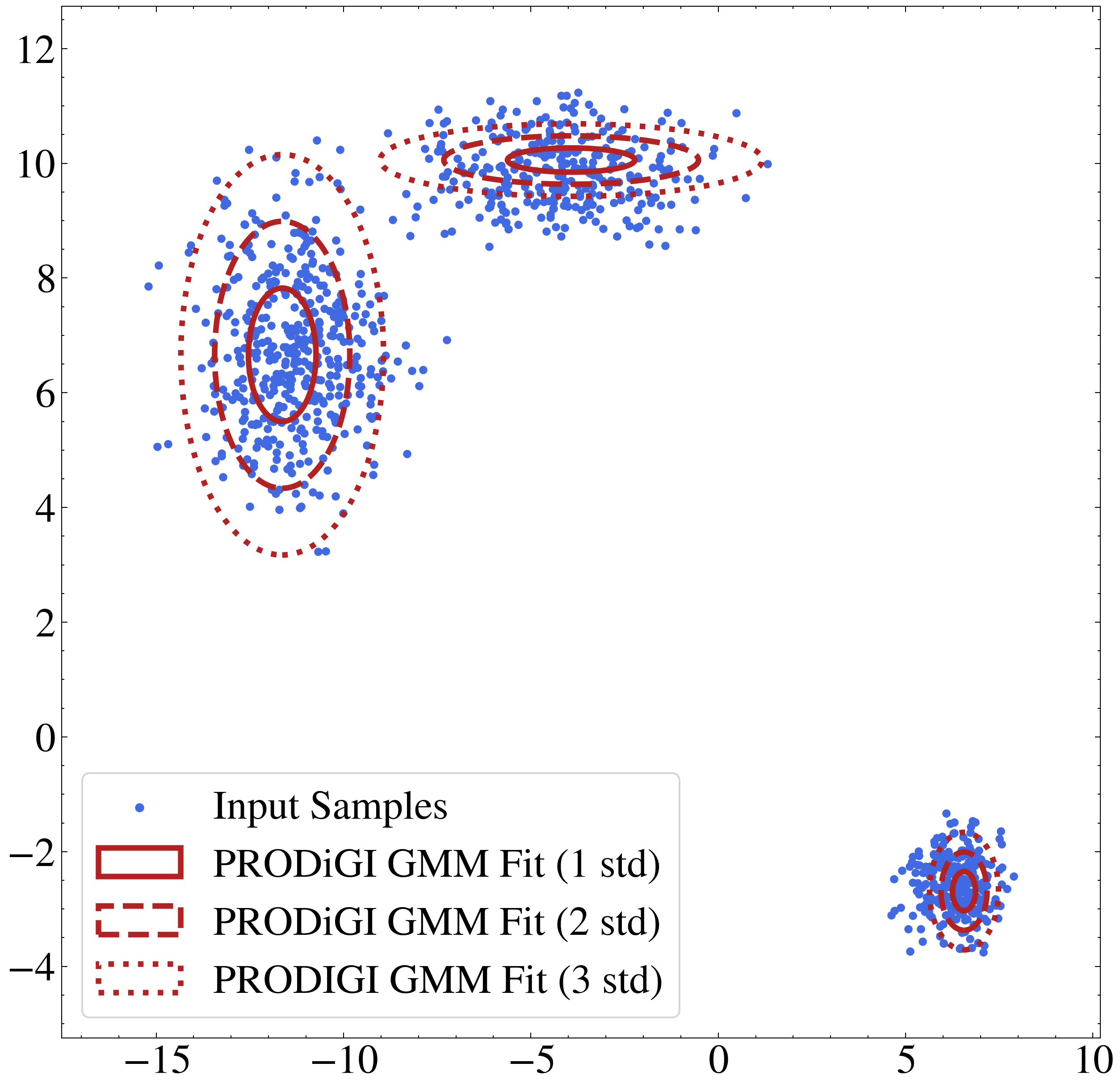}
        \caption{GMM}
        \label{fig:sub1}
    \end{subfigure}
    \hfill
    \begin{subfigure}[b]{0.32\linewidth}
        \centering
        \includegraphics[width=\linewidth]{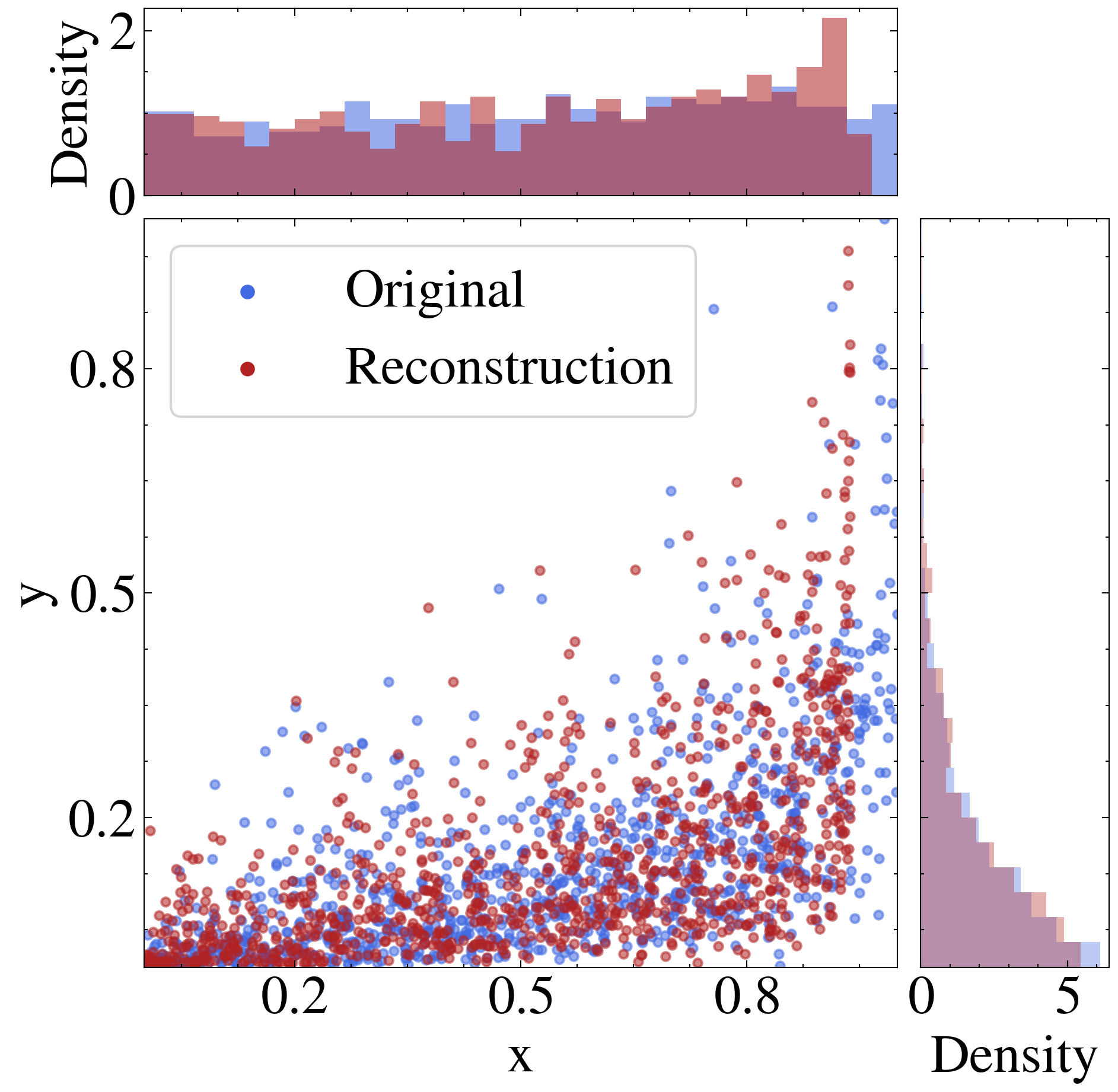}
        \caption{Copula}
        \label{fig:sub2}
    \end{subfigure}
    \hfill
    \begin{subfigure}[b]{0.32\linewidth}
        \centering
        \includegraphics[width=\linewidth]{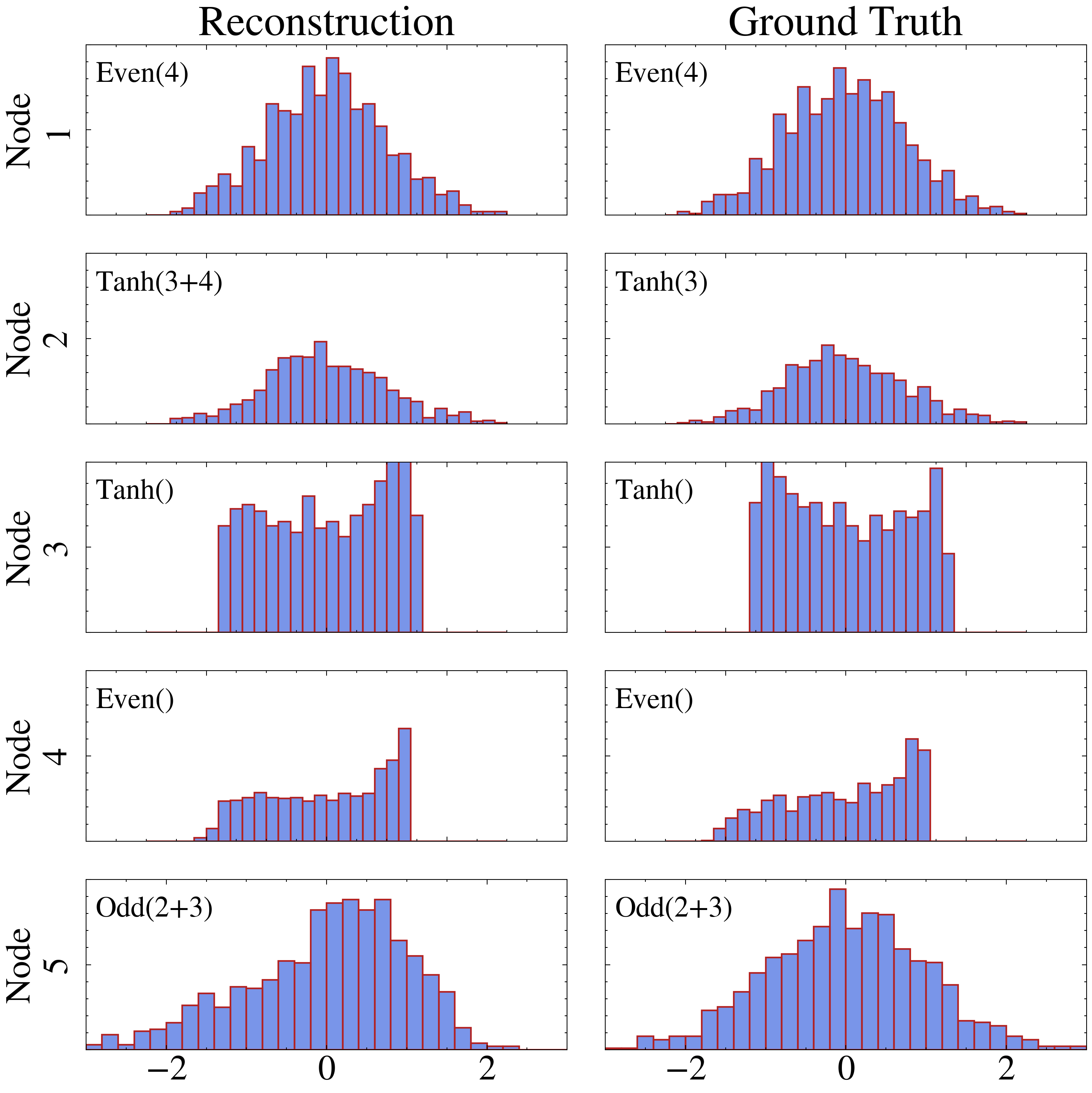}
        \caption{SCM}
        \label{fig:sub3}
    \end{subfigure}
    \caption{Example inferred programs and generations by \method for each prior family: \textbf{(a) A 2-d GMM} with 3 components; \textbf{(b) A 2-d Copula} with a uniform and a long-tail marginal, respectively; \textbf{(c) A 5-d SCM} with structural mechanisms \texttt{odd}, \texttt{even}, \texttt{tanh} with (right) ground-truth and (left) inferred parents per node/feature in parentheses (empty implies root variable).}
    \label{fig:prior_examples}
\end{figure}



We further investigate cases in which \method predicts an incorrect prior family. While \method correctly identifies the prior family for the vast majority of in-distribution examples, Figure \ref{fig:casestudy} shows that approximately $0.13\%$ of cases are assigned to a different prior family. We examine one such example in Figure \ref{fig:casestudy}, where the input distribution is generated by a two-dimensional SCM, whereas \method reconstructs it using a Copula. Despite this discrepancy in prior family, the reconstructed distribution and marginals closely match those of the input. This suggests that these cases may arise from overlap among our prior families rather than from a failure to model the underlying distribution accurately. For example, a one-component GMM with a diagonal covariance matrix can represent the same distribution as a Copula with Gaussian marginals and no dependence. Consistent with this hypothesis, $75\%$ of incorrect prior-family predictions occur on one-dimensional datasets, where different prior families admit greater overlap.

\begin{figure}
    \centering
    \includegraphics[width=0.69\linewidth]{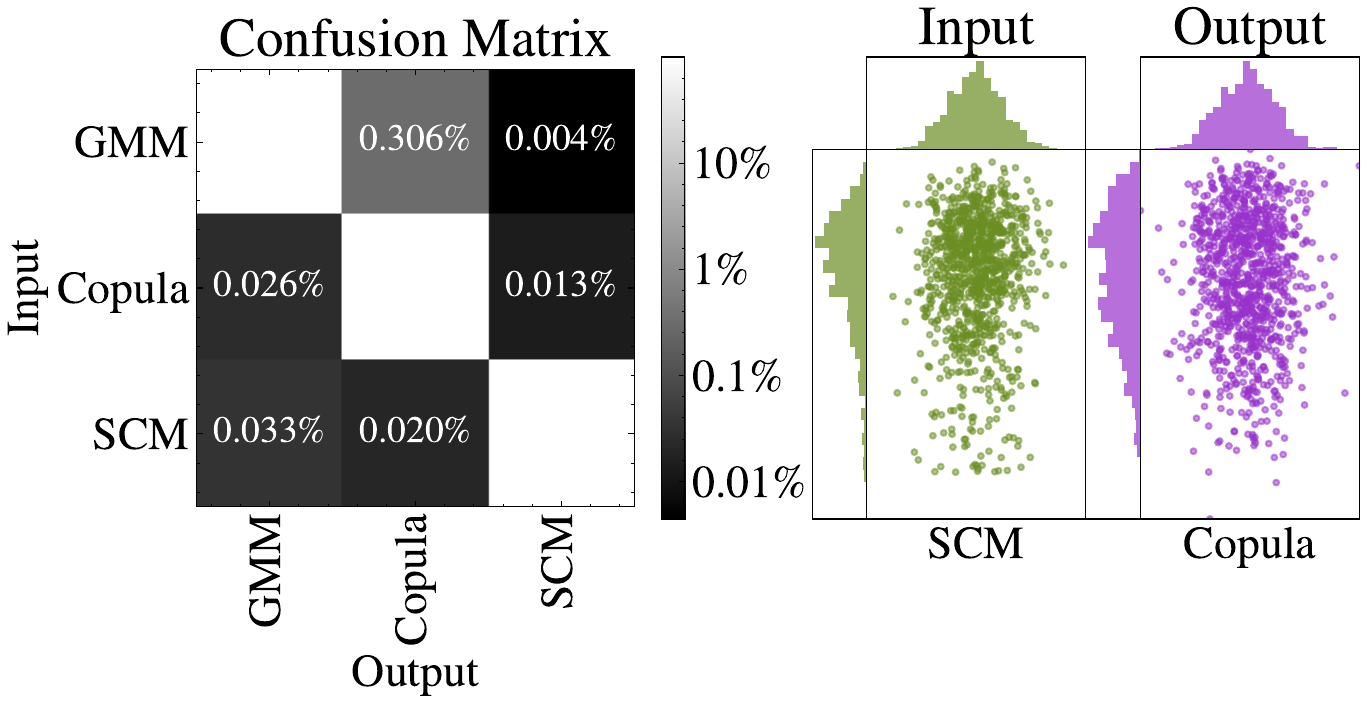}
    \vspace{-0.1in}
    \caption{Predicting the family of the program to be reconstructed is a learned task, \method may occasionally misclassify the data prior. Such confusion is rare, however, \textbf{with only $1/744$ programs misclassified} (left). Despite the misclassification, the \textbf{model continues to fit the data well} (right), suggesting that these errors arise from overlapping data priors rather than priors that are inherently difficult for \method.}
    \label{fig:casestudy}
\end{figure}

\newpage

\subsection{Program reconstruction performance}\label{app:progrecon}

Our evaluation in Section \ref{sec:exp} primarily covers downstream performance in generation, density estimation, and score estimation. In contrast, Figure \ref{fig:quality} evaluates \method's ability to reconstruct individual program parameters.

Overall, GMM and Copula reconstruction is relatively accurate, with floating-point MAEs $\ll 1$ and categorical accuracies $\geq 75\%$. The lowest categorical accuracy is observed for the predicted number of GMM components ($75\%$). This highlights a fundamental ambiguity in program reconstruction: multiple programs can generate very similar distributions (see Figure \ref{fig:casestudy}). For example, two strongly overlapping Gaussian components can be nearly indistinguishable from a single component with a larger mixture weight.

This ambiguity is particularly pronounced for SCMs. The maximum in-degree of the causal graph is recovered with $52\%$ accuracy ($20\%$ random chance) and an MAE of $0.58$. Parent indices are predicted with $68\%$ accuracy, while structural factors have an MAE of $1.39$. Small differences in graph structure or functional parameters may nevertheless result in similar observational distributions, particularly when the mechanisms compensate for one another.

\textbf{SCM graph recovery.}
To quantify structural reconstruction directly, we additionally compare predicted and ground-truth graphs on 32 synthetic SCM datasets, with an average of 36.1 ground-truth edges. As a reference, we construct random graphs that preserve the ground-truth number of parents for each node but randomly select their identities. Table \ref{tab:scm_recovery} summarizes the results.

\begin{table}[t]
    \centering
    \caption{SCM graph reconstruction on 32 synthetic datasets.
    The random baseline preserves each node's ground-truth in-degree.}
    \label{tab:scm_recovery}
    \begin{tabular}{lcc}
        \toprule
        \textbf{Metric} & \textbf{\method} & \textbf{Random graph} \\
        \midrule
        Directed edge F1 ($\uparrow$) & $\mathbf{0.59 \pm 0.05}$ & 0.39 \\
        Precision ($\uparrow$) & $\mathbf{0.67 \pm 0.04}$ & 0.39 \\
        Recall ($\uparrow$) & $\mathbf{0.54 \pm 0.05}$ & 0.39 \\
        Skeleton F1 ($\uparrow$) & $\mathbf{0.63 \pm 0.05}$ & 0.44 \\
        SHD ($\downarrow$) & $\mathbf{32.3 \pm 6.6}$ & 61.5 \\
        \bottomrule
    \end{tabular}
\end{table}

Despite the ambiguity of observational graph recovery, \method substantially outperforms the random baseline, achieving a directed-edge F1 of $0.59$ versus $0.39$ and reducing structural Hamming distance (SHD) by $47\%$. Notably, the baseline receives the ground-truth in-degrees, whereas \method must infer the graph structure from the data.

Importantly, our primary objective is not exact causal structure recovery, but inference of an executable program that accurately reproduces the observed distribution. Different SCMs can induce similar observational distributions, so exact graph recovery is neither guaranteed nor necessary for our downstream tasks. Nevertheless, these results demonstrate that \method learns meaningful structural information beyond distribution matching alone.

\begin{figure}[t]
    \centering
    \includegraphics[width=0.89\linewidth]{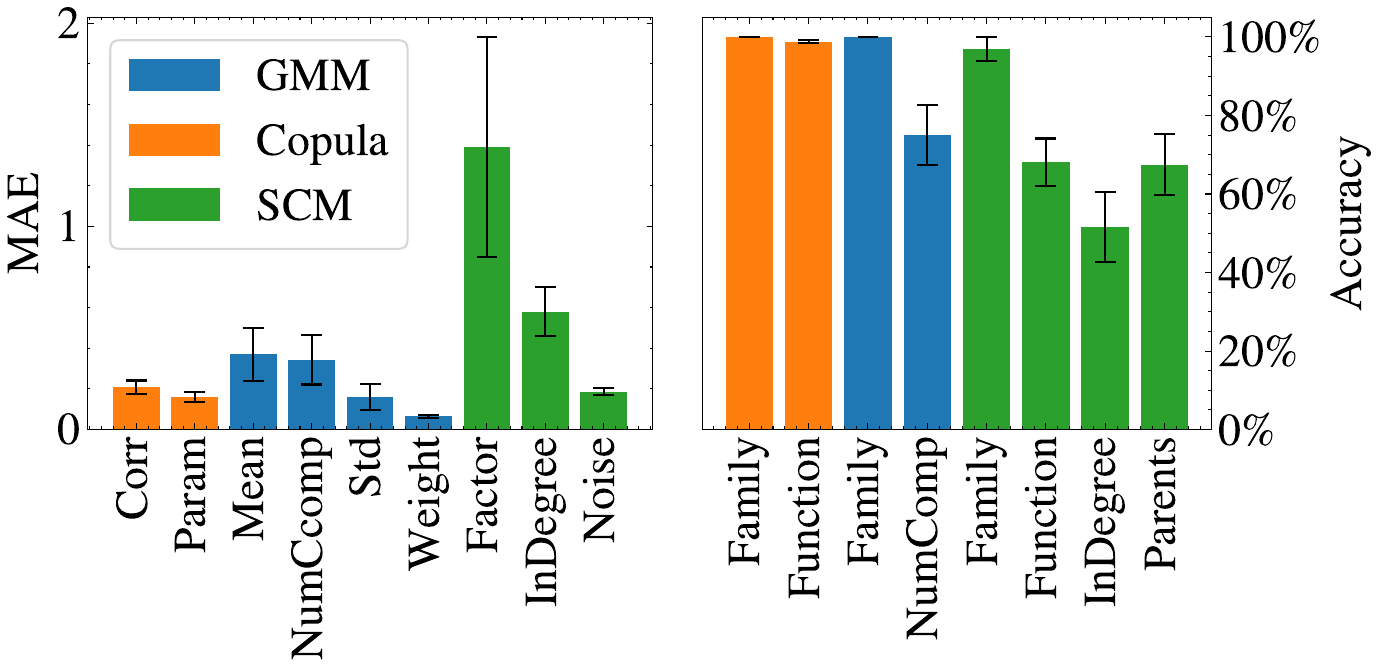}
    \vspace{-0.1in}
    \caption{Program reconstruction performance of \method.
    Left: MAE for floating-point parameters.
    Right: Accuracy for categorical and integer parameters.
    Results are averaged over applicable datasets and parameters
    within each prior family.}
    \label{fig:quality}
\end{figure}

\newpage
\subsection{Real-World Results and Examples}

We evaluate \method and baselines on $96$ datasets extracted from OddBench \cite{ding2026macrodata}. We ignore labeled anomalies and choose a random subset of features for each dataset to guarantee an equivalent distribution of feature dimensions compared to our in-distribution datasets.
A list of all used datasets can be found in Table \ref{tab:rw_sources} and in our supplementary material.

We further show examples of the data generation of \method and baselines in Figure on Real-World datasets in Figure \ref{fig:rw_examples}.

To obtain approximate ground-truth densities and scores, we fit a Gaussian mixture model (GMM) as described in Appdx. \ref{app:baseline:gmm}, using a maximum of $100$ components per dataset. We retain only datasets for which we fail to reject the null hypothesis when evaluating newly generated samples. This procedure yields $37$ datasets with GMM-approximated density and score ground truth.

\begin{table}[h]
\centering
\caption{OddBench datasets used as Real-World Datasets  }
\begin{tabular}{lll}
\hline
$\text{Dim}=1$ & $\text{Dim}=2$ & $\text{Dim}=5$ \\
\hline
DifferentialExpression & AmusementParkSafety & StudentProgrammePerformance \\
TrafficIncident & MunicipalExpenditure & AbandonedMineLand \\
HeartMurmurDetection & PlanetaryEmissions & AutoencoderAnomaly \\
AgilePriority & RxJavaPackage & RestaurantViolation \\
SpeciesValidation & FinancialBeneficiary & MaritimeVessel \\
VotingRecall & ServiceErrorLocation & BusStopPrior \\
CallPerformance & FraudProcessSteps & KidneyDisease \\
GameItemPassive & FarmerPayments & SoftwareDevelopmentEffort \\
FishingAnomalies & FinancialTransactionAnomalies & WaterQualityAssessment \\
TextMessageSpam & SystemMessageAnomalies & LabTestAbnormality \\
NetworkTrafficDOS & RealEstateBlockType & HousingConstruction \\
NetworkFirewall & LanguageLocationAnomaly & NetworkActivity \\
\hline
$\text{Dim}=10$ & $\text{Dim}=20$ & $\text{Dim}=30$ \\
\hline
WindmillAnomaly & WaterSources & TravelWeatherScores \\
QuadrotorNavigation & EnvironmentalFlux & EconomicLifeExpectancy \\
CodeCommitMetrics & IndustrialSensor & BaseballPitchingAnomalies \\
BuildingDamage & EyeTrackingAnomalies & ProsperLoan \\
BiologicalClassifications & CreditRiskAssessment & FootballPlayerPosition \\
ChicagoFoodInspection & MineAccident & ChemicalReactionAnalysis \\
TreeUsage & SoccerActions & EnvironmentalAnomalies \\
KickstarterProjects & MoldProcess & SatelliteAnomaly \\
BarExamPrediction & ICUPhysiology & WeatherUVIndex \\
BudgetGoal & BugSeverity & WorkforceDemographics \\
BirdAudioAnomaly & BuildingInspection & VolleyballMatch \\
RealEstateIrregularity & CorporateBondDefault & LondonCrimeFraud \\
\hline
$\text{Dim}=40$ & $\text{Dim}=50$ &  \\
\hline
MiningProjectOutcome & CreativeSchoolCertification &  \\
GamePositionAnomaly & SystemHealth &  \\
BoliviaUnpaidLabor & GalaxyAnomaly &  \\
SentencingFraud & GasBenchmark &  \\
AdRequestFraud & EmailSpamDetection &  \\
RNASequencingAnomaly & ElectrocaloricAnomalies &  \\
CodeQualityMetrics & CancerGeneExpression &  \\
DataValidationEntries & FinanceJobCategories &  \\
RNAAmplificationEphys & SoftwareQualityMetrics &  \\
CodeDefects & BugReportNgramIDF &  \\
SlowlorisAttackDetection & VehicleNetworkTraffic &  \\
MassBalanceCheck & MachineFailureSensors &  \\
\noalign{\smallskip}\cline{1-2}\noalign{\smallskip}
\end{tabular}
\label{tab:rw_sources}
\end{table}

\begin{figure}
    \centering
    \includegraphics[width=0.89\linewidth]{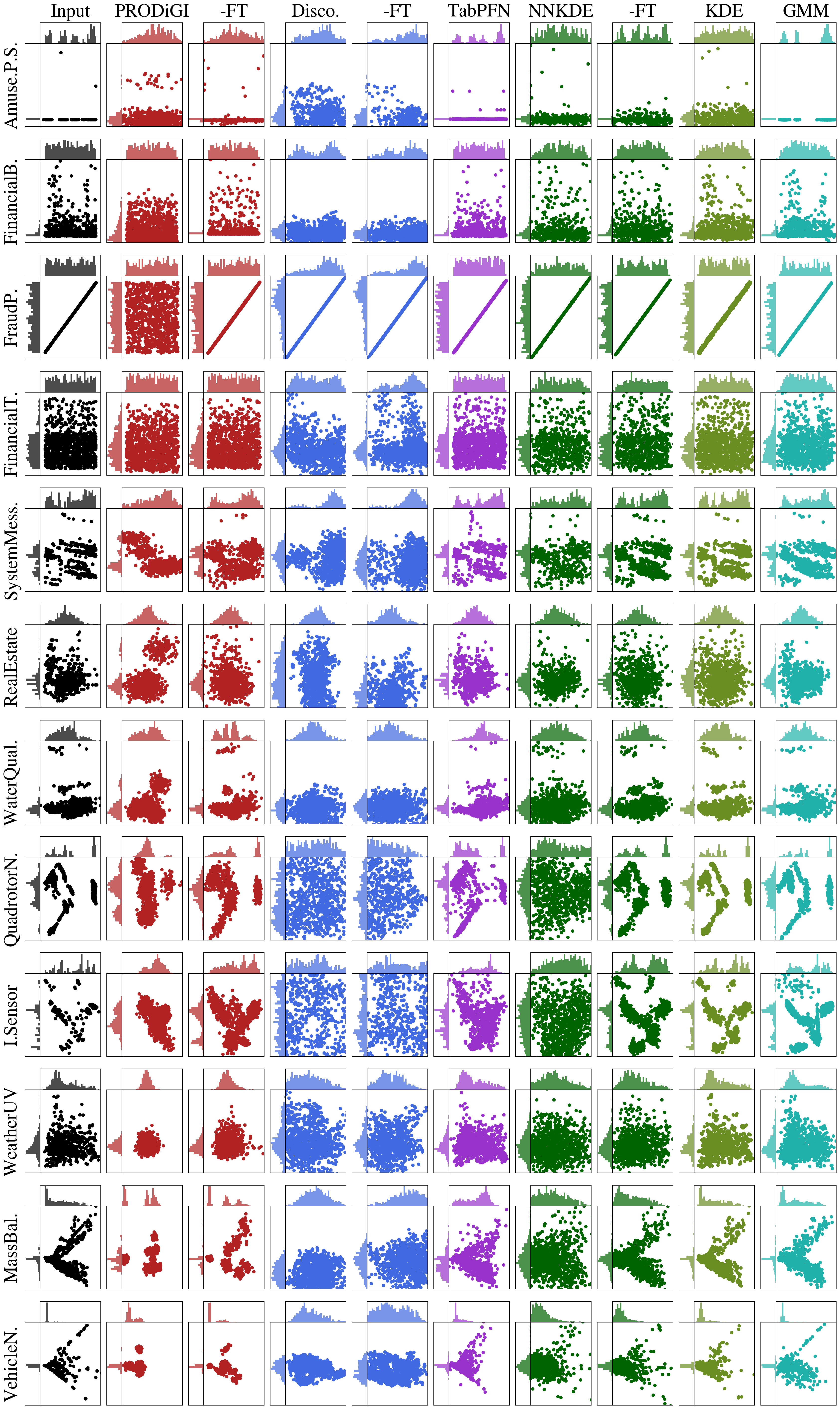}
    \vspace{-0.1in}
    \caption{Example real-world datasets and generations by all methods, including fine-tuned variants. The first six examples are two-dimensional, while the last six are PCA representations of one 5-, 10-, 20-, 30-, 40-, and 50-dimensional dataset each. \textbf{Real-world manifolds challenge all methods, where \methodft notably boosts \method.}}
    \label{fig:rw_examples}
\end{figure}



\begin{figure}[h]
\vspace{-0.1in}
    \centering
\includegraphics[width=1.0\linewidth]{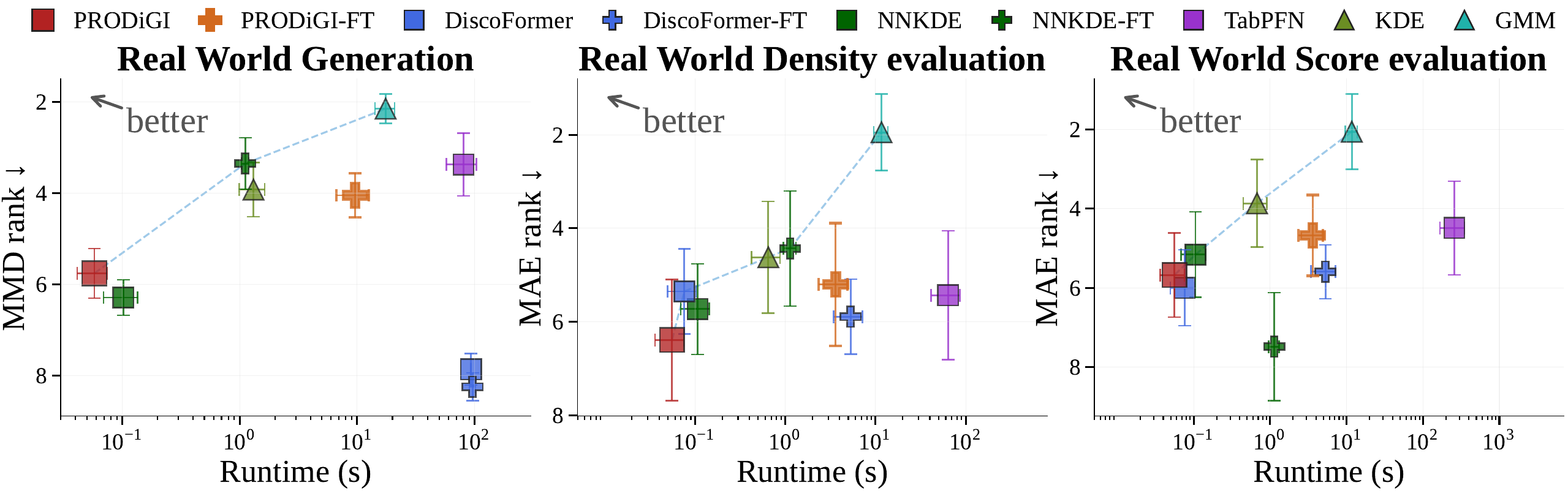}
\vspace{-0.2in}
    \caption{Performance–inference time trade-off for Generation, Density, and Score Estimation on real-world datasets. \method ranks among the fastest methods across all tasks, while \methodft consistently improves performance on each task. However, the limited coverage of prior families constrains performance on datasets that cannot be adequately described by the available priors.
}
    \label{fig:rank_comparison_rw}
\end{figure}

\begin{table}[h]
    \centering
      \caption{Performance comparison across tasks and metrics on \textbf{real-world datasets}$^\text{a}$: (1) \textbf{Generation distribution match} via MMD ($\downarrow$) where $\% p$$<$$0.05$ ($\downarrow$) denotes the proportion of datasets where the null hypothesis ($H_0$: generated and real distributions are identical) is rejected;
    (2) \textbf{Density estimation} via point-wise MAE ($\downarrow$), population-wide Spearman correlation ($\uparrow$), and NLL of predicted densities ($\downarrow$); 
    (3) \textbf{Score estimation} via MAE ($\downarrow$) and cosine similarity ($\uparrow$). Average rank ($\downarrow$) per metric is across methods over all datasets. Time ($\downarrow$) per task (s) is averaged over datasets.}  
    \label{tab:all_metrics_rw}
        \vspace{-0.1in}

    \setlength{\tabcolsep}{1.5pt} 
    \renewcommand{\arraystretch}{1.05}

    \resizebox{\columnwidth}{!}{%
    \begin{threeparttable}

\begin{tabular}{c@{\hspace{2pt}}l|@{\hspace{2pt}}*{2}{c}|@{\hspace{2pt}}*{7}{c}}
        \toprule
        & \textbf{Metric}  &\textbf{PRODiGI} & \textbf{-FT} & \textbf{DiscoFormer} & \textbf{-FT} & \textbf{NNKDE} & \textbf{-FT} & \textbf{TabPFN} & \textbf{KDE} & \textbf{GMM} \\
        \midrule
\multirow{5}{*}{\rotatebox[origin=c]{90}{\textbf{Generation}}}
                & MMD ($\downarrow$) {\tiny$\times 10^3$} & \pmv{68.53}{5.71} & \pmv{37.29}{4.58} & \pmv{251.91}{22.73} & \pmv{264.23}{21.95} & \pmv{79.31}{8.96} & \pmv{\underline{31.07}}{5.96} & \pmv{45.23}{10.53} & \pmv{67.91}{12.78} & \pmv{\textbf{16.25}}{3.21} \\

                & Rank ($\downarrow$) & \pmv{5.76}{0.18} & \pmv{4.05}{0.16} & \pmv{7.86}{0.12} & \pmv{8.25}{0.10} & \pmv{6.29}{0.13} & \pmv{\underline{3.35}}{0.19} & \pmv{3.37}{0.23} & \pmv{3.92}{0.20} & \pmv{\textbf{2.15}}{0.11} \\

        \noalign{\smallskip}\cline{2-11}\noalign{\smallskip}
                & \% $p$$<$$0.05$ ($\downarrow$) & 97.9\% & 89.6\% & 100\% & 100\% & 97.9\% & 82.3\% & 83.3\% & \underline{79.2\%} & \textbf{77.1\%} \\

        \noalign{\smallskip}\cline{2-11}\noalign{\smallskip}
                & Time ($\downarrow$) & \pmv{\textbf{0.06}}{0.01} & \pmv{9.64}{1.00} & \pmv{93.44}{0.11} & \pmv{96.01}{1.81} & \pmv{\underline{0.10}}{0.01} & \pmv{1.12}{0.03} & \pmv{80.53}{7.66} & \pmv{1.31}{0.11} & \pmv{17.48}{1.11} \\

                & Rank ($\downarrow$) & \pmv{\textbf{1.12}}{0.03} & \pmv{5.06}{0.04} & \pmv{7.64}{0.05} & \pmv{8.62}{0.05} & \pmv{\underline{1.91}}{0.04} & \pmv{3.58}{0.06} & \pmv{7.48}{0.12} & \pmv{3.42}{0.06} & \pmv{6.17}{0.06} \\

        \midrule
        \midrule
\multirow{8}{*}{\rotatebox[origin=c]{90}{\textbf{Density}}}
                & MAE ($\downarrow$) & \pmv{332}{226} & \pmv{\underline{332}}{226} & \pmv{360}{228} & \pmv{360}{228} & \pmv{360}{228} & \pmv{340}{226} & \pmv{362}{228} & \pmv{362}{227} & \pmv{\textbf{326}}{202} \\

                & Rank ($\downarrow$) & \pmv{6.39}{0.43} & \pmv{5.20}{0.44} & \pmv{5.35}{0.30} & \pmv{5.89}{0.27} & \pmv{5.73}{0.32} & \pmv{\underline{4.43}}{0.41} & \pmv{5.43}{0.46} & \pmv{4.62}{0.40} & \pmv{\textbf{1.95}}{0.27} \\

        \noalign{\smallskip}\cline{2-11}\noalign{\smallskip}
                & Spear. ($\uparrow$) & \pmv{0.30}{0.04} & \pmv{0.43}{0.05} & \pmv{0.45}{0.04} & \pmv{0.48}{0.05} & \pmv{0.51}{0.05} & \pmv{\underline{0.57}}{0.05} & \pmv{0.49}{0.05} & \pmv{0.55}{0.05} & \pmv{\textbf{0.67}}{0.05} \\

                & Rank ($\downarrow$) & \pmv{7.31}{0.36} & \pmv{5.76}{0.41} & \pmv{5.83}{0.29} & \pmv{5.20}{0.39} & \pmv{4.77}{0.27} & \pmv{4.66}{0.38} & \pmv{5.16}{0.44} & \pmv{\underline{3.97}}{0.39} & \pmv{\textbf{2.34}}{0.33} \\

        \noalign{\smallskip}\cline{2-11}\noalign{\smallskip}
                & NLL ($\downarrow$) & \pmv{11.68}{2.73} & \pmv{\underline{8.21}}{2.12} & -\tnote{b} & -\tnote{b} & \pmv{\textbf{1.21}}{1.17} & \pmv{46.56}{32.07} & -\tnote{b} & \pmv{9.11}{2.97} & \pmv{34.30}{27.25} \\

                & Rank ($\downarrow$) & \pmv{4.53}{0.28} & \pmv{3.64}{0.27} & -\tnote{b} & -\tnote{b} & \pmv{\textbf{3.00}}{0.26} & \pmv{\underline{3.05}}{0.29} & -\tnote{b} & \pmv{3.38}{0.22} & \pmv{3.41}{0.28} \\

        \noalign{\smallskip}\cline{2-11}\noalign{\smallskip}
                & Time ($\downarrow$) & \pmv{\textbf{0.06}}{0.01} & \pmv{3.62}{0.73} & \pmv{\underline{0.08}}{0.04} & \pmv{5.29}{4.59} & \pmv{0.11}{0.02} & \pmv{1.13}{0.06} & \pmv{63.57}{16.37} & \pmv{0.65}{0.08} & \pmv{11.65}{0.70} \\

                & Rank ($\downarrow$) & \pmv{\underline{1.97}}{0.09} & \pmv{7.14}{0.10} & \pmv{\textbf{1.46}}{0.21} & \pmv{4.70}{0.16} & \pmv{2.78}{0.08} & \pmv{6.03}{0.14} & \pmv{7.92}{0.19} & \pmv{4.51}{0.14} & \pmv{8.49}{0.11} \\

        \midrule
        \midrule
\multirow{6}{*}{\rotatebox[origin=c]{90}{\textbf{Score}}}
                & MAE ($\downarrow$) & \pmv{5810}{3018} & \pmv{5831}{3017} & \pmv{6039}{3008} & \pmv{5957}{3011} & \pmv{6134}{3008} & \pmv{11164}{4206} & \pmv{5799}{3017} & \pmv{\underline{5790}}{3014} & \pmv{\textbf{5076}}{2446} \\

                & Rank ($\downarrow$) & \pmv{5.68}{0.36} & \pmv{4.68}{0.34} & \pmv{6.00}{0.32} & \pmv{5.59}{0.23} & \pmv{5.16}{0.36} & \pmv{7.49}{0.45} & \pmv{4.49}{0.39} & \pmv{\underline{3.86}}{0.37} & \pmv{\textbf{2.05}}{0.32} \\

        \noalign{\smallskip}\cline{2-11}\noalign{\smallskip}
                & Cos. ($\uparrow$) & \pmv{0.06}{0.02} & \pmv{0.16}{0.03} & \pmv{0.18}{0.03} & \pmv{0.16}{0.03} & \pmv{0.22}{0.04} & \pmv{0.32}{0.05} & \pmv{0.23}{0.04} & \pmv{\underline{0.33}}{0.05} & \pmv{\textbf{0.52}}{0.06} \\

                & Rank ($\downarrow$) & \pmv{7.01}{0.33} & \pmv{5.77}{0.31} & \pmv{5.68}{0.31} & \pmv{5.85}{0.37} & \pmv{5.05}{0.35} & \pmv{4.59}{0.43} & \pmv{5.05}{0.43} & \pmv{\underline{3.61}}{0.37} & \pmv{\textbf{2.38}}{0.32} \\

        \noalign{\smallskip}\cline{2-11}\noalign{\smallskip}
                & Time ($\downarrow$) & \pmv{\textbf{0.06}}{0.01} & \pmv{3.62}{0.73} & \pmv{\underline{0.08}}{0.04} & \pmv{5.29}{4.59} & \pmv{0.11}{0.02} & \pmv{1.13}{0.06} & \pmv{257.74}{94.76} & \pmv{0.67}{0.08} & \pmv{11.71}{0.71} \\

                & Rank ($\downarrow$) & \pmv{\underline{1.97}}{0.09} & \pmv{7.03}{0.10} & \pmv{\textbf{1.43}}{0.19} & \pmv{4.65}{0.15} & \pmv{2.78}{0.08} & \pmv{5.84}{0.12} & \pmv{8.24}{0.12} & \pmv{4.57}{0.14} & \pmv{8.49}{0.11} \\

        \bottomrule
    \end{tabular}%

\begin{tablenotes}
\item[a] Detailed results on individual priors and dimensions are given in Appdx. \ref{app:detailed_tables}. 
\item[b] NLL is not reported for DiscoFormer and TabPFN because their local density estimates are not normalized to integrate to one.\end{tablenotes}
\end{threeparttable}

    }
\end{table}

We quantitatively evaluate our methods on these datasets in Figure \ref{fig:rank_comparison_rw} and Table \ref{tab:all_metrics_rw}.

Overall, the MMD values are approximately an order of magnitude higher than those obtained on the prior families (Table \ref{tab:all_metrics_priors}), indicating that real-world datasets are substantially more challenging for all methods. This is also reflected in the rejection rate: for datasets with $30$ or more dimensions, no method generates samples sufficiently similar to the input data to fail our $p$-test.

Despite this increased difficulty, \method remains competitive. It generates samples in the shortest amount of time, while \methodft achieves the third-lowest average MMD.

Performance on the density and score estimation tasks similarly degrades across methods. Since the ground truth in this setting is derived from a GMM, the GMM baseline has a natural advantage. Nevertheless, its performance is substantially lower than on the prior families, with the Spearman correlation decreasing from $0.82$ to $0.67$ and the average cosine similarity decreasing from $0.59$ to $0.52$.

For \method, we retain Copula and SCM\footnote{Out of $37$ datasets with a GMM-derived ground truth approximation, \method classifies $20$ as a GMM and $17$ as a Copula, but none as a SCM.} templates despite expecting them to provide a poor fit to the GMM-derived ground truth. Consequently, \method generally underperforms baselines that either explicitly model the data using GMMs or are trained exclusively on GMM-generated data. Nevertheless, \methodft remains competitive, achieving, for example, the second-lowest density MAE and NLL. 

We further report the distribution of the predicted prior families in Figure \ref{fig:families}. While the prior families are not equally prevalent across real-world datasets, each captures characteristics observed in real-world data.

Both of these results suggest that expanding the set of supported prior families could further improve performance on real-world datasets.

\begin{figure}[h!]
    \centering
    \includegraphics[width=0.49\linewidth]{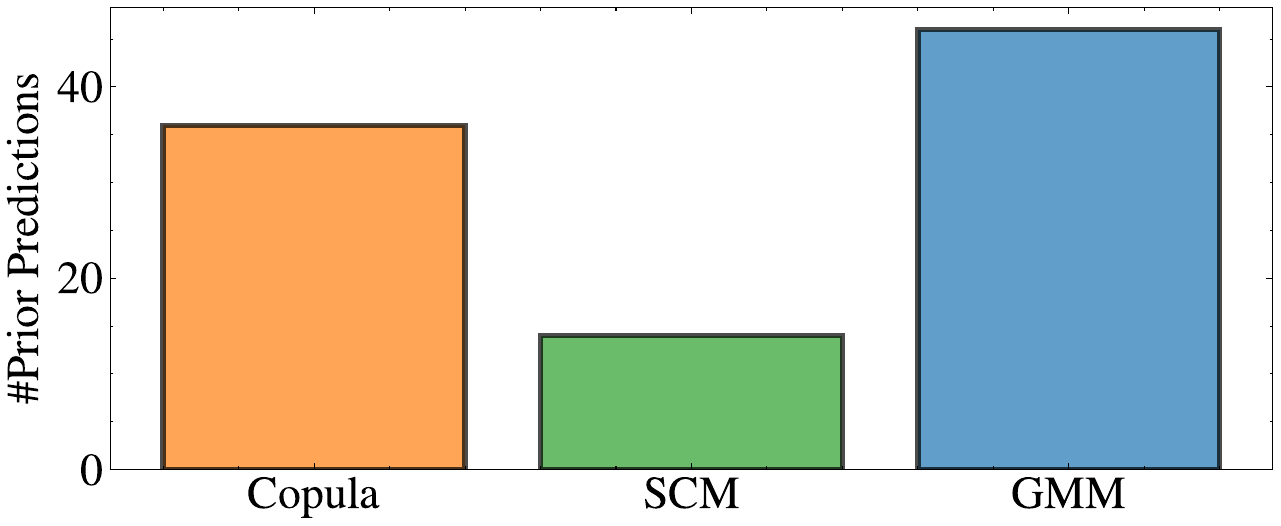}
    \vspace{-0.1in}
    \caption{Distribution of predicted families on Real-World examples. Each Family is predicted between 14 and 46 times, showing that \textbf{each of our priors helps to model Real-World data more accurately}.}
        \label{fig:families}
\end{figure}

\newpage
\subsection{Finetuning without \method}\label{app:ftnoprodigi}

To validate whether finetuning requires \method or whether we can achieve comparable results with random initialization, we compare finetuning random programs to \method predictions in Figures \ref{fig:random_init} (in-distribution) and \ref{fig:random_init_rw} (real-world data). While sometimes random initializations do help reach a better performance, overall in each setting it is better to use \method to initialize the finetuning process.

\begin{figure}
    \centering
    \includegraphics[width=1\linewidth]{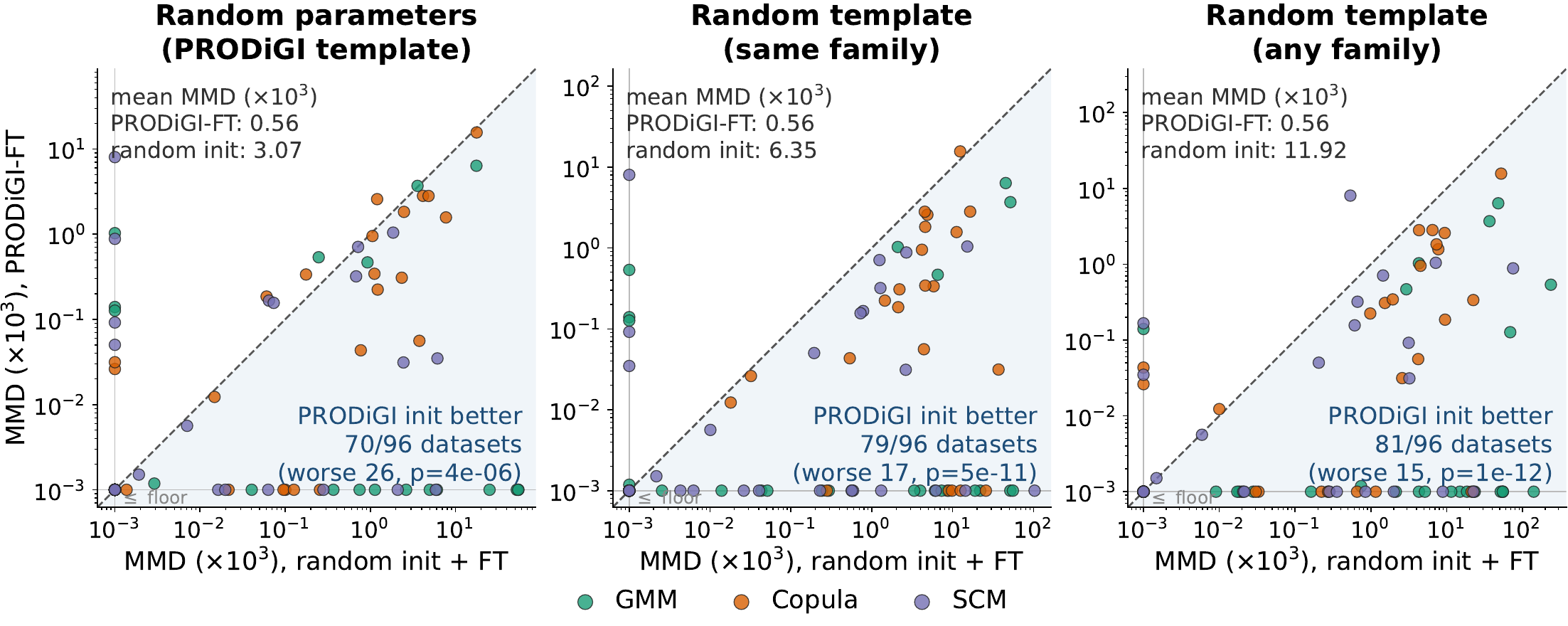}
    \vspace{-0.1in}
    \caption{Performance of our fine-tuning routine on our in-distribution generation tasks, when starting with random programs instead of \method output. From left to right, we replace a \method prediction with a random program of the same template, with a random program of the same prior family and with an entirely random program. \textbf{Starting Fine-Tuning with \method is a significantly better choice in each of these settings.}}
    \label{fig:random_init}
\end{figure}
\begin{figure}
    \centering
    \includegraphics[width=1\linewidth]{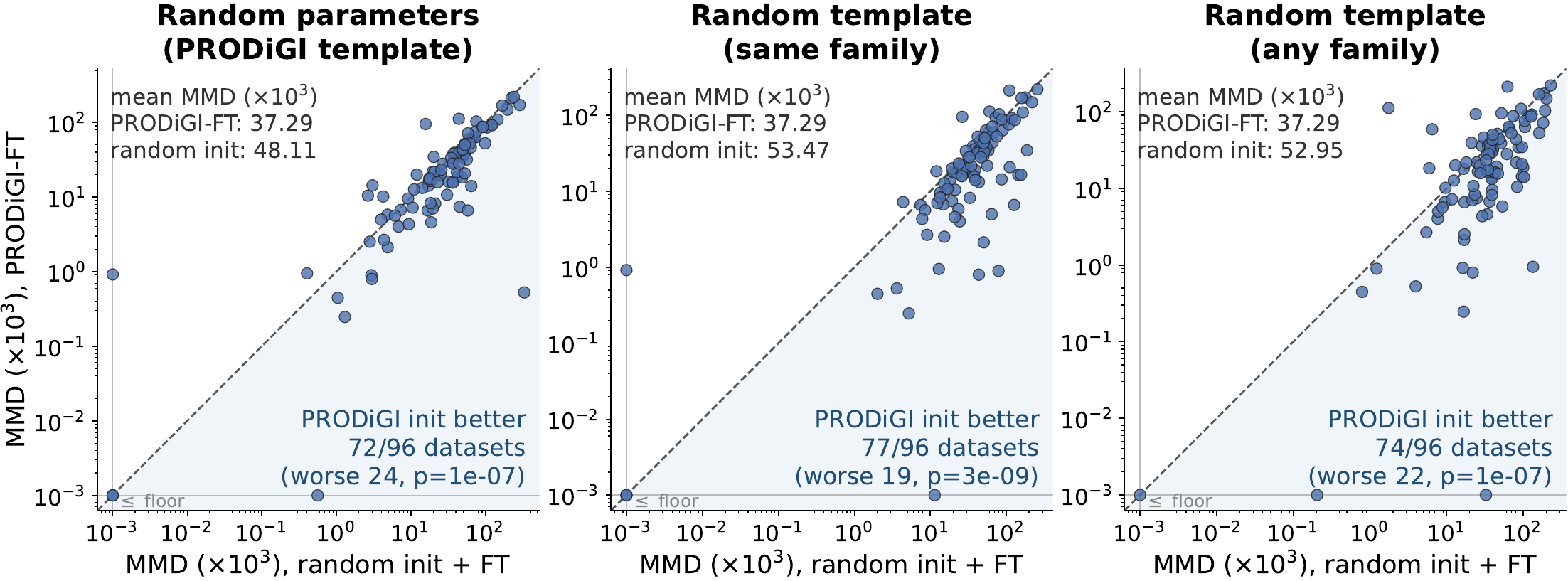}
    \vspace{-0.1in}
    \caption{Performance of our fine-tuning routine on our real-world generation tasks, when starting with random programs instead of \method output. From left to right, we replace a \method prediction with a random program of the same template, with a random program of the same prior family and with an entirely random program. \textbf{Starting Fine-Tuning with \method is a significantly better choice in each of these settings.}}
    \label{fig:random_init_rw}
\end{figure}


\newpage
\subsection{Out-of-Distribution Results}\label{app:ood}

\method and \methodft perform worse on real-world datasets when compared to their excellent in-distribution performance. To further analyze this behavior, we compare their performance on simulated but out-of-distribution datasets. For this, we generate $96$ new datasets following our priors, but with critical parameters not seen during training, forcing \method and \methodft to approximate the new parameter settings with those seen during training. Our $96$ datasets consist of $32$ datasets per modified prior, $12$ datasets per studied dimension, closely following the analysis for In-Distribution results. The priors are modified in the following ways:

\begin{itemize}
    \item \textbf{GMM:} We increase the number of components from $1-10$ to $11-20$. Each component is still guaranteed to have a weight $>1\%$, but the differences between component weights are unrestricted.
    \item \textbf{SCM:} We replace the three activation functions shown in Figure \ref{fig:app_scmfunc} with the five activation functions shown in Figure \ref{fig:app_scmfunc_ood}.
    \item \textbf{Copula:} We replace the five marginals shown in Figure \ref{fig:app_marginals} with the five marginals shown in Figure \ref{fig:app_marginals_ood}.
\end{itemize}

The results are shown in Table \ref{tab:all_metrics_ood} and are compared to our main In-Distribution results in Figure \ref{fig:ood_diff}. Almost all our competitors show only very limited performance decreases compared to the In-Distribution case (e.g., the MMD of GMM increases from $0.78$ to $1.35$ and the cosine similarity of KDE even slightly improves from $0.47$ to $0.58$). \method is more significantly hurt by this change (e.g., the Spearman correlation decreases from $0.72/0.90$ to $0.56/0.71$ for the raw/finetuned versions). This is expected, since due to the reliance on explicit programs, \method and \methodft are incapable of representing the new OOD distributions exactly. However, the performance of \methodft remains quite competitive, with only GMM and TabPFN outperforming our method on most metrics, though at significantly higher runtime. This indicates that an improved version of \method with more diverse priors would not have the same approximation limitation and would likely perform even better on real-world datasets.

\begin{figure}
    \centering
    \includegraphics[width=1.0\linewidth]{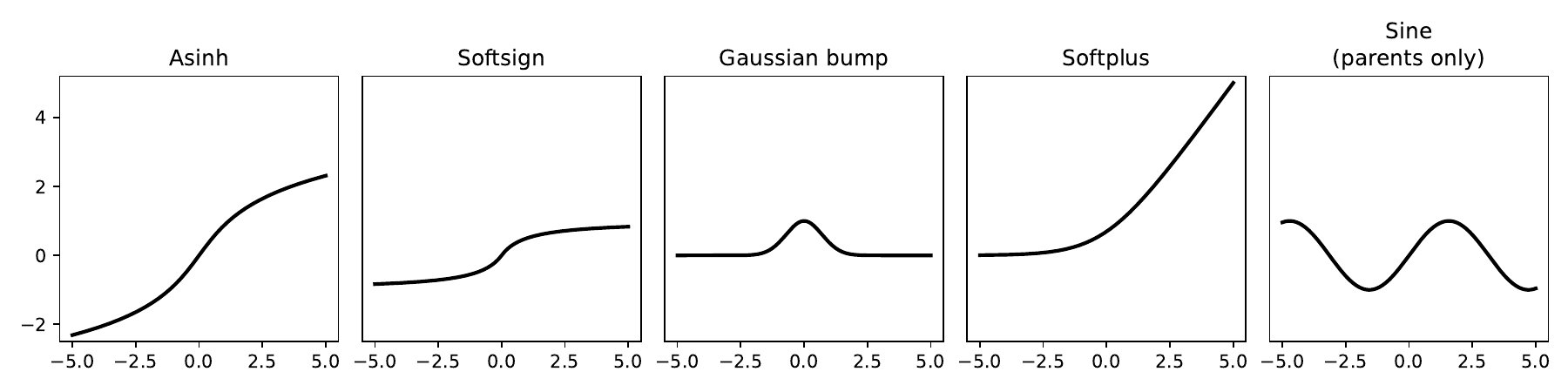}
    \vspace{-0.1in}
    \caption{SCM activation functions used for the OOD analysis.}
    \label{fig:app_scmfunc_ood}
\end{figure}

\begin{figure}
    \centering
    \includegraphics[width=1.0\linewidth]{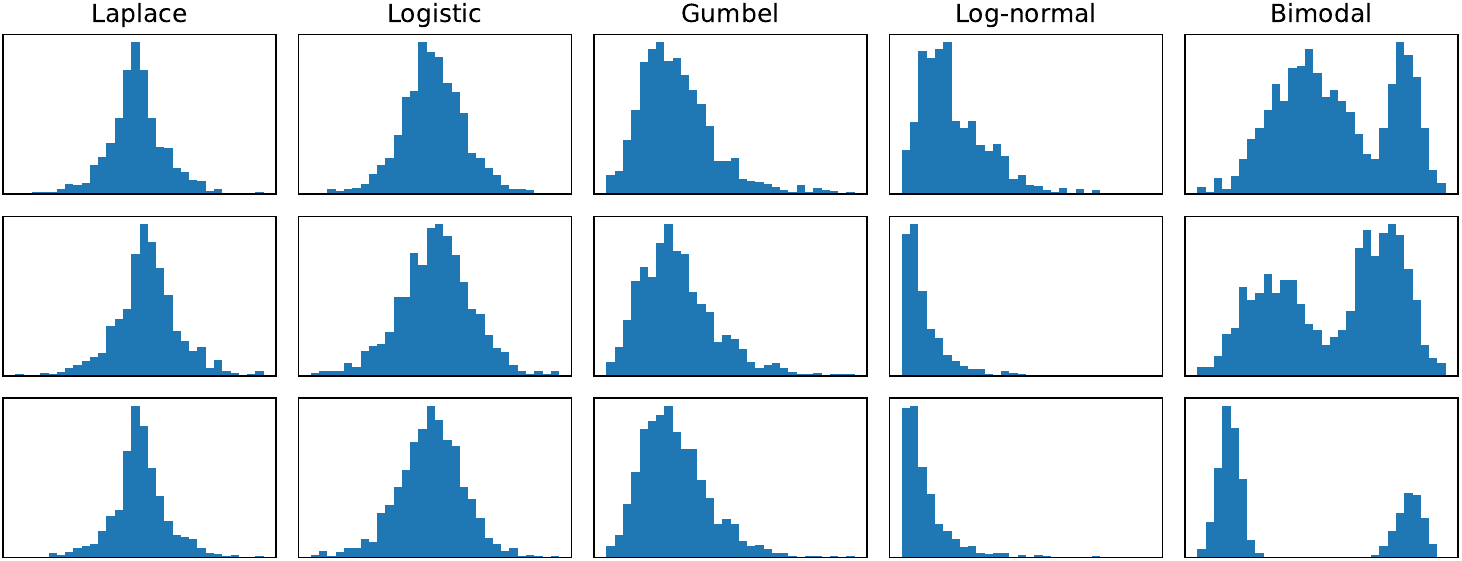}
    \vspace{-0.1in}
    \caption{Examples of each copula marginal used for the OOD analysis.}
    \label{fig:app_marginals_ood}
\end{figure}

\begin{table}[!th]
\vspace{-0.1in}
    \centering
    \caption{Performance comparison across tasks and metrics over all prior families for out-of-distribution datasets: (1) \textbf{Generation distribution match} via MMD ($\downarrow$) where $\% p$$<$$0.05$ ($\downarrow$) denotes the proportion of datasets where the null hypothesis ($H_0$: generated and real distributions are identical) is rejected; 
    (2) \textbf{Density estimation} via point-wise MAE ($\downarrow$), population-wide Spearman correlation ($\uparrow$), and NLL of predicted densities ($\downarrow$); 
    (3) \textbf{Score estimation} via MAE ($\downarrow$) and cosine similarity ($\uparrow$). Average rank ($\downarrow$) per metric is across methods over all datasets.}
    \label{tab:all_metrics_ood}
    \vspace{-0.1in}

    \setlength{\tabcolsep}{1.5pt} 
    \renewcommand{\arraystretch}{1.05}

    \resizebox{\columnwidth}{!}{%
    \begin{threeparttable}

\begin{tabular}{c@{\hspace{2pt}}l|@{\hspace{2pt}}*{2}{c}|@{\hspace{2pt}}*{7}{c}}
        \toprule
        & \textbf{Metric}  &\textbf{PRODiGI} & \textbf{-FT} & \textbf{DiscoFormer} & \textbf{-FT} & \textbf{NNKDE} & \textbf{-FT} & \textbf{TabPFN} & \textbf{KDE} & \textbf{GMM} \\
        \midrule
\multirow{3}{*}{\rotatebox[origin=c]{90}{\textbf{Gen.}}}
                & MMD ($\downarrow$) {\tiny$\times 10^3$} & \pmv{8.70}{1.67} & \pmv{1.82}{0.45} & \pmv{36.01}{7.04} & \pmv{79.11}{21.09} & \pmv{15.89}{2.40} & \pmv{7.98}{1.38} & \pmv{\textbf{0.04}}{0.02} & \pmv{3.22}{0.74} & \pmv{\underline{1.35}}{0.56} \\

                & Rank ($\downarrow$) & \pmv{5.15}{0.18} & \pmv{3.31}{0.14} & \pmv{8.01}{0.10} & \pmv{8.03}{0.12} & \pmv{6.65}{0.14} & \pmv{5.59}{0.16} & \pmv{\textbf{1.99}}{0.08} & \pmv{3.89}{0.18} & \pmv{\underline{2.38}}{0.07} \\

        \noalign{\smallskip}\cline{2-11}\noalign{\smallskip}
                & \% $p$$<$$0.05$ ($\downarrow$) & 65.3\% & 34.7\% & 100\% & 100\% & 91.6\% & 75.8\% & \textbf{1.1\%} & 49.5\% & \underline{16.8\%} \\

        \midrule
        \midrule
\multirow{6}{*}{\rotatebox[origin=c]{90}{\textbf{Density}}}
                & MAE ($\downarrow$) & \pmv{16.95}{2.88} & \pmv{6.75}{0.98} & \pmv{6.99}{0.88} & \pmv{5.94}{0.81} & \pmv{7.25}{0.95} & \pmv{7.13}{0.96} & \pmv{\textbf{2.45}}{0.20} & \pmv{7.49}{1.14} & \pmv{\underline{4.88}}{0.77} \\

                & Rank ($\downarrow$) & \pmv{7.74}{0.18} & \pmv{4.71}{0.27} & \pmv{5.83}{0.20} & \pmv{5.08}{0.24} & \pmv{5.33}{0.15} & \pmv{5.10}{0.15} & \pmv{\underline{3.85}}{0.37} & \pmv{4.95}{0.31} & \pmv{\textbf{2.40}}{0.14} \\

        \noalign{\smallskip}\cline{2-11}\noalign{\smallskip}
                & Spear. ($\uparrow$) & \pmv{0.56}{0.03} & \pmv{0.71}{0.03} & \pmv{0.69}{0.02} & \pmv{0.72}{0.02} & \pmv{0.82}{0.01} & \pmv{0.80}{0.01} & \pmv{\textbf{0.86}}{0.01} & \pmv{0.82}{0.01} & \pmv{\underline{0.84}}{0.01} \\

                & Rank ($\downarrow$) & \pmv{7.87}{0.20} & \pmv{5.79}{0.28} & \pmv{7.03}{0.13} & \pmv{5.65}{0.21} & \pmv{3.91}{0.16} & \pmv{4.49}{0.19} & \pmv{\underline{3.25}}{0.27} & \pmv{4.15}{0.27} & \pmv{\textbf{2.85}}{0.19} \\

        \noalign{\smallskip}\cline{2-11}\noalign{\smallskip}
                & NLL ($\downarrow$) & \pmv{8.16}{3.61} & \pmv{\underline{-3.54}}{3.31} & -\tnote{a} & -\tnote{a} & \pmv{-2.65}{3.20} & \pmv{-2.83}{3.20} & -\tnote{a} & \pmv{-2.44}{3.29} & \pmv{\textbf{-5.41}}{3.28} \\

                & Rank ($\downarrow$) & \pmv{5.07}{0.13} & \pmv{\underline{3.12}}{0.17} & -\tnote{a} & -\tnote{a} & \pmv{4.11}{0.12} & \pmv{3.56}{0.10} & -\tnote{a} & \pmv{3.66}{0.18} & \pmv{\textbf{1.48}}{0.08} \\

        \midrule
        \midrule
\multirow{4}{*}{\rotatebox[origin=c]{90}{\textbf{Score}}}
                & MAE ($\downarrow$) & \pmv{148.30}{52.82} & \pmv{201.34}{63.32} & \pmv{140.11}{51.64} & \pmv{140.17}{51.51} & \pmv{142.74}{51.55} & \pmv{159.98}{51.57} & \pmv{\underline{138.11}}{48.97} & \pmv{142.08}{51.47} & \pmv{\textbf{138.03}}{51.38} \\

                & Rank ($\downarrow$) & \pmv{6.82}{0.27} & \pmv{4.95}{0.33} & \pmv{\underline{4.06}}{0.18} & \pmv{4.22}{0.21} & \pmv{5.18}{0.21} & \pmv{7.39}{0.17} & \pmv{4.83}{0.32} & \pmv{4.62}{0.20} & \pmv{\textbf{2.93}}{0.20} \\

        \noalign{\smallskip}\cline{2-11}\noalign{\smallskip}
                & Cos. ($\uparrow$) & \pmv{0.32}{0.02} & \pmv{0.58}{0.03} & \pmv{0.55}{0.03} & \pmv{0.56}{0.03} & \pmv{0.57}{0.02} & \pmv{0.40}{0.02} & \pmv{\underline{0.60}}{0.02} & \pmv{0.58}{0.03} & \pmv{\textbf{0.67}}{0.03} \\

                & Rank ($\downarrow$) & \pmv{7.80}{0.21} & \pmv{4.45}{0.30} & \pmv{4.78}{0.22} & \pmv{4.68}{0.24} & \pmv{4.96}{0.16} & \pmv{6.57}{0.18} & \pmv{\underline{4.21}}{0.30} & \pmv{4.82}{0.25} & \pmv{\textbf{2.71}}{0.20} \\

        \bottomrule
    \end{tabular}%

\begin{tablenotes}
\item[a] NLL is not reported for DiscoFormer and TabPFN because their local density estimates are not normalized to integrate to one.
\end{tablenotes}
\end{threeparttable}
    }
    \vspace{-0.225in}
\end{table}

\begin{figure}
    \centering
    \includegraphics[width=1\linewidth]{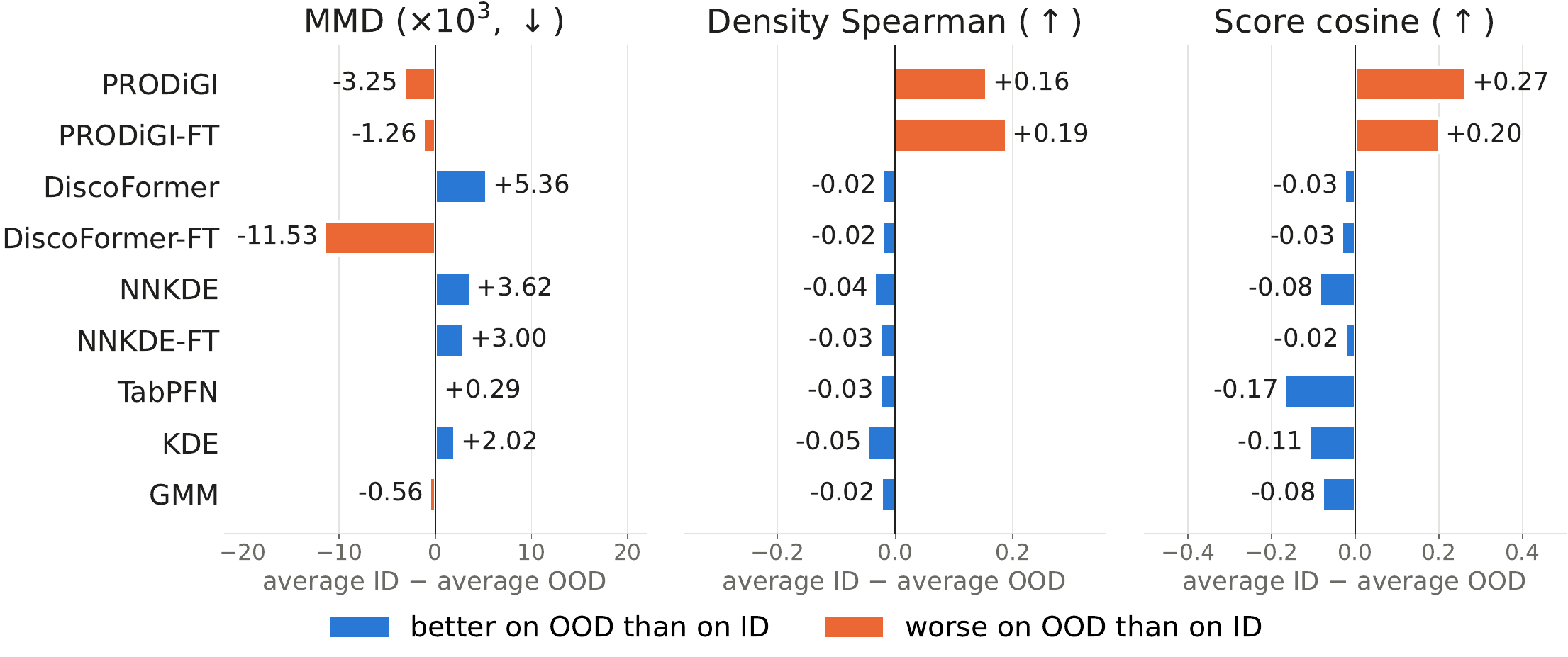}
    \vspace{-0.1in}
    \caption{Performance difference for different methods and metrics between In-Distribution and Out-Of-Distribution datasets. While the individual datasets are different in both runs, their overall statistics are similar. This is further shown, as \textbf{most methods achieve similar performance on both datasets, while \method and \methodft, which can only natively represent In-Distribution datasets, show the largest decrease in performance.}}
    \label{fig:ood_diff}
\end{figure}


\newpage
\subsection{LLM performance}\label{app:llmperf}

Large language models (LLMs) offer an alternative route to program inference: instead of predicting the parameters of a fixed template vocabulary, an LLM can write an arbitrary executable generative model. We compare PRODiGI against such a pipeline built on Claude Haiku~5.5.

\paragraph{Pipeline.}
The LLM receives the dataset name, the feature names (real-world data only), and the first 50 rows of the shuffled data rounded to three decimals. It is asked for a Python program that defines a class with \texttt{fit}, \texttt{sample}, and \texttt{log\_density} methods. The program may use any model and any fitting procedure built from \texttt{numpy}, \texttt{scipy}, \texttt{sklearn}, and \texttt{torch}. It runs on the CPU (4 threads, no network or file access) with a time limit of 300\,s. The prompt states that the program is evaluated by the mean log density of unseen rows. The 50 rows serve only to propose the program; \texttt{fit} receives the full training partition, so all available data is used for parameter estimation. Each program is fitted on 75\% of the training data and evaluated on the remaining 25\%. The LLM then receives the validation log likelihood, the five columns with the largest Kolmogorov--Smirnov distance between generated and held-out rows, and the five largest correlation errors. It may propose an improved program in up to two further rounds. If a reply cannot be used (no code, a crash, a timeout, non-finite outputs, or samples that copy the training rows), the error message is returned and up to two repairs are allowed per round. Among the valid programs, we select the one with the highest validation log likelihood, refit it on the full training partition, and evaluate it on the same test sets as all other methods. Scores are obtained by finite differences of \texttt{log\_density}. The initial prompt is given below.

\begin{tcolorbox}[ colback=blue!5!white, colframe=blue!75!black, title=\textbf{Initial Prompt for the LLM Baseline}, fonttitle=\bfseries, fontupper=\small ]
\textbf{System:} You are an expert in statistics and machine learning who writes correct, efficient Python.

\medskip
\textbf{User:} Write a Python program that fits a generative model to a tabular dataset and draws new synthetic rows from it.

The dataset is called ``$\langle$\textit{dataset name}$\rangle$''.
The dataset has $\langle d\rangle$ numeric columns. Each column was standardized to zero mean and unit variance on the full dataset.

These are the first 50 of $\langle n\rangle$ rows, in random order, rounded to 3 decimals:\\
\texttt{$\langle$\textit{column names, comma separated}$\rangle$}\\
\texttt{$\langle$\textit{50 rows, comma separated}$\rangle$}

Your program has to define exactly this interface:
\begin{verbatim}
class Model:
    def fit(self, data):
        # data: numpy array of shape (n_rows, d), receives all rows
    def sample(self, n):
        # returns a numpy array of shape (n, d)
    def log_density(self, x):
        # x: numpy array of shape (m, d),
        # returns the log density of each row, shape (m,)
\end{verbatim}

Evaluation: the program is scored by the mean \texttt{log\_density} it assigns to unseen rows of the same dataset. Higher is better.

\texttt{log\_density} has to be the normalized log density of the same distribution that \texttt{sample} draws from, and it has to be finite on unseen rows.

You choose both the model and how it is fitted. This can be a library call such as a sklearn estimator's \texttt{fit} or your own optimization loop.

Rules:
\begin{itemize}\setlength\itemsep{0pt}
\item Available libraries: numpy, scipy, sklearn, torch. CPU only, 4 threads, no network, no file access.
\item \texttt{fit} and \texttt{sample} together have to finish within 300 seconds.
\item Samples must be new rows. Programs whose samples copy or nearly copy the rows given to \texttt{fit} are rejected.
\item Do not hardcode values from the rows above. The program receives all rows in \texttt{fit}.
\end{itemize}

Think about the structure of this data, then reply with one fenced python code block containing the complete program.
\end{tcolorbox}

\paragraph{Results.}
Table~\ref{tab:llm} reports the results after the first round (\emph{1 round}) and after all three rounds (\emph{3 rounds}).

\begin{table}[h]
\centering
\caption{LLM-based program inference (Claude Haiku~5.5) compared with PRODiGI-FT and a BIC-selected GMM on 96 in-distribution (ID) and 96 real-world (RW) datasets. Mean $\pm$ SEM across datasets.}
\label{tab:llm}
\begin{tabular}{lcccc}
\toprule
Method & ID MMD ($\times10^3$) $\downarrow$ & ID Spearman $\uparrow$ & ID score cosine $\uparrow$ & RW MMD ($\times10^3$) $\downarrow$ \\
\midrule
Haiku, 1 round  & \textbf{0.29 $\pm$ 0.15} & 0.832 $\pm$ 0.015 & 0.587 $\pm$ 0.034 & 18.79 $\pm$ 3.84 \\
Haiku, 3 rounds & 0.41 $\pm$ 0.30 & \textbf{0.900 $\pm$ 0.011} & 0.652 $\pm$ 0.034 & 19.00 $\pm$ 3.92 \\
PRODiGI-FT      & 0.56 $\pm$ 0.20 & 0.895 $\pm$ 0.014 & \textbf{0.780 $\pm$ 0.027} & 37.29 $\pm$ 4.58 \\
GMM (BIC)       & 0.78 $\pm$ 0.30 & 0.817 $\pm$ 0.016 & 0.590 $\pm$ 0.035 & \textbf{16.25 $\pm$ 3.21} \\
\bottomrule
\end{tabular}
\end{table}

LLM-generated programs achieve strong generation quality and a lower mean MMD than PRODiGI-FT on both in-distribution and real-world data. On in-distribution data, density ranking is on par with PRODiGI-FT (Spearman 0.900 vs.\ 0.895), while PRODiGI-FT estimates score directions considerably better (cosine 0.780 vs.\ 0.652). Further rounds improve Spearman (0.832$\rightarrow$0.900) and score cosine (0.587$\rightarrow$0.652), but not MMD (0.29$\rightarrow$0.41 ID, 18.79$\rightarrow$19.00 RW). This is consistent with selection by validation likelihood: a higher likelihood does not necessarily mean a closer match of the samples.

\paragraph{Chosen program structures.}
Although the LLM is free to propose any program, it almost always writes a Gaussian mixture. This holds for 94/96 in-distribution datasets (38 plain GMMs, 56 combined with transformations) and 89/96 real-world datasets (57 plain GMMs, 32 with transformations). The strong real-world results of the LLM pipeline are therefore closely tied to the strength of fitted full-covariance GMMs on these datasets.

\paragraph{Cost.}
On in-distribution data, the three-round pipeline takes 162\,s per dataset: 47\,s of API latency and 113\,s executing the generated programs. This compares with 12.90\,s for PRODiGI-FT and 0.06\,s for zero-shot PRODiGI. On real-world data, the times are 120\,s, 9.64\,s, and 0.06\,s. The LLM pipeline is therefore about $12.5\times$ slower than PRODiGI-FT and 2{,}000--2{,}700$\times$ slower than zero-shot PRODiGI. It also incurs about \$0.007--0.008 of API charges per dataset, excluding execution. PRODiGI predicts an explicit program in a single forward pass, sees the full input set rather than a 50-row excerpt, and does not depend on an external service.

\subsection{Comparing to Set Transformer}\label{app:settransformer}

The Set Transformer~\citep{SetTransformer} demonstrated amortized inference of GMM parameters from a set of observations, and PRODiGI uses a Set Transformer encoder. The original model, however, predicts the parameters of a GMM with a fixed dimensionality and a fixed number of components. PRODiGI's template-based non-autoregressive decoder instead predicts programs with variable structure from three families (GMM, SCM, Copula) and across dimensionalities. To quantify the difference, we reimplemented the Set Transformer GMM predictor of the original paper. We trained it on our 2D GMM prior with a fixed output of 10 components, following the original paper's optimization pipeline of $50000$ steps. Because the Set Transformer is restricted to a fixed dimensionality, we evaluate on the 2D in-distribution datasets: 32 GMM, 16 Copula, and 16 SCM. We compare against PRODiGI and PRODiGI-2D, a variant trained only on 1D and 2D data.

\begin{table}[h]
\centering
\caption{Zero-shot comparison with a Set Transformer GMM predictor on 2D in-distribution datasets (number of datasets in parentheses).}
\label{tab:settf_zs}
\begin{tabular}{llccc}
\toprule
Data & Metric & PRODiGI & PRODiGI-2D & Set Transformer \\
\midrule
GMM (32)    & MMD ($\times10^3$) $\downarrow$ & 21.45 & \textbf{12.42} & 42.62 \\
            & Spearman $\uparrow$             & 0.46  & \textbf{0.59}  & 0.50 \\
            & Score cosine $\uparrow$         & 0.30  & \textbf{0.46}  & 0.40 \\
\midrule
Copula (16) & MMD ($\times10^3$) $\downarrow$ & 5.52  & \textbf{0.34}  & 27.34 \\
            & Spearman $\uparrow$             & 0.75  & \textbf{0.99}  & 0.54 \\
            & Score cosine $\uparrow$         & 0.69  & \textbf{0.96}  & 0.44 \\
\midrule
SCM (16)    & MMD ($\times10^3$) $\downarrow$ & 25.63 & \textbf{11.77} & 52.29 \\
            & Spearman $\uparrow$             & 0.79  & \textbf{0.84}  & 0.30 \\
            & Score cosine $\uparrow$         & 0.74  & \textbf{0.78}  & 0.03 \\
\bottomrule
\end{tabular}
\end{table}

\paragraph{Zero-shot.}
Zero-shot, PRODiGI-2D outperforms the Set Transformer on every family and metric (Table~\ref{tab:settf_zs}). Even on GMM data, the setting the Set Transformer was designed for, PRODiGI-2D reaches a 71\% lower MMD and better density ranking and score directions. The general PRODiGI model, which covers all dimensionality settings, also outperforms the Set Transformer on MMD across all three families.

\begin{table}[h]
\centering
\caption{Comparison after refinement. PRODiGI-2D-FT uses 300 Adam steps (learning rate 0.01) on the NLL; the Set Transformer prediction is refined by EM until convergence (scikit-learn default tolerance $10^{-3}$).}
\label{tab:settf_ft}
\begin{tabular}{llcc}
\toprule
Data & Metric & PRODiGI-2D-FT & Set Transformer + EM \\
\midrule
GMM (32)    & MMD ($\times10^3$) $\downarrow$ & 0.74 & \textbf{0.30} \\
            & Spearman $\uparrow$             & \textbf{0.95} & \textbf{0.95} \\
            & Score cosine $\uparrow$         & 0.87 & \textbf{0.89} \\
\midrule
Copula (16) & MMD ($\times10^3$) $\downarrow$ & 0.72 & \textbf{0.66} \\
            & Spearman $\uparrow$             & \textbf{0.99} & 0.86 \\
            & Score cosine $\uparrow$         & \textbf{0.97} & 0.68 \\
\midrule
SCM (16)    & MMD ($\times10^3$) $\downarrow$ & 0.82 & \textbf{0.58} \\
            & Spearman $\uparrow$             & \textbf{0.98} & 0.80 \\
            & Score cosine $\uparrow$         & \textbf{0.96} & 0.28 \\
\bottomrule
\end{tabular}
\end{table}

\paragraph{After refinement.}
Each method is refined with its own procedure (Table~\ref{tab:settf_ft}): EM for the Set Transformer's GMM, and program-space refinement for PRODiGI-2D. The two procedures therefore do not have matched budgets; EM is considerably faster (less than $0.1$ seconds on average). On GMM data, Set Transformer + EM reaches a lower MMD (0.30 vs.\ 0.74$\times10^{-3}$), with comparable density ranking and score directions. On Copula and SCM data, EM still matches the samples slightly better. However, PRODiGI-2D-FT is far more accurate in density ranking (Spearman 0.99 vs.\ 0.86 and 0.98 vs.\ 0.80) and score direction (cosine 0.97 vs.\ 0.68 and 0.96 vs.\ 0.28). This is because a GMM can match samples from a non-Gaussian distribution, but its densities and scores remain those of a mixture of Gaussians. PRODiGI instead infers a program from the matching family.

These results show that PRODiGI improves on the Set Transformer even in its original, restricted zero-shot GMM setting. They also show that PRODiGI's main benefit lies beyond it: amortized inference of heterogeneous, variable-structure programs, that are not restricted to a single prior family and structure and whose densities and scores remain faithful after refinement.


\clearpage
\subsection{PRODiGI Outputs}\label{app:prodigioutput}

While \method can generate programs containing thousands of tokens, we restrict the appendix to the two examples shown in Figure \ref{fig:gen_rw_heatmap} for brevity. Both examples generate compact Copula programs.

\begin{tcolorbox}[ colback=blue!5!white, colframe=blue!75!black, title=\textbf{PRODiGI Output for the FinancialBeneficiary Dataset}, fonttitle=\bfseries, fontupper=\small ]
\begin{verbatim}
[COPULA][DIMENSIONALITY][2]
[START_COMPONENT]
[INDEX][0][MARGINAL][UNIFORM]
[PARAMS][-0.0248][1.0000][0.0094][0.0115]
[END_COMPONENT]
[START_COMPONENT]
[INDEX][1][MARGINAL][KUMARASWAMY]
[PARAMS][-0.0108][0.8946][0.7638][2.8795]
[END_COMPONENT]
[CORRELATIONS][0.0026][END_CORRELATIONS]
[END_COPULA]
\end{verbatim}
\end{tcolorbox}

\begin{tcolorbox}[ colback=blue!5!white, colframe=blue!75!black, title=\textbf{PRODiGI-FT Output for the FinancialBeneficiary Dataset}, fonttitle=\bfseries, fontupper=\small ]
\begin{verbatim}
[COPULA][DIMENSIONALITY][2]
[START_COMPONENT]
[INDEX][0][MARGINAL][UNIFORM]
[PARAMS][-0.0248][1.0000][0.0094][0.0115]
[END_COMPONENT]
[START_COMPONENT]
[INDEX][1][MARGINAL][KUMARASWAMY]
[PARAMS][-0.0108][0.8946][0.0951][1.4659]
[END_COMPONENT]
[CORRELATIONS][-0.0017][END_CORRELATIONS]
[END_COPULA]
\end{verbatim}
\end{tcolorbox}

\begin{tcolorbox}[ colback=blue!5!white, colframe=blue!75!black, title=\textbf{PRODiGI Output for the FinancialTransactionAnomalies Dataset}, fonttitle=\bfseries, fontupper=\small ]
\begin{verbatim}
[COPULA][DIMENSIONALITY][2]
[START_COMPONENT]
[INDEX][0][MARGINAL][UNIFORM]
[PARAMS][-0.0043][1.0083][0.0056][0.0254]
[END_COMPONENT]
[START_COMPONENT]
[INDEX][1][MARGINAL][KUMARASWAMY]
[PARAMS][0.0086][0.9578][1.4854][3.6404]
[END_COMPONENT]
[CORRELATIONS][-0.0012][END_CORRELATIONS]
[END_COPULA]
\end{verbatim}
\end{tcolorbox}

\begin{tcolorbox}[ colback=blue!5!white, colframe=blue!75!black, title=\textbf{PRODiGI-FT Output for the FinancialTransactionAnomalies Dataset}, fonttitle=\bfseries, fontupper=\small ]
\begin{verbatim}
[COPULA][DIMENSIONALITY][2]
[START_COMPONENT]
[INDEX][0][MARGINAL][UNIFORM]
[PARAMS][-0.0043][1.0083][0.0056][0.0254]
[END_COMPONENT]
[START_COMPONENT]
[INDEX][1][MARGINAL][KUMARASWAMY]
[PARAMS][0.0086][0.9578][1.2654][4.3308]
[END_COMPONENT]
[CORRELATIONS][-0.0056][END_CORRELATIONS]
[END_COPULA]
\end{verbatim}
\end{tcolorbox}



\section{Detailed Results}
\label{app:detailed_tables}

We provide detailed results for each algorithm and metric split across dataset dimensions and prior family in the following tables.




\begin{table}[h]
    \centering
    \caption{\textbf{Distribution match} performance via MMD ($\downarrow$) and rank, as well as runtime in seconds ($\downarrow$) and runtime rank for varying data dimensions.
    Average rank per metric (lower is better) is across methods over test datasets from ALL prior families.}
    \vspace{-0.1in}
    \label{tab:mmd_all_priors}
    \resizebox{\columnwidth}{!}{%
\begin{tabular}{lc|cc|ccccccc}
        \toprule
        \textbf{Dimension} & \textbf{Metric} & \textbf{PRODiGI} & \textbf{PRODiGI-FT} & \textbf{DiscoFormer} & \textbf{DiscoFormer-FT}  &\textbf{NNKDE} & \textbf{NNKDE-FT} & \textbf{TabPFN} & \textbf{KDE} & \textbf{GMM} \\
        \midrule

        \multirow{4}{*}{\textbf{Dim 1}}
        & $\mathrm{MMD}\;(\times 10^3)$ & \pmv{2.81}{1.37} & \pmv{0.85}{0.63} & \pmv{91.35}{42.99} & \pmv{276.79}{98.84} & \pmv{8.34}{1.25} & \pmv{0.53}{0.28} & \pmv{0.41}{0.13} & \pmv{\textbf{0.20}}{0.13} & \pmv{\underline{0.35}}{0.18} \\
        & Rank & \pmv{4.79}{0.49} & \pmv{\underline{2.92}}{0.50} & \pmv{7.96}{0.43} & \pmv{8.33}{0.18} & \pmv{7.00}{0.29} & \pmv{3.96}{0.45} & \pmv{4.21}{0.49} & \pmv{\textbf{2.75}}{0.44} & \pmv{3.08}{0.33} \\
        \noalign{\smallskip}\cline{2-11}\noalign{\smallskip}
        & Time & \pmv{\textbf{0.03}}{0.00} & \pmv{1.47}{0.16} & \pmv{93.28}{0.61} & \pmv{93.80}{0.61} & \pmv{0.38}{0.19} & \pmv{1.17}{0.01} & \pmv{1.52}{0.02} & \pmv{\underline{0.11}}{0.00} & \pmv{8.71}{0.18} \\
        & Time Rank & \pmv{\textbf{1.00}}{0.00} & \pmv{4.83}{0.23} & \pmv{8.00}{0.00} & \pmv{9.00}{0.00} & \pmv{3.25}{0.24} & \pmv{4.25}{0.17} & \pmv{5.67}{0.14} & \pmv{\underline{2.00}}{0.00} & \pmv{7.00}{0.00} \\
        \midrule

        \multirow{4}{*}{\textbf{Dim 2}}
        & $\mathrm{MMD}\;(\times 10^3)$ & \pmv{1.38}{0.90} & \pmv{\textbf{0.05}}{0.04} & \pmv{12.26}{3.79} & \pmv{18.70}{8.47} & \pmv{66.38}{8.58} & \pmv{8.47}{2.72} & \pmv{0.15}{0.07} & \pmv{0.45}{0.33} & \pmv{\underline{0.09}}{0.08} \\
        & Rank & \pmv{3.58}{0.35} & \pmv{\textbf{2.50}}{0.27} & \pmv{7.42}{0.25} & \pmv{6.83}{0.45} & \pmv{8.83}{0.11} & \pmv{6.50}{0.22} & \pmv{3.54}{0.40} & \pmv{3.00}{0.24} & \pmv{\underline{2.79}}{0.33} \\
\bottomrule
    \end{tabular}%
    }
\end{table}

\begin{table}[h]
    \centering
    \caption{\textbf{Rejection rate of two-sample tests for distribution match} w.r.t. MMD  over test datasets from ALL prior families.
    Tests under generated data that successfully match the input data distribution fail to reject the null (i.e., $p>0.05$) that the two samples come from the same distribution.}
    \vspace{-0.1in}
    \label{tab:pval_mmd_all_priors}
    \resizebox{\columnwidth}{!}{%
\begin{tabular}{l|cc|ccccccc}
    \toprule
        \textbf{Dimension}  & \textbf{PRODiGI} & \textbf{PRODiGI-FT} & \textbf{DiscoFormer} & \textbf{DiscoFormer-FT}  &\textbf{NNKDE} & \textbf{NNKDE-FT} & \textbf{TabPFN} & \textbf{KDE} & \textbf{GMM} \\
        \midrule
    \textbf{Dim 1}     & 41.7\% & \textbf{8.3\%} & 91.7\% & 100.0\% & 91.7\% & 25.0\% & \textbf{8.3\%} & \textbf{8.3\%} & \textbf{8.3\%} \\
    \textbf{Dim 2}     & 16.7\% & \textbf{0.0\%} & 100.0\% & 83.3\% & 100.0\% & 91.7\% & 16.7\% & 8.3\% & \textbf{0.0\%} \\
    \textbf{Dim 5}     & 66.7\% & \textbf{25.0\%} & 100.0\% & 100.0\% & 91.7\% & 83.3\% & \underline{33.3\%} & 75.0\% & \underline{33.3\%} \\
    \textbf{Dim 10}    & 75.0\% & \textbf{25.0\%} & 100.0\% & 100.0\% & 100.0\% & 83.3\% & \textbf{25.0\%} & 66.7\% & 41.7\% \\
    \textbf{Dim 20}    & 83.3\% & \textbf{8.3\%} & 100.0\% & 100.0\% & 91.7\% & 91.7\% & 41.7\% & 83.3\% & \underline{16.7\%} \\
    \textbf{Dim 30}    & 66.7\% & \textbf{25.0\%} & 100.0\% & 100.0\% & 100.0\% & 100.0\% & \underline{33.3\%} & 91.7\% & 50.0\% \\
    \textbf{Dim 40}    & 58.3\% & \textbf{8.3\%} & 100.0\% & 100.0\% & 83.3\% & 83.3\% & \underline{33.3\%} & 83.3\% & 58.3\% \\
    \textbf{Dim 50}    & 50.0\% & \textbf{16.7\%} & 100.0\% & 100.0\% & 83.3\% & 83.3\% & \underline{41.7\%} & 75.0\% & \underline{41.7\%} \\
    \midrule
    \textbf{Overall}   & 57.3\% & \textbf{14.6\%} & 99.0\% & 97.9\% & 92.7\% & 80.2\% & \underline{29.2\%} & 61.5\% & 31.3\% \\
    \bottomrule
\end{tabular}

    }
\end{table}

\begin{table}[t]
    \centering
    \caption{\textbf{Distribution match} performance via MMD ($\downarrow$) and rank, as well as runtime in seconds ($\downarrow$) and runtime rank for varying data dimensions.
    Average rank per metric (lower is better) is across methods over test datasets from the GMM prior.}
    \vspace{-0.1in}
    \label{tab:mmd_gmm}
    \resizebox{\columnwidth}{!}{%
    \begin{tabular}{lc|cc|ccccccc}
        \toprule
        \textbf{Dimension} & \textbf{Metric} & \textbf{PRODiGI} & \textbf{PRODiGI-FT} & \textbf{DiscoFormer} & \textbf{DiscoFormer-FT}  &\textbf{NNKDE} & \textbf{NNKDE-FT} & \textbf{TabPFN} & \textbf{KDE} & \textbf{GMM} \\
        \midrule

        \multirow{4}{*}{\textbf{Dim 1}}
        & $\mathrm{MMD}\;(\times 10^3)$ & \pmv{5.31}{2.70} & \pmv{0.32}{0.21} & \pmv{59.23}{41.13} & \pmv{76.56}{59.00} & \pmv{6.77}{0.95} & \pmv{1.22}{0.71} & \pmv{\underline{0.23}}{0.11} & \pmv{\textbf{0.00}}{0.00} & \pmv{0.66}{0.45} \\
        & Rank & \pmv{6.50}{0.25} & \pmv{3.00}{0.61} & \pmv{8.25}{0.41} & \pmv{8.25}{0.22} & \pmv{7.00}{0.61} & \pmv{4.38}{0.54} & \pmv{\underline{2.88}}{0.37} & \pmv{\textbf{1.62}}{0.32} & \pmv{3.12}{0.45} \\
        \noalign{\smallskip}\cline{2-11}\noalign{\smallskip}
        & Time & \pmv{\textbf{0.03}}{0.00} & \pmv{1.12}{0.01} & \pmv{92.61}{0.07} & \pmv{93.18}{0.08} & \pmv{0.18}{0.01} & \pmv{1.16}{0.00} & \pmv{1.56}{0.04} & \pmv{\underline{0.11}}{0.00} & \pmv{8.30}{0.05} \\
        & Time Rank & \pmv{\textbf{1.00}}{0.00} & \pmv{4.00}{0.00} & \pmv{8.00}{0.00} & \pmv{9.00}{0.00} & \pmv{3.00}{0.00} & \pmv{5.00}{0.00} & \pmv{6.00}{0.00} & \pmv{\underline{2.00}}{0.00} & \pmv{7.00}{0.00} \\
        \midrule

        \multirow{4}{*}{\textbf{Dim 2}}
        & $\mathrm{MMD}\;(\times 10^3)$ & \pmv{4.02}{2.15} & \pmv{\underline{0.13}}{0.12} & \pmv{15.23}{2.69} & \pmv{8.24}{2.43} & \pmv{52.62}{17.98} & \pmv{8.83}{2.49} & \pmv{0.22}{0.19} & \pmv{0.28}{0.24} & \pmv{\textbf{0.00}}{0.00} \\
        & Rank & \pmv{4.62}{0.65} & \pmv{\underline{2.75}}{0.67} & \pmv{8.00}{0.35} & \pmv{6.75}{0.54} & \pmv{8.75}{0.22} & \pmv{6.25}{0.22} & \pmv{3.00}{0.31} & \pmv{\underline{2.75}}{0.38} & \pmv{\textbf{2.12}}{0.21} \\
\bottomrule
    \end{tabular}%
    }
\end{table}

\begin{table}[t]
    \centering
   \caption{\textbf{Distribution match} performance via MMD ($\downarrow$) and rank, as well as runtime in seconds ($\downarrow$) and runtime rank for varying data dimensions.
    Average rank per metric (lower is better) is across methods over test datasets from the SCM prior.}
    \vspace{-0.1in}
    \label{tab:mmd_scm}
    \resizebox{\columnwidth}{!}{%
    \begin{tabular}{lc|cc|ccccccc}
        \toprule
        \textbf{Dimension} & \textbf{Metric} & \textbf{PRODiGI} & \textbf{PRODiGI-FT} & \textbf{DiscoFormer} & \textbf{DiscoFormer-FT}  &\textbf{NNKDE} & \textbf{NNKDE-FT} & \textbf{TabPFN} & \textbf{KDE} & \textbf{GMM} \\
        \midrule

        \multirow{4}{*}{\textbf{Dim 1}}
        & $\mathrm{MMD}\;(\times 10^3)$ & \pmv{3.07}{2.47} & \pmv{2.23}{1.68} & \pmv{179.43}{108.64} & \pmv{538.42}{175.53} & \pmv{9.01}{3.47} & \pmv{\underline{0.00}}{0.00} & \pmv{0.41}{0.22} & \pmv{0.24}{0.17} & \pmv{\textbf{0.00}}{0.00} \\
        & Rank & \pmv{4.75}{0.74} & \pmv{3.75}{1.14} & \pmv{8.25}{0.22} & \pmv{8.50}{0.43} & \pmv{6.50}{0.25} & \pmv{\underline{2.75}}{0.67} & \pmv{4.88}{1.01} & \pmv{3.38}{0.60} & \pmv{\textbf{2.25}}{0.28} \\
        \noalign{\smallskip}\cline{2-11}\noalign{\smallskip}
        & Time & \pmv{\textbf{0.03}}{0.00} & \pmv{1.55}{0.05} & \pmv{92.65}{0.15} & \pmv{93.13}{0.09} & \pmv{0.17}{0.01} & \pmv{1.18}{0.02} & \pmv{1.51}{0.03} & \pmv{\underline{0.10}}{0.00} & \pmv{9.31}{0.35} \\
        & Time Rank & \pmv{\textbf{1.00}}{0.00} & \pmv{5.50}{0.25} & \pmv{8.00}{0.00} & \pmv{9.00}{0.00} & \pmv{3.00}{0.00} & \pmv{4.00}{0.00} & \pmv{5.50}{0.25} & \pmv{\underline{2.00}}{0.00} & \pmv{7.00}{0.00} \\
        \midrule

        \multirow{4}{*}{\textbf{Dim 2}}
        & $\mathrm{MMD}\;(\times 10^3)$ & \pmv{0.02}{0.02} & \pmv{\textbf{0.00}}{0.00} & \pmv{7.50}{2.78} & \pmv{25.72}{17.60} & \pmv{87.66}{4.73} & \pmv{5.60}{3.88} & \pmv{0.22}{0.07} & \pmv{0.04}{0.03} & \pmv{\underline{0.01}}{0.01} \\
        & Rank & \pmv{2.75}{0.12} & \pmv{\textbf{2.25}}{0.38} & \pmv{7.25}{0.41} & \pmv{7.25}{0.41} & \pmv{9.00}{0.00} & \pmv{6.25}{0.41} & \pmv{4.75}{0.54} & \pmv{3.00}{0.31} & \pmv{\underline{2.50}}{0.18} \\
\bottomrule
    \end{tabular}%
    }
\end{table}

\begin{table}[t]
    \centering
    \caption{\textbf{Distribution match} performance via MMD ($\downarrow$) and rank, as well as runtime in seconds ($\downarrow$) and runtime rank for varying data dimensions.
    Average rank per metric (lower is better) is across methods over test datasets from the Copula prior.}
    \vspace{-0.1in}
    \label{tab:mmd_copula}
    \resizebox{\columnwidth}{!}{%
    \begin{tabular}{lc|cc|ccccccc}
        \toprule
        \textbf{Dimension} & \textbf{Metric} & \textbf{PRODiGI} & \textbf{PRODiGI-FT} & \textbf{DiscoFormer} & \textbf{DiscoFormer-FT}  &\textbf{NNKDE} & \textbf{NNKDE-FT} & \textbf{TabPFN} & \textbf{KDE} & \textbf{GMM} \\
        \midrule

        \multirow{4}{*}{\textbf{Dim 1}}
        & $\mathrm{MMD}\;(\times 10^3)$ & \pmv{\underline{0.06}}{0.03} & \pmv{\textbf{0.00}}{0.00} & \pmv{35.40}{12.62} & \pmv{215.40}{159.87} & \pmv{9.24}{0.44} & \pmv{0.38}{0.19} & \pmv{0.60}{0.27} & \pmv{0.38}{0.33} & \pmv{0.40}{0.19} \\
        & Rank & \pmv{\underline{3.12}}{0.37} & \pmv{\textbf{2.00}}{0.47} & \pmv{7.38}{1.14} & \pmv{8.25}{0.22} & \pmv{7.50}{0.43} & \pmv{4.75}{0.74} & \pmv{4.88}{0.57} & \pmv{3.25}{0.88} & \pmv{3.88}{0.62} \\
        \noalign{\smallskip}\cline{2-11}\noalign{\smallskip}
        & Time & \pmv{\textbf{0.02}}{0.00} & \pmv{1.75}{0.43} & \pmv{94.58}{1.64} & \pmv{95.07}{1.65} & \pmv{0.78}{0.52} & \pmv{1.16}{0.00} & \pmv{1.50}{0.03} & \pmv{\underline{0.11}}{0.00} & \pmv{8.53}{0.10} \\
        & Time Rank & \pmv{\textbf{1.00}}{0.00} & \pmv{5.00}{0.35} & \pmv{8.00}{0.00} & \pmv{9.00}{0.00} & \pmv{3.75}{0.65} & \pmv{3.75}{0.22} & \pmv{5.50}{0.25} & \pmv{\underline{2.00}}{0.00} & \pmv{7.00}{0.00} \\
        \midrule

        \multirow{4}{*}{\textbf{Dim 2}}
        & $\mathrm{MMD}\;(\times 10^3)$ & \pmv{0.09}{0.05} & \pmv{\underline{0.01}}{0.01} & \pmv{14.03}{10.29} & \pmv{22.14}{16.97} & \pmv{58.87}{11.96} & \pmv{10.98}{6.46} & \pmv{\textbf{0.00}}{0.00} & \pmv{1.02}{0.89} & \pmv{0.25}{0.22} \\
        & Rank & \pmv{3.38}{0.48} & \pmv{\textbf{2.50}}{0.18} & \pmv{7.00}{0.35} & \pmv{6.50}{1.15} & \pmv{8.75}{0.22} & \pmv{7.00}{0.35} & \pmv{\underline{2.88}}{0.72} & \pmv{3.25}{0.52} & \pmv{3.75}{0.72} \\
\bottomrule
    \end{tabular}%
    }
\end{table}


\begin{table}[t]
    \centering
    \caption{\textbf{Density estimation} performance via MAE ($\downarrow$) and rank, as well as runtime in seconds ($\downarrow$) and runtime rank for varying data dimensions.
    Average rank per metric (lower is better) is across methods over test datasets from ALL prior families.}
    \vspace{-0.1in}
    \label{tab:mae_dense_all_priors}
    \resizebox{\columnwidth}{!}{%
    \begin{tabular}{lc|cc|ccccccc}
        \toprule
        \textbf{Dimension} & \textbf{Metric} & \textbf{PRODiGI} & \textbf{PRODiGI-FT} & \textbf{DiscoFormer} & \textbf{DiscoFormer-FT}  &\textbf{NNKDE} & \textbf{NNKDE-FT} & \textbf{TabPFN} & \textbf{KDE} & \textbf{GMM} \\
        \midrule

        \multirow{4}{*}{\textbf{Dim 1}}
        & MAE & \pmv{0.31}{0.11} & \pmv{\textbf{0.13}}{0.04} & \pmv{0.66}{0.10} & \pmv{0.56}{0.10} & \pmv{0.27}{0.04} & \pmv{0.21}{0.01} & \pmv{2.41}{0.02} & \pmv{\underline{0.16}}{0.03} & \pmv{0.19}{0.04} \\
        & Rank & \pmv{3.71}{0.62} & \pmv{\textbf{2.04}}{0.51} & \pmv{7.83}{0.16} & \pmv{6.92}{0.08} & \pmv{5.00}{0.20} & \pmv{4.33}{0.32} & \pmv{9.00}{0.00} & \pmv{\underline{2.75}}{0.49} & \pmv{3.42}{0.30} \\
        \noalign{\smallskip}\cline{2-11}\noalign{\smallskip}
        & Time & \pmv{0.31}{0.01} & \pmv{1.52}{0.17} & \pmv{\textbf{0.04}}{0.02} & \pmv{0.62}{0.04} & \pmv{0.17}{0.00} & \pmv{1.16}{0.01} & \pmv{3.38}{0.99} & \pmv{\underline{0.14}}{0.00} & \pmv{8.72}{0.18} \\
        & Time Rank & \pmv{4.00}{0.00} & \pmv{7.25}{0.17} & \pmv{\textbf{1.17}}{0.16} & \pmv{5.00}{0.00} & \pmv{2.83}{0.11} & \pmv{6.42}{0.14} & \pmv{7.42}{0.30} & \pmv{\underline{2.00}}{0.12} & \pmv{8.92}{0.08} \\
        \midrule

        \multirow{4}{*}{\textbf{Dim 2}}
        & MAE & \pmv{0.93}{0.32} & \pmv{\textbf{0.29}}{0.07} & \pmv{0.65}{0.09} & \pmv{0.71}{0.08} & \pmv{0.80}{0.07} & \pmv{0.71}{0.03} & \pmv{1.66}{0.18} & \pmv{0.35}{0.05} & \pmv{\underline{0.33}}{0.07} \\
        & Rank & \pmv{5.00}{0.87} & \pmv{\textbf{1.83}}{0.23} & \pmv{5.17}{0.28} & \pmv{5.92}{0.18} & \pmv{7.25}{0.21} & \pmv{5.83}{0.48} & \pmv{8.83}{0.11} & \pmv{2.67}{0.18} & \pmv{\underline{2.50}}{0.42} \\
        \noalign{\smallskip}\cline{2-11}\noalign{\smallskip}
        & Time & \pmv{0.53}{0.22} & \pmv{2.19}{0.22} & \pmv{\textbf{0.01}}{0.00} & \pmv{0.56}{0.00} & \pmv{\underline{0.19}}{0.01} & \pmv{1.16}{0.00} & \pmv{5.48}{0.54} & \pmv{0.46}{0.01} & \pmv{11.35}{0.54} \\
\bottomrule
    \end{tabular}%
    }
\end{table}

\begin{table}[t]
    \centering
   \caption{\textbf{Density estimation} performance via MAE ($\downarrow$) and rank, as well as runtime in seconds ($\downarrow$) and runtime rank for varying data dimensions.
    Average rank per metric (lower is better) is across methods over test datasets from the GMM prior.}
    \vspace{-0.1in}
    \label{tab:mae_dense_gmm}
    \resizebox{\columnwidth}{!}{%
    \begin{tabular}{lc|cc|ccccccc}
        \toprule
        \textbf{Dimension} & \textbf{Metric} & \textbf{PRODiGI} & \textbf{PRODiGI-FT} & \textbf{DiscoFormer} & \textbf{DiscoFormer-FT}  &\textbf{NNKDE} & \textbf{NNKDE-FT} & \textbf{TabPFN} & \textbf{KDE} & \textbf{GMM} \\
        \midrule

        \multirow{4}{*}{\textbf{Dim 1}}
        & MAE & \pmv{0.64}{0.23} & \pmv{0.16}{0.05} & \pmv{0.54}{0.08} & \pmv{0.51}{0.07} & \pmv{0.23}{0.05} & \pmv{0.18}{0.01} & \pmv{2.38}{0.01} & \pmv{\textbf{0.09}}{0.01} & \pmv{\underline{0.12}}{0.01} \\
        & Rank & \pmv{6.50}{0.43} & \pmv{\underline{2.50}}{0.56} & \pmv{7.50}{0.43} & \pmv{7.00}{0.00} & \pmv{4.75}{0.22} & \pmv{4.00}{0.35} & \pmv{9.00}{0.00} & \pmv{\textbf{1.25}}{0.22} & \pmv{\underline{2.50}}{0.25} \\
        \noalign{\smallskip}\cline{2-11}\noalign{\smallskip}
        & Time & \pmv{0.30}{0.00} & \pmv{1.16}{0.02} & \pmv{\textbf{0.01}}{0.00} & \pmv{0.56}{0.00} & \pmv{0.18}{0.01} & \pmv{1.16}{0.00} & \pmv{5.20}{2.61} & \pmv{\underline{0.13}}{0.00} & \pmv{8.30}{0.05} \\
        & Time Rank & \pmv{4.00}{0.00} & \pmv{7.00}{0.35} & \pmv{\textbf{1.00}}{0.00} & \pmv{5.00}{0.00} & \pmv{3.00}{0.00} & \pmv{6.50}{0.25} & \pmv{7.75}{0.54} & \pmv{\underline{2.00}}{0.00} & \pmv{8.75}{0.22} \\
        \midrule

        \multirow{4}{*}{\textbf{Dim 2}}
        & MAE & \pmv{0.79}{0.23} & \pmv{0.22}{0.12} & \pmv{0.44}{0.09} & \pmv{0.54}{0.11} & \pmv{0.61}{0.06} & \pmv{0.63}{0.03} & \pmv{1.42}{0.23} & \pmv{\underline{0.19}}{0.03} & \pmv{\textbf{0.10}}{0.03} \\
        & Rank & \pmv{6.50}{1.60} & \pmv{\underline{2.25}}{0.22} & \pmv{4.50}{0.25} & \pmv{5.75}{0.41} & \pmv{6.50}{0.25} & \pmv{6.25}{0.74} & \pmv{8.75}{0.22} & \pmv{3.00}{0.35} & \pmv{\textbf{1.50}}{0.43} \\
        \noalign{\smallskip}\cline{2-11}\noalign{\smallskip}
        & Time & \pmv{0.29}{0.00} & \pmv{1.60}{0.05} & \pmv{\textbf{0.01}}{0.00} & \pmv{0.56}{0.00} & \pmv{\underline{0.17}}{0.01} & \pmv{1.16}{0.00} & \pmv{4.68}{0.78} & \pmv{0.43}{0.00} & \pmv{9.34}{0.39} \\
\bottomrule
    \end{tabular}%
    }
\end{table}

\begin{table}[t]
    \centering
  \caption{\textbf{Density estimation} performance via MAE ($\downarrow$) and rank, as well as runtime in seconds ($\downarrow$) and runtime rank for varying data dimensions.
    Average rank per metric (lower is better) is across methods over test datasets from the SCM prior.}
    \vspace{-0.1in}
    \label{tab:mae_dense_scm}
    \resizebox{\columnwidth}{!}{%
    \begin{tabular}{lc|cc|ccccccc}
        \toprule
        \textbf{Dimension} & \textbf{Metric} & \textbf{PRODiGI} & \textbf{PRODiGI-FT} & \textbf{DiscoFormer} & \textbf{DiscoFormer-FT}  &\textbf{NNKDE} & \textbf{NNKDE-FT} & \textbf{TabPFN} & \textbf{KDE} & \textbf{GMM} \\
        \midrule

        \multirow{4}{*}{\textbf{Dim 1}}
        & MAE & \pmv{0.23}{0.09} & \pmv{\underline{0.19}}{0.08} & \pmv{0.99}{0.18} & \pmv{0.83}{0.22} & \pmv{0.39}{0.09} & \pmv{0.25}{0.02} & \pmv{2.46}{0.04} & \pmv{\textbf{0.19}}{0.03} & \pmv{0.33}{0.08} \\
        & Rank & \pmv{\underline{2.50}}{0.56} & \pmv{\underline{2.50}}{1.30} & \pmv{8.00}{0.00} & \pmv{6.75}{0.22} & \pmv{5.50}{0.43} & \pmv{4.00}{0.50} & \pmv{9.00}{0.00} & \pmv{\textbf{2.25}}{0.22} & \pmv{4.50}{0.25} \\
        \noalign{\smallskip}\cline{2-11}\noalign{\smallskip}
        & Time & \pmv{0.33}{0.04} & \pmv{1.61}{0.05} & \pmv{\textbf{0.06}}{0.02} & \pmv{0.56}{0.00} & \pmv{0.17}{0.01} & \pmv{1.17}{0.02} & \pmv{2.97}{0.00} & \pmv{\underline{0.14}}{0.00} & \pmv{9.32}{0.35} \\
        & Time Rank & \pmv{4.00}{0.00} & \pmv{7.00}{0.00} & \pmv{\textbf{1.00}}{0.00} & \pmv{5.00}{0.00} & \pmv{3.00}{0.00} & \pmv{6.00}{0.00} & \pmv{8.00}{0.00} & \pmv{\underline{2.00}}{0.00} & \pmv{9.00}{0.00} \\
        \midrule

        \multirow{4}{*}{\textbf{Dim 2}}
        & MAE & \pmv{1.56}{0.83} & \pmv{\textbf{0.36}}{0.09} & \pmv{0.99}{0.10} & \pmv{0.98}{0.10} & \pmv{1.09}{0.09} & \pmv{0.82}{0.05} & \pmv{2.04}{0.32} & \pmv{\underline{0.58}}{0.03} & \pmv{0.60}{0.13} \\
        & Rank & \pmv{4.50}{1.48} & \pmv{\textbf{1.75}}{0.41} & \pmv{6.25}{0.41} & \pmv{6.00}{0.35} & \pmv{7.75}{0.22} & \pmv{4.50}{0.43} & \pmv{8.75}{0.22} & \pmv{\underline{2.50}}{0.25} & \pmv{3.00}{0.79} \\
        \noalign{\smallskip}\cline{2-11}\noalign{\smallskip}
        & Time & \pmv{1.00}{0.60} & \pmv{2.05}{0.14} & \pmv{\textbf{0.01}}{0.00} & \pmv{0.56}{0.00} & \pmv{\underline{0.20}}{0.01} & \pmv{1.15}{0.00} & \pmv{5.43}{0.06} & \pmv{0.53}{0.00} & \pmv{13.08}{0.48} \\
\bottomrule
    \end{tabular}%
    }
\end{table}

\begin{table}[t]
    \centering
  \caption{\textbf{Density estimation} performance via MAE ($\downarrow$) and rank, as well as runtime in seconds ($\downarrow$) and runtime rank for varying data dimensions.
    Average rank per metric (lower is better) is across methods over test datasets from the Copula prior.}
    \vspace{-0.1in}
    \label{tab:mae_dense_copula}
    \resizebox{\columnwidth}{!}{%
    \begin{tabular}{lc|cc|ccccccc}
        \toprule
        \textbf{Dimension} & \textbf{Metric} & \textbf{PRODiGI} & \textbf{PRODiGI-FT} & \textbf{DiscoFormer} & \textbf{DiscoFormer-FT}  &\textbf{NNKDE} & \textbf{NNKDE-FT} & \textbf{TabPFN} & \textbf{KDE} & \textbf{GMM} \\
        \midrule

        \multirow{4}{*}{\textbf{Dim 1}}
        & MAE & \pmv{\underline{0.05}}{0.02} & \pmv{\textbf{0.04}}{0.01} & \pmv{0.45}{0.05} & \pmv{0.34}{0.06} & \pmv{0.18}{0.02} & \pmv{0.19}{0.01} & \pmv{2.39}{0.02} & \pmv{0.20}{0.05} & \pmv{0.13}{0.03} \\
        & Rank & \pmv{\underline{2.12}}{0.27} & \pmv{\textbf{1.12}}{0.11} & \pmv{8.00}{0.00} & \pmv{7.00}{0.00} & \pmv{4.75}{0.22} & \pmv{5.00}{0.61} & \pmv{9.00}{0.00} & \pmv{4.75}{0.65} & \pmv{3.25}{0.41} \\
        \noalign{\smallskip}\cline{2-11}\noalign{\smallskip}
        & Time & \pmv{0.29}{0.00} & \pmv{1.79}{0.45} & \pmv{\textbf{0.06}}{0.04} & \pmv{0.73}{0.10} & \pmv{0.17}{0.00} & \pmv{1.15}{0.00} & \pmv{1.97}{0.84} & \pmv{\underline{0.15}}{0.00} & \pmv{8.54}{0.10} \\
        & Time Rank & \pmv{4.00}{0.00} & \pmv{7.75}{0.22} & \pmv{\textbf{1.50}}{0.43} & \pmv{5.00}{0.00} & \pmv{2.50}{0.25} & \pmv{6.75}{0.22} & \pmv{6.50}{0.43} & \pmv{\underline{2.00}}{0.35} & \pmv{9.00}{0.00} \\
        \midrule

        \multirow{4}{*}{\textbf{Dim 2}}
        & MAE & \pmv{0.42}{0.16} & \pmv{\textbf{0.28}}{0.11} & \pmv{0.52}{0.09} & \pmv{0.59}{0.08} & \pmv{0.70}{0.08} & \pmv{0.68}{0.02} & \pmv{1.53}{0.29} & \pmv{0.29}{0.02} & \pmv{\underline{0.29}}{0.02} \\
        & Rank & \pmv{4.00}{1.06} & \pmv{\textbf{1.50}}{0.43} & \pmv{4.75}{0.22} & \pmv{6.00}{0.00} & \pmv{7.50}{0.25} & \pmv{6.75}{0.82} & \pmv{9.00}{0.00} & \pmv{\underline{2.50}}{0.25} & \pmv{3.00}{0.61} \\
        \noalign{\smallskip}\cline{2-11}\noalign{\smallskip}
        & Time & \pmv{0.30}{0.00} & \pmv{2.91}{0.43} & \pmv{\textbf{0.01}}{0.00} & \pmv{0.55}{0.00} & \pmv{\underline{0.19}}{0.01} & \pmv{1.16}{0.00} & \pmv{6.33}{1.31} & \pmv{0.44}{0.00} & \pmv{11.62}{0.65} \\
\bottomrule
    \end{tabular}%
    }
\end{table}

\begin{table}[t]
    \centering
   \caption{\textbf{Density estimation} performance via Spearman correlation between the overall ranking by estimates and  ground-truth ($\uparrow$) and rank, as well as runtime in seconds ($\downarrow$) and runtime rank for varying data dimensions.
    Average rank per metric (lower is better) is across methods over test datasets from ALL prior families.}
    \vspace{-0.1in}
    \label{tab:corr_dense_all_priors}
    \resizebox{\columnwidth}{!}{%
    \begin{tabular}{lc|cc|ccccccc}
        \toprule
        \textbf{Dimension} & \textbf{Metric} & \textbf{PRODiGI} & \textbf{PRODiGI-FT} & \textbf{DiscoFormer} & \textbf{DiscoFormer-FT}  &\textbf{NNKDE} & \textbf{NNKDE-FT} & \textbf{TabPFN} & \textbf{KDE} & \textbf{GMM} \\
        \midrule

        \multirow{4}{*}{\textbf{Dim 1}}
        & Corr & \pmv{0.77}{0.09} & \pmv{\underline{0.88}}{0.08} & \pmv{0.38}{0.10} & \pmv{0.47}{0.12} & \pmv{0.84}{0.03} & \pmv{0.84}{0.05} & \pmv{0.80}{0.05} & \pmv{\textbf{0.90}}{0.04} & \pmv{0.85}{0.05} \\
        & Rank & \pmv{4.68}{0.85} & \pmv{\textbf{2.68}}{0.62} & \pmv{8.77}{0.10} & \pmv{7.14}{0.48} & \pmv{4.55}{0.65} & \pmv{4.82}{0.53} & \pmv{5.73}{0.32} & \pmv{\underline{2.73}}{0.32} & \pmv{3.91}{0.57} \\
        \noalign{\smallskip}\cline{2-11}\noalign{\smallskip}
        & Time & \pmv{0.31}{0.02} & \pmv{1.54}{0.18} & \pmv{\textbf{0.04}}{0.02} & \pmv{0.62}{0.04} & \pmv{0.17}{0.00} & \pmv{1.16}{0.01} & \pmv{3.24}{1.07} & \pmv{\underline{0.14}}{0.00} & \pmv{8.76}{0.19} \\
        & Time Rank & \pmv{4.00}{0.00} & \pmv{7.27}{0.19} & \pmv{\textbf{1.18}}{0.17} & \pmv{5.00}{0.00} & \pmv{2.91}{0.09} & \pmv{6.45}{0.15} & \pmv{7.36}{0.32} & \pmv{\underline{1.91}}{0.09} & \pmv{8.91}{0.09} \\
        \midrule

        \multirow{4}{*}{\textbf{Dim 2}}
        & Corr & \pmv{0.79}{0.06} & \pmv{\underline{0.93}}{0.03} & \pmv{0.78}{0.03} & \pmv{0.79}{0.03} & \pmv{0.88}{0.02} & \pmv{0.81}{0.02} & \pmv{0.76}{0.05} & \pmv{\textbf{0.93}}{0.01} & \pmv{0.90}{0.03} \\
        & Rank & \pmv{5.21}{0.93} & \pmv{\textbf{2.38}}{0.45} & \pmv{7.50}{0.38} & \pmv{6.50}{0.43} & \pmv{4.75}{0.39} & \pmv{5.75}{0.53} & \pmv{6.50}{0.72} & \pmv{\underline{2.83}}{0.31} & \pmv{3.58}{0.68} \\
        \noalign{\smallskip}\cline{2-11}\noalign{\smallskip}
        & Time & \pmv{0.53}{0.22} & \pmv{2.19}{0.22} & \pmv{\textbf{0.01}}{0.00} & \pmv{0.56}{0.00} & \pmv{\underline{0.19}}{0.01} & \pmv{1.16}{0.00} & \pmv{5.48}{0.54} & \pmv{0.46}{0.01} & \pmv{11.35}{0.54} \\
\bottomrule
    \end{tabular}%
    }
\end{table}

\begin{table}[t]
    \centering
 \caption{\textbf{Density estimation} performance via Spearman correlation between the overall ranking by estimates and  ground-truth ($\uparrow$) and rank, as well as runtime in seconds ($\downarrow$) and runtime rank for varying data dimensions.
    Average rank per metric (lower is better) is across methods over test datasets from the GMM prior.}
    \vspace{-0.1in}
    \label{tab:corr_dense_gmm}
    \resizebox{\columnwidth}{!}{%
    \begin{tabular}{lc|cc|ccccccc}
        \toprule
        \textbf{Dimension} & \textbf{Metric} & \textbf{PRODiGI} & \textbf{PRODiGI-FT} & \textbf{DiscoFormer} & \textbf{DiscoFormer-FT}  &\textbf{NNKDE} & \textbf{NNKDE-FT} & \textbf{TabPFN} & \textbf{KDE} & \textbf{GMM} \\
        \midrule

        \multirow{4}{*}{\textbf{Dim 1}}
        & Corr & \pmv{0.68}{0.04} & \pmv{0.91}{0.04} & \pmv{0.50}{0.11} & \pmv{0.64}{0.14} & \pmv{0.92}{0.03} & \pmv{0.90}{0.02} & \pmv{0.87}{0.04} & \pmv{\textbf{0.96}}{0.01} & \pmv{\underline{0.95}}{0.02} \\
        & Rank & \pmv{7.50}{0.25} & \pmv{3.25}{0.96} & \pmv{9.00}{0.00} & \pmv{7.00}{0.61} & \pmv{3.00}{0.79} & \pmv{5.00}{0.61} & \pmv{5.50}{0.56} & \pmv{\textbf{2.00}}{0.35} & \pmv{\underline{2.75}}{0.54} \\
        \noalign{\smallskip}\cline{2-11}\noalign{\smallskip}
        & Time & \pmv{0.30}{0.00} & \pmv{1.16}{0.02} & \pmv{\textbf{0.01}}{0.00} & \pmv{0.56}{0.00} & \pmv{0.18}{0.01} & \pmv{1.16}{0.00} & \pmv{5.20}{2.61} & \pmv{\underline{0.13}}{0.00} & \pmv{8.30}{0.05} \\
        & Time Rank & \pmv{4.00}{0.00} & \pmv{7.00}{0.35} & \pmv{\textbf{1.00}}{0.00} & \pmv{5.00}{0.00} & \pmv{3.00}{0.00} & \pmv{6.50}{0.25} & \pmv{7.75}{0.54} & \pmv{\underline{2.00}}{0.00} & \pmv{8.75}{0.22} \\
        \midrule

        \multirow{4}{*}{\textbf{Dim 2}}
        & Corr & \pmv{0.67}{0.14} & \pmv{0.92}{0.06} & \pmv{0.89}{0.05} & \pmv{0.86}{0.07} & \pmv{0.95}{0.02} & \pmv{0.82}{0.04} & \pmv{0.75}{0.07} & \pmv{\underline{0.98}}{0.00} & \pmv{\textbf{0.99}}{0.00} \\
        & Rank & \pmv{7.12}{1.62} & \pmv{3.12}{0.87} & \pmv{6.00}{0.35} & \pmv{5.75}{0.82} & \pmv{4.25}{0.41} & \pmv{6.25}{0.89} & \pmv{8.00}{0.35} & \pmv{\underline{3.00}}{0.50} & \pmv{\textbf{1.50}}{0.43} \\
        \noalign{\smallskip}\cline{2-11}\noalign{\smallskip}
        & Time & \pmv{0.29}{0.00} & \pmv{1.60}{0.05} & \pmv{\textbf{0.01}}{0.00} & \pmv{0.56}{0.00} & \pmv{\underline{0.17}}{0.01} & \pmv{1.16}{0.00} & \pmv{4.68}{0.78} & \pmv{0.43}{0.00} & \pmv{9.34}{0.39} \\
\bottomrule
    \end{tabular}%
    }
\end{table}
\begin{table}[t]
    \centering
\caption{\textbf{Density estimation} performance via Spearman correlation between the overall ranking by estimates and  ground-truth ($\uparrow$) and rank, as well as runtime in seconds ($\downarrow$) and runtime rank for varying data dimensions.
    Average rank per metric (lower is better) is across methods over test datasets from the SCM prior.}
    \vspace{-0.1in}
    \label{tab:corr_dense_scm}
    \resizebox{\columnwidth}{!}{%
    \begin{tabular}{lc|cc|ccccccc}
        \toprule
        \textbf{Dimension} & \textbf{Metric} & \textbf{PRODiGI} & \textbf{PRODiGI-FT} & \textbf{DiscoFormer} & \textbf{DiscoFormer-FT}  &\textbf{NNKDE} & \textbf{NNKDE-FT} & \textbf{TabPFN} & \textbf{KDE} & \textbf{GMM} \\
        \midrule

        \multirow{4}{*}{\textbf{Dim 1}}
        & Corr & \pmv{0.69}{0.20} & \pmv{0.76}{0.21} & \pmv{-0.00}{0.00} & \pmv{-0.00}{0.00} & \pmv{0.73}{0.04} & \pmv{\underline{0.76}}{0.11} & \pmv{0.68}{0.11} & \pmv{\textbf{0.82}}{0.08} & \pmv{0.67}{0.09} \\
        & Rank & \pmv{3.75}{1.34} & \pmv{\textbf{2.50}}{1.30} & \pmv{8.38}{0.11} & \pmv{8.38}{0.11} & \pmv{4.50}{1.03} & \pmv{3.75}{0.54} & \pmv{5.50}{0.25} & \pmv{\underline{2.75}}{0.22} & \pmv{5.50}{0.56} \\
        \noalign{\smallskip}\cline{2-11}\noalign{\smallskip}
        & Time & \pmv{0.33}{0.04} & \pmv{1.61}{0.05} & \pmv{\textbf{0.06}}{0.02} & \pmv{0.56}{0.00} & \pmv{0.17}{0.01} & \pmv{1.17}{0.02} & \pmv{2.97}{0.00} & \pmv{\underline{0.14}}{0.00} & \pmv{9.32}{0.35} \\
        & Time Rank & \pmv{4.00}{0.00} & \pmv{7.00}{0.00} & \pmv{\textbf{1.00}}{0.00} & \pmv{5.00}{0.00} & \pmv{3.00}{0.00} & \pmv{6.00}{0.00} & \pmv{8.00}{0.00} & \pmv{\underline{2.00}}{0.00} & \pmv{9.00}{0.00} \\
        \midrule

        \multirow{4}{*}{\textbf{Dim 2}}
        & Corr & \pmv{0.84}{0.09} & \pmv{\textbf{0.96}}{0.02} & \pmv{0.74}{0.02} & \pmv{0.77}{0.02} & \pmv{0.81}{0.02} & \pmv{0.87}{0.01} & \pmv{0.82}{0.04} & \pmv{\underline{0.91}}{0.01} & \pmv{0.85}{0.05} \\
        & Rank & \pmv{4.25}{1.56} & \pmv{\textbf{1.50}}{0.25} & \pmv{8.50}{0.43} & \pmv{7.00}{0.50} & \pmv{6.25}{0.41} & \pmv{4.25}{0.41} & \pmv{5.25}{1.19} & \pmv{\underline{3.25}}{0.22} & \pmv{4.75}{1.29} \\
        \noalign{\smallskip}\cline{2-11}\noalign{\smallskip}
        & Time & \pmv{1.00}{0.60} & \pmv{2.05}{0.14} & \pmv{\textbf{0.01}}{0.00} & \pmv{0.56}{0.00} & \pmv{\underline{0.20}}{0.01} & \pmv{1.15}{0.00} & \pmv{5.43}{0.06} & \pmv{0.53}{0.00} & \pmv{13.08}{0.48} \\
\bottomrule
    \end{tabular}%
    }
\end{table}

\begin{table}[t]
    \centering
\caption{\textbf{Density estimation} performance via Spearman correlation between the overall ranking by estimates and  ground-truth ($\uparrow$) and rank, as well as runtime in seconds ($\downarrow$) and runtime rank for varying data dimensions.
    Average rank per metric (lower is better) is across methods over test datasets from the Copula prior.}
    \vspace{-0.1in}
    \label{tab:corr_dense_copula}
    \resizebox{\columnwidth}{!}{%
    \begin{tabular}{lc|cc|ccccccc}
        \toprule
        \textbf{Dimension} & \textbf{Metric} & \textbf{PRODiGI} & \textbf{PRODiGI-FT} & \textbf{DiscoFormer} & \textbf{DiscoFormer-FT}  &\textbf{NNKDE} & \textbf{NNKDE-FT} & \textbf{TabPFN} & \textbf{KDE} & \textbf{GMM} \\
        \midrule

        \multirow{4}{*}{\textbf{Dim 1}}
        & Corr & \pmv{\textbf{1.00}}{0.00} & \pmv{\textbf{1.00}}{0.00} & \pmv{0.73}{0.05} & \pmv{0.88}{0.06} & \pmv{0.88}{0.03} & \pmv{0.87}{0.05} & \pmv{0.88}{0.02} & \pmv{0.92}{0.04} & \pmv{0.96}{0.03} \\
        & Rank & \pmv{\textbf{2.17}}{0.54} & \pmv{\textbf{2.17}}{0.54} & \pmv{9.00}{0.00} & \pmv{5.67}{0.98} & \pmv{6.67}{0.27} & \pmv{6.00}{1.25} & \pmv{6.33}{0.72} & \pmv{3.67}{0.72} & \pmv{3.33}{1.19} \\
        \noalign{\smallskip}\cline{2-11}\noalign{\smallskip}
        & Time & \pmv{0.29}{0.00} & \pmv{1.95}{0.57} & \pmv{\textbf{0.07}}{0.05} & \pmv{0.78}{0.12} & \pmv{0.17}{0.00} & \pmv{1.15}{0.00} & \pmv{1.00}{0.04} & \pmv{\underline{0.14}}{0.00} & \pmv{8.64}{0.05} \\
        & Time Rank & \pmv{4.00}{0.00} & \pmv{8.00}{0.00} & \pmv{\textbf{1.67}}{0.54} & \pmv{5.00}{0.00} & \pmv{2.67}{0.27} & \pmv{7.00}{0.00} & \pmv{6.00}{0.00} & \pmv{\textbf{1.67}}{0.27} & \pmv{9.00}{0.00} \\
        \midrule

        \multirow{4}{*}{\textbf{Dim 2}}
        & Corr & \pmv{0.86}{0.08} & \pmv{\underline{0.90}}{0.07} & \pmv{0.72}{0.04} & \pmv{0.75}{0.05} & \pmv{0.89}{0.02} & \pmv{0.75}{0.03} & \pmv{0.71}{0.11} & \pmv{\textbf{0.92}}{0.02} & \pmv{0.86}{0.02} \\
        & Rank & \pmv{4.25}{1.14} & \pmv{\underline{2.50}}{0.83} & \pmv{8.00}{0.35} & \pmv{6.75}{0.74} & \pmv{3.75}{0.41} & \pmv{6.75}{0.82} & \pmv{6.25}{1.47} & \pmv{\textbf{2.25}}{0.65} & \pmv{4.50}{0.83} \\
        \noalign{\smallskip}\cline{2-11}\noalign{\smallskip}
        & Time & \pmv{0.30}{0.00} & \pmv{2.91}{0.43} & \pmv{\textbf{0.01}}{0.00} & \pmv{0.55}{0.00} & \pmv{\underline{0.19}}{0.01} & \pmv{1.16}{0.00} & \pmv{6.33}{1.31} & \pmv{0.44}{0.00} & \pmv{11.62}{0.65} \\
\bottomrule
    \end{tabular}%
    }
\end{table}

\begin{table}[t]
    \centering
    \caption{\textbf{Density estimation} performance via NLL ($\downarrow$) and rank, as well as runtime in seconds ($\downarrow$) and runtime rank for varying data dimensions.
    Average rank per metric (lower is better) is across methods over test datasets from ALL prior families. We do not report NLL's for TabPFN and Disco variants, as their predicted densities do not integrate to one.}
    \vspace{-0.1in}
    \label{tab:mae_dense_all_priors}
    \resizebox{\columnwidth}{!}{%
    \begin{tabular}{lc|cc|cccc}
        \toprule
        \textbf{Dimension} & \textbf{Metric} & \textbf{PRODiGI} & \textbf{PRODiGI-FT} & \textbf{NNKDE} & \textbf{NNKDE-FT} & \textbf{KDE} & \textbf{GMM} \\
        \midrule

        \multirow{4}{*}{\textbf{Dim 1}}
        & NLL & \pmv{0.26}{0.42} & \pmv{\textbf{0.11}}{0.46} & \pmv{0.28}{0.44} & \pmv{0.12}{0.46} & \pmv{\underline{0.12}}{0.46} & \pmv{0.13}{0.44} \\
        & Rank & \pmv{3.71}{0.54} & \pmv{\textbf{2.29}}{0.35} & \pmv{5.67}{0.14} & \pmv{3.75}{0.31} & \pmv{\underline{2.67}}{0.46} & \pmv{2.92}{0.32} \\
        \noalign{\smallskip}\cline{2-8}\noalign{\smallskip}
        & Time & \pmv{0.31}{0.01} & \pmv{1.52}{0.17} & \pmv{\underline{0.17}}{0.00} & \pmv{1.16}{0.01} & \pmv{\textbf{0.14}}{0.00} & \pmv{8.72}{0.18} \\
        & Time Rank & \pmv{3.00}{0.00} & \pmv{4.92}{0.08} & \pmv{\underline{1.92}}{0.08} & \pmv{4.08}{0.08} & \pmv{\textbf{1.08}}{0.08} & \pmv{6.00}{0.00} \\
        \midrule

        \multirow{4}{*}{\textbf{Dim 2}}
        & NLL & \pmv{-0.87}{0.87} & \pmv{\underline{-1.49}}{0.94} & \pmv{-0.88}{0.91} & \pmv{-1.11}{0.94} & \pmv{-1.44}{0.93} & \pmv{\textbf{-1.50}}{0.93} \\
        & Rank & \pmv{4.00}{0.55} & \pmv{\textbf{1.92}}{0.25} & \pmv{5.67}{0.14} & \pmv{4.50}{0.14} & \pmv{2.92}{0.25} & \pmv{\underline{2.00}}{0.26} \\
        \noalign{\smallskip}\cline{2-8}\noalign{\smallskip}
        & Time & \pmv{0.53}{0.22} & \pmv{2.19}{0.22} & \pmv{\textbf{0.19}}{0.01} & \pmv{1.16}{0.00} & \pmv{\underline{0.46}}{0.01} & \pmv{11.35}{0.54} \\
\bottomrule
    \end{tabular}%
    }
\end{table}

\begin{table}[t]
    \centering
   \caption{\textbf{Density estimation} performance via NLL ($\downarrow$) and rank, as well as runtime in seconds ($\downarrow$) and runtime rank for varying data dimensions.
    Average rank per metric (lower is better) is across methods over test datasets from the GMM prior.  We do not report NLL's for TabPFN and Disco variants, as their predicted densities do not integrate to one.}
    \vspace{-0.1in}
    \label{tab:mae_dense_gmm}
    
    \resizebox{\columnwidth}{!}{%
    \begin{tabular}{lc|cc|cccc}
        \toprule
        \textbf{Dimension} & \textbf{Metric} & \textbf{PRODiGI} & \textbf{PRODiGI-FT} & \textbf{NNKDE} & \textbf{NNKDE-FT} & \textbf{KDE} & \textbf{GMM} \\
        \midrule

        \multirow{4}{*}{\textbf{Dim 1}}
        & NLL & \pmv{0.83}{0.53} & \pmv{0.49}{0.64} & \pmv{0.60}{0.64} & \pmv{0.49}{0.63} & \pmv{\textbf{0.46}}{0.63} & \pmv{\underline{0.46}}{0.63} \\
        & Rank & \pmv{5.50}{0.25} & \pmv{2.75}{0.41} & \pmv{5.50}{0.25} & \pmv{3.75}{0.22} & \pmv{\textbf{1.50}}{0.43} & \pmv{\underline{2.00}}{0.35} \\
        \noalign{\smallskip}\cline{2-8}\noalign{\smallskip}
        & Time & \pmv{0.30}{0.00} & \pmv{1.16}{0.02} & \pmv{\underline{0.18}}{0.01} & \pmv{1.16}{0.00} & \pmv{\textbf{0.13}}{0.00} & \pmv{8.30}{0.05} \\
        & Time Rank & \pmv{3.00}{0.00} & \pmv{4.75}{0.22} & \pmv{\underline{2.00}}{0.00} & \pmv{4.25}{0.22} & \pmv{\textbf{1.00}}{0.00} & \pmv{6.00}{0.00} \\
        \midrule

        \multirow{4}{*}{\textbf{Dim 2}}
        & NLL & \pmv{2.10}{1.06} & \pmv{1.63}{1.18} & \pmv{2.10}{1.18} & \pmv{1.97}{1.20} & \pmv{\underline{1.57}}{1.20} & \pmv{\textbf{1.54}}{1.20} \\
        & Rank & \pmv{4.75}{1.08} & \pmv{\underline{2.25}}{0.41} & \pmv{5.25}{0.22} & \pmv{4.25}{0.22} & \pmv{2.75}{0.41} & \pmv{\textbf{1.75}}{0.41} \\
        \noalign{\smallskip}\cline{2-8}\noalign{\smallskip}
        & Time & \pmv{\underline{0.29}}{0.00} & \pmv{1.60}{0.05} & \pmv{\textbf{0.17}}{0.01} & \pmv{1.16}{0.00} & \pmv{0.43}{0.00} & \pmv{9.34}{0.39} \\
\bottomrule
    \end{tabular}%
    }
\end{table}

\begin{table}[t]
    \centering
  \caption{\textbf{Density estimation} performance via NLL ($\downarrow$) and rank, as well as runtime in seconds ($\downarrow$) and runtime rank for varying data dimensions.
    Average rank per metric (lower is better) is across methods over test datasets from the SCM prior. We do not report NLL's for TabPFN and Disco variants, as their predicted densities do not integrate to one.}
    \vspace{-0.1in}
    \label{tab:mae_dense_SCM}
    
    \resizebox{\columnwidth}{!}{%
    \begin{tabular}{lc|cc|cccc}
        \toprule
        \textbf{Dimension} & \textbf{Metric} & \textbf{PRODiGI} & \textbf{PRODiGI-FT} & \textbf{NNKDE} & \textbf{NNKDE-FT} & \textbf{KDE} & \textbf{GMM} \\
        \midrule

        \multirow{4}{*}{\textbf{Dim 1}}
        & NLL & \pmv{-0.29}{1.07} & \pmv{-0.39}{1.16} & \pmv{-0.11}{1.12} & \pmv{\underline{-0.39}}{1.18} & \pmv{\textbf{-0.41}}{1.17} & \pmv{-0.32}{1.11} \\
        & Rank & \pmv{3.50}{1.03} & \pmv{\underline{2.75}}{0.74} & \pmv{5.75}{0.22} & \pmv{3.25}{0.74} & \pmv{\textbf{2.00}}{0.35} & \pmv{3.75}{0.22} \\
        \noalign{\smallskip}\cline{2-8}\noalign{\smallskip}
        & Time & \pmv{0.33}{0.04} & \pmv{1.61}{0.05} & \pmv{\underline{0.17}}{0.01} & \pmv{1.17}{0.02} & \pmv{\textbf{0.14}}{0.00} & \pmv{9.32}{0.35} \\
        & Time Rank & \pmv{3.00}{0.00} & \pmv{5.00}{0.00} & \pmv{\underline{2.00}}{0.00} & \pmv{4.00}{0.00} & \pmv{\textbf{1.00}}{0.00} & \pmv{6.00}{0.00} \\
        \midrule

        \multirow{4}{*}{\textbf{Dim 2}}
        & NLL & \pmv{-2.75}{0.71} & \pmv{\textbf{-4.00}}{0.75} & \pmv{-3.17}{0.71} & \pmv{-3.59}{0.75} & \pmv{-3.82}{0.75} & \pmv{\underline{-3.90}}{0.73} \\
        & Rank & \pmv{4.00}{0.94} & \pmv{\textbf{1.75}}{0.41} & \pmv{5.75}{0.22} & \pmv{4.50}{0.25} & \pmv{2.75}{0.41} & \pmv{\underline{2.25}}{0.41} \\
        \noalign{\smallskip}\cline{2-8}\noalign{\smallskip}
        & Time & \pmv{1.00}{0.60} & \pmv{2.05}{0.14} & \pmv{\textbf{0.20}}{0.01} & \pmv{1.15}{0.00} & \pmv{\underline{0.53}}{0.00} & \pmv{13.08}{0.48} \\
\bottomrule
    \end{tabular}%
    }
\end{table}

\begin{table}[t]
    \centering
  \caption{\textbf{Density estimation} performance via NLL ($\downarrow$) and rank, as well as runtime in seconds ($\downarrow$) and runtime rank for varying data dimensions.
    Average rank per metric (lower is better) is across methods over test datasets from the Copula prior. We do not report NLL's for TabPFN and Disco variants, as their predicted densities do not integrate to one.}
    \vspace{-0.1in}
    \label{tab:mae_dense_copula}
    
    \resizebox{\columnwidth}{!}{%
    \begin{tabular}{lc|cc|cccc}
        \toprule
        \textbf{Dimension} & \textbf{Metric} & \textbf{PRODiGI} & \textbf{PRODiGI-FT} & \textbf{NNKDE} & \textbf{NNKDE-FT} & \textbf{KDE} & \textbf{GMM} \\
        \midrule

        \multirow{4}{*}{\textbf{Dim 1}}
        & NLL & \pmv{\underline{0.23}}{0.18} & \pmv{\textbf{0.22}}{0.18} & \pmv{0.36}{0.19} & \pmv{0.26}{0.18} & \pmv{0.31}{0.19} & \pmv{0.24}{0.18} \\
        & Rank & \pmv{\underline{2.12}}{0.27} & \pmv{\textbf{1.38}}{0.21} & \pmv{5.75}{0.22} & \pmv{4.25}{0.41} & \pmv{4.50}{0.56} & \pmv{3.00}{0.61} \\
        \noalign{\smallskip}\cline{2-8}\noalign{\smallskip}
        & Time & \pmv{0.29}{0.00} & \pmv{1.79}{0.45} & \pmv{\underline{0.17}}{0.00} & \pmv{1.15}{0.00} & \pmv{\textbf{0.15}}{0.00} & \pmv{8.54}{0.10} \\
        & Time Rank & \pmv{3.00}{0.00} & \pmv{5.00}{0.00} & \pmv{\underline{1.75}}{0.22} & \pmv{4.00}{0.00} & \pmv{\textbf{1.25}}{0.22} & \pmv{6.00}{0.00} \\
        \midrule

        \multirow{4}{*}{\textbf{Dim 2}}
        & NLL & \pmv{-1.96}{1.37} & \pmv{\underline{-2.09}}{1.38} & \pmv{-1.57}{1.38} & \pmv{-1.71}{1.41} & \pmv{-2.07}{1.40} & \pmv{\textbf{-2.13}}{1.40} \\
        & Rank & \pmv{3.25}{0.65} & \pmv{\textbf{1.75}}{0.41} & \pmv{6.00}{0.00} & \pmv{4.75}{0.22} & \pmv{3.25}{0.41} & \pmv{\underline{2.00}}{0.50} \\
        \noalign{\smallskip}\cline{2-8}\noalign{\smallskip}
        & Time & \pmv{\underline{0.30}}{0.00} & \pmv{2.91}{0.43} & \pmv{\textbf{0.19}}{0.01} & \pmv{1.16}{0.00} & \pmv{0.44}{0.00} & \pmv{11.62}{0.65} \\
\bottomrule
    \end{tabular}%
    }
\end{table}

\begin{table}[t]
    \centering
  \caption{\textbf{Score estimation}    performance via MAE ($\downarrow$) and rank, as well as runtime in seconds ($\downarrow$) and runtime rank for varying data dimensions.
    Average rank per metric (lower is better) is across methods over test datasets from ALL prior families.}
    \vspace{-0.1in}
    \label{tab:mae_score_all_priors}
    \resizebox{\columnwidth}{!}{%
    \begin{tabular}{lc|cc|ccccccc}
        \toprule
        \textbf{Dimension} & \textbf{Metric} & \textbf{PRODiGI} & \textbf{PRODiGI-FT} & \textbf{DiscoFormer} & \textbf{DiscoFormer-FT}  &\textbf{NNKDE} & \textbf{NNKDE-FT} & \textbf{TabPFN} & \textbf{KDE} & \textbf{GMM} \\
        \midrule

        \multirow{4}{*}{\textbf{Dim 1}}
        & MAE & \pmv{\textbf{158.49}}{140.91} & \pmv{828.06}{790.26} & \pmv{546.36}{519.90} & \pmv{542.86}{516.50} & \pmv{\underline{532.95}}{508.15} & \pmv{735.29}{686.78} & \pmv{595.26}{563.83} & \pmv{561.77}{535.06} & \pmv{541.06}{515.33} \\
        & Rank & \pmv{4.88}{0.87} & \pmv{\underline{3.38}}{0.74} & \pmv{4.83}{0.26} & \pmv{4.92}{0.49} & \pmv{\textbf{2.75}}{0.36} & \pmv{8.83}{0.11} & \pmv{7.08}{0.42} & \pmv{4.42}{0.58} & \pmv{3.92}{0.61} \\
        \noalign{\smallskip}\cline{2-11}\noalign{\smallskip}
        & Time & \pmv{0.31}{0.01} & \pmv{1.52}{0.17} & \pmv{\textbf{0.04}}{0.02} & \pmv{0.62}{0.04} & \pmv{0.17}{0.00} & \pmv{1.16}{0.01} & \pmv{1.85}{0.43} & \pmv{\underline{0.15}}{0.00} & \pmv{8.72}{0.18} \\
        & Time Rank & \pmv{4.00}{0.00} & \pmv{7.58}{0.18} & \pmv{\textbf{1.17}}{0.16} & \pmv{5.00}{0.00} & \pmv{2.75}{0.12} & \pmv{6.33}{0.14} & \pmv{7.08}{0.22} & \pmv{\underline{2.08}}{0.14} & \pmv{9.00}{0.00} \\
        \midrule

        \multirow{4}{*}{\textbf{Dim 2}}
        & MAE & \pmv{\underline{332.22}}{199.24} & \pmv{\textbf{325.47}}{257.55} & \pmv{379.61}{292.29} & \pmv{376.06}{290.29} & \pmv{373.40}{289.89} & \pmv{613.22}{350.71} & \pmv{440.37}{317.47} & \pmv{385.43}{294.12} & \pmv{384.11}{294.60} \\
        & Rank & \pmv{5.21}{0.86} & \pmv{\textbf{2.79}}{0.58} & \pmv{4.50}{0.25} & \pmv{4.67}{0.36} & \pmv{\underline{3.33}}{0.50} & \pmv{8.92}{0.08} & \pmv{7.42}{0.25} & \pmv{4.17}{0.44} & \pmv{4.00}{0.80} \\
\bottomrule
    \end{tabular}%
    }
\end{table}

\begin{table}[t]
    \centering
  \caption{\textbf{Score estimation} performance via MAE ($\downarrow$) and rank, as well as runtime in seconds ($\downarrow$) and runtime rank for varying data dimensions.
    Average rank per metric (lower is better) is across methods over test datasets from the GMM prior.}
   \label{tab:mae_score_gmm}
    \resizebox{\columnwidth}{!}{%
    \begin{tabular}{lc|cc|ccccccc}
        \toprule
        \textbf{Dimension} & \textbf{Metric} & \textbf{PRODiGI} & \textbf{PRODiGI-FT} & \textbf{DiscoFormer} & \textbf{DiscoFormer-FT}  &\textbf{NNKDE} & \textbf{NNKDE-FT} & \textbf{TabPFN} & \textbf{KDE} & \textbf{GMM} \\
        \midrule

        \multirow{4}{*}{\textbf{Dim 1}}
        & MAE & \pmv{9.05}{5.11} & \pmv{1.83}{0.46} & \pmv{4.65}{1.94} & \pmv{4.98}{2.03} & \pmv{2.46}{0.77} & \pmv{22.28}{9.12} & \pmv{6.89}{2.76} & \pmv{\underline{1.82}}{0.57} & \pmv{\textbf{1.69}}{0.50} \\
        & Rank & \pmv{7.50}{0.25} & \pmv{3.25}{0.74} & \pmv{5.25}{0.22} & \pmv{6.25}{0.22} & \pmv{3.00}{0.50} & \pmv{9.00}{0.00} & \pmv{6.50}{1.03} & \pmv{\textbf{1.75}}{0.41} & \pmv{\underline{2.50}}{0.56} \\
        \noalign{\smallskip}\cline{2-11}\noalign{\smallskip}
        & Time & \pmv{0.30}{0.00} & \pmv{1.16}{0.02} & \pmv{\textbf{0.01}}{0.00} & \pmv{0.56}{0.00} & \pmv{0.18}{0.01} & \pmv{1.16}{0.00} & \pmv{2.73}{1.14} & \pmv{\underline{0.14}}{0.00} & \pmv{8.30}{0.05} \\
        & Time Rank & \pmv{4.00}{0.00} & \pmv{7.00}{0.35} & \pmv{\textbf{1.00}}{0.00} & \pmv{5.00}{0.00} & \pmv{3.00}{0.00} & \pmv{6.50}{0.25} & \pmv{7.50}{0.43} & \pmv{\underline{2.00}}{0.00} & \pmv{9.00}{0.00} \\
        \midrule

        \multirow{4}{*}{\textbf{Dim 2}}
        & MAE & \pmv{5.94}{2.95} & \pmv{\underline{2.52}}{1.29} & \pmv{3.42}{1.68} & \pmv{3.99}{1.91} & \pmv{3.03}{1.40} & \pmv{18.47}{7.87} & \pmv{4.72}{1.95} & \pmv{2.54}{1.24} & \pmv{\textbf{1.51}}{0.76} \\
        & Rank & \pmv{6.12}{1.35} & \pmv{\underline{2.88}}{0.91} & \pmv{4.00}{0.35} & \pmv{5.75}{0.54} & \pmv{4.75}{0.74} & \pmv{9.00}{0.00} & \pmv{7.00}{0.61} & \pmv{4.00}{0.79} & \pmv{\textbf{1.50}}{0.43} \\
        \noalign{\smallskip}\cline{2-11}\noalign{\smallskip}
        & Time & \pmv{0.29}{0.00} & \pmv{1.60}{0.05} & \pmv{\textbf{0.01}}{0.00} & \pmv{0.56}{0.00} & \pmv{\underline{0.17}}{0.01} & \pmv{1.16}{0.00} & \pmv{6.79}{0.15} & \pmv{0.43}{0.00} & \pmv{9.35}{0.39} \\
\bottomrule
    \end{tabular}%
    }
\end{table}

\begin{table}[t]
    \centering
  \caption{\textbf{Score estimation} performance via MAE ($\downarrow$) and rank, as well as runtime in seconds ($\downarrow$) and runtime rank for varying data dimensions.
    Average rank per metric (lower is better) is across methods over test datasets from the SCM prior.}
    \vspace{-0.1in}
    \label{tab:mae_score_copula}
    \resizebox{\columnwidth}{!}{%
    \begin{tabular}{lc|cc|ccccccc}
        \toprule
        \textbf{Dimension} & \textbf{Metric} & \textbf{PRODiGI} & \textbf{PRODiGI-FT} & \textbf{DiscoFormer} & \textbf{DiscoFormer-FT}  &\textbf{NNKDE} & \textbf{NNKDE-FT} & \textbf{TabPFN} & \textbf{KDE} & \textbf{GMM} \\
        \midrule

        \multirow{4}{*}{\textbf{Dim 1}}
        & MAE & \pmv{\textbf{466.16}}{378.38} & \pmv{2482.19}{2143.48} & \pmv{1632.80}{1410.70} & \pmv{1621.91}{1401.54} & \pmv{\underline{1595.12}}{1378.73} & \pmv{2168.64}{1864.00} & \pmv{1772.82}{1530.07} & \pmv{1681.65}{1451.32} & \pmv{1619.40}{1397.86} \\
        & Rank & \pmv{5.00}{1.77} & \pmv{5.50}{1.48} & \pmv{4.00}{0.35} & \pmv{\underline{3.00}}{0.35} & \pmv{\textbf{1.50}}{0.25} & \pmv{8.50}{0.25} & \pmv{6.75}{0.41} & \pmv{6.00}{0.35} & \pmv{4.75}{0.74} \\
        \noalign{\smallskip}\cline{2-11}\noalign{\smallskip}
        & Time & \pmv{0.33}{0.04} & \pmv{1.61}{0.05} & \pmv{\textbf{0.06}}{0.02} & \pmv{0.56}{0.00} & \pmv{0.17}{0.01} & \pmv{1.17}{0.02} & \pmv{1.43}{0.00} & \pmv{\underline{0.15}}{0.00} & \pmv{9.32}{0.35} \\
        & Time Rank & \pmv{4.00}{0.00} & \pmv{8.00}{0.00} & \pmv{\textbf{1.00}}{0.00} & \pmv{5.00}{0.00} & \pmv{2.75}{0.22} & \pmv{6.00}{0.00} & \pmv{7.00}{0.00} & \pmv{\underline{2.25}}{0.22} & \pmv{9.00}{0.00} \\
        \midrule

        \multirow{4}{*}{\textbf{Dim 2}}
        & MAE & \pmv{\underline{957.98}}{458.45} & \pmv{\textbf{944.20}}{673.20} & \pmv{1081.92}{763.65} & \pmv{1072.32}{758.89} & \pmv{1069.18}{757.75} & \pmv{1634.66}{839.10} & \pmv{1213.76}{823.60} & \pmv{1104.61}{764.17} & \pmv{1095.06}{768.53} \\
        & Rank & \pmv{5.25}{1.67} & \pmv{\underline{2.50}}{1.03} & \pmv{4.75}{0.22} & \pmv{3.75}{0.22} & \pmv{\textbf{2.25}}{0.22} & \pmv{8.75}{0.22} & \pmv{7.50}{0.25} & \pmv{5.25}{0.65} & \pmv{5.00}{1.22} \\
\bottomrule
    \end{tabular}%
    }
\end{table}

\begin{table}[t]
    \centering
  \caption{\textbf{Score estimation} performance via MAE ($\downarrow$) and rank, as well as runtime in seconds ($\downarrow$) and runtime rank for varying data dimensions.
    Average rank per metric (lower is better) is across methods over test datasets from the Copula prior.}
    \vspace{-0.1in}
    \label{tab:mae_score_copula}
    \resizebox{\columnwidth}{!}{%
    \begin{tabular}{lc|cc|ccccccc}
        \toprule
        \textbf{Dimension} & \textbf{Metric} & \textbf{PRODiGI} & \textbf{PRODiGI-FT} & \textbf{DiscoFormer} & \textbf{DiscoFormer-FT}  &\textbf{NNKDE} & \textbf{NNKDE-FT} & \textbf{TabPFN} & \textbf{KDE} & \textbf{GMM} \\
        \midrule

        \multirow{4}{*}{\textbf{Dim 1}}
        & MAE & \pmv{\underline{0.26}}{0.16} & \pmv{\textbf{0.17}}{0.09} & \pmv{1.64}{0.18} & \pmv{1.68}{0.19} & \pmv{1.26}{0.19} & \pmv{14.94}{2.95} & \pmv{6.08}{0.64} & \pmv{1.85}{0.27} & \pmv{2.10}{1.00} \\
        & Rank & \pmv{\underline{2.12}}{0.27} & \pmv{\textbf{1.38}}{0.21} & \pmv{5.25}{0.41} & \pmv{5.50}{0.75} & \pmv{3.75}{0.41} & \pmv{9.00}{0.00} & \pmv{8.00}{0.00} & \pmv{5.50}{0.25} & \pmv{4.50}{1.30} \\
        \noalign{\smallskip}\cline{2-11}\noalign{\smallskip}
        & Time & \pmv{0.29}{0.00} & \pmv{1.79}{0.45} & \pmv{\textbf{0.06}}{0.04} & \pmv{0.73}{0.10} & \pmv{0.17}{0.00} & \pmv{1.15}{0.00} & \pmv{1.38}{0.22} & \pmv{\underline{0.15}}{0.00} & \pmv{8.54}{0.10} \\
        & Time Rank & \pmv{4.00}{0.00} & \pmv{7.75}{0.22} & \pmv{\textbf{1.50}}{0.43} & \pmv{5.00}{0.00} & \pmv{2.50}{0.25} & \pmv{6.50}{0.25} & \pmv{6.75}{0.41} & \pmv{\underline{2.00}}{0.35} & \pmv{9.00}{0.00} \\
        \midrule

        \multirow{4}{*}{\textbf{Dim 2}}
        & MAE & \pmv{\underline{32.74}}{12.00} & \pmv{\textbf{29.69}}{10.54} & \pmv{53.48}{21.24} & \pmv{51.88}{19.80} & \pmv{47.98}{19.20} & \pmv{186.55}{89.77} & \pmv{102.64}{56.84} & \pmv{49.14}{19.74} & \pmv{55.76}{23.44} \\
        & Rank & \pmv{4.25}{1.24} & \pmv{\textbf{3.00}}{1.06} & \pmv{4.75}{0.54} & \pmv{4.50}{0.56} & \pmv{\textbf{3.00}}{0.94} & \pmv{9.00}{0.00} & \pmv{7.75}{0.22} & \pmv{3.25}{0.41} & \pmv{5.50}{1.30} \\
        \noalign{\smallskip}\cline{2-11}\noalign{\smallskip}
        & Time & \pmv{0.30}{0.00} & \pmv{2.92}{0.43} & \pmv{\textbf{0.01}}{0.00} & \pmv{0.55}{0.00} & \pmv{\underline{0.19}}{0.01} & \pmv{1.16}{0.00} & \pmv{7.68}{0.43} & \pmv{0.45}{0.00} & \pmv{11.62}{0.65} \\
\bottomrule
    \end{tabular}%
    }
\end{table}

\begin{table}[t]
    \centering
 \caption{\textbf{Score estimation} performance via cosine similarity between score vector estimates and  ground-truth ($\uparrow$) and rank, as well as runtime in seconds ($\downarrow$) and runtime rank for varying data dimensions.
    Average rank per metric (lower is better) is across methods over test datasets from ALL prior families.}
    \vspace{-0.1in}
    \label{tab:cos_score_all_priors}
    \resizebox{\columnwidth}{!}{%
    \begin{tabular}{lc|cc|ccccccc}
        \toprule
        \textbf{Dimension} & \textbf{Metric} & \textbf{PRODiGI} & \textbf{PRODiGI-FT} & \textbf{DiscoFormer} & \textbf{DiscoFormer-FT}  &\textbf{NNKDE} & \textbf{NNKDE-FT} & \textbf{TabPFN} & \textbf{KDE} & \textbf{GMM} \\
        \midrule

        \multirow{4}{*}{\textbf{Dim 1}}
        & Cosine & \pmv{\underline{0.58}}{0.11} & \pmv{\textbf{0.69}}{0.11} & \pmv{0.31}{0.10} & \pmv{0.33}{0.10} & \pmv{0.49}{0.08} & \pmv{0.13}{0.02} & \pmv{0.30}{0.06} & \pmv{0.50}{0.09} & \pmv{0.48}{0.11} \\
        & Rank & \pmv{4.21}{0.78} & \pmv{\textbf{2.92}}{0.57} & \pmv{6.92}{0.41} & \pmv{6.42}{0.64} & \pmv{3.92}{0.48} & \pmv{7.38}{0.52} & \pmv{5.79}{0.37} & \pmv{\underline{3.42}}{0.43} & \pmv{4.04}{0.63} \\
        \noalign{\smallskip}\cline{2-11}\noalign{\smallskip}
        & Time & \pmv{0.31}{0.01} & \pmv{1.52}{0.17} & \pmv{\textbf{0.04}}{0.02} & \pmv{0.62}{0.04} & \pmv{0.17}{0.00} & \pmv{1.16}{0.01} & \pmv{1.85}{0.43} & \pmv{\underline{0.15}}{0.00} & \pmv{8.72}{0.18} \\
        & Time Rank & \pmv{4.00}{0.00} & \pmv{7.58}{0.18} & \pmv{\textbf{1.17}}{0.16} & \pmv{5.00}{0.00} & \pmv{2.75}{0.12} & \pmv{6.33}{0.14} & \pmv{7.08}{0.22} & \pmv{\underline{2.08}}{0.14} & \pmv{9.00}{0.00} \\
        \midrule

        \multirow{4}{*}{\textbf{Dim 2}}
        & Cosine & \pmv{\underline{0.66}}{0.09} & \pmv{\textbf{0.82}}{0.06} & \pmv{0.50}{0.10} & \pmv{0.49}{0.10} & \pmv{0.59}{0.09} & \pmv{0.16}{0.03} & \pmv{0.41}{0.07} & \pmv{0.61}{0.08} & \pmv{0.58}{0.11} \\
        & Rank & \pmv{4.88}{0.93} & \pmv{\textbf{3.04}}{0.66} & \pmv{5.96}{0.42} & \pmv{5.79}{0.49} & \pmv{3.58}{0.46} & \pmv{8.54}{0.18} & \pmv{6.00}{0.42} & \pmv{\underline{3.50}}{0.28} & \pmv{3.71}{0.83} \\
        \noalign{\smallskip}\cline{2-11}\noalign{\smallskip}
        & Time & \pmv{0.53}{0.22} & \pmv{2.19}{0.22} & \pmv{\textbf{0.01}}{0.00} & \pmv{0.56}{0.00} & \pmv{\underline{0.19}}{0.01} & \pmv{1.16}{0.00} & \pmv{7.21}{0.19} & \pmv{0.48}{0.01} & \pmv{11.35}{0.53} \\
\bottomrule
    \end{tabular}%
    }
\end{table}

\begin{table}[t]
    \centering
 \caption{\textbf{Score estimation} performance via cosine similarity between score vector estimates and  ground-truth ($\uparrow$) and rank, as well as runtime in seconds ($\downarrow$) and runtime rank for varying data dimensions.
    Average rank per metric (lower is better) is across methods over test datasets from the GMM prior.}
    \vspace{-0.1in}
    \label{tab:cos_score_gmm}
    \resizebox{\columnwidth}{!}{%
    \begin{tabular}{lc|cc|ccccccc}
        \toprule
        \textbf{Dimension} & \textbf{Metric} & \textbf{PRODiGI} & \textbf{PRODiGI-FT} & \textbf{DiscoFormer} & \textbf{DiscoFormer-FT}  &\textbf{NNKDE} & \textbf{NNKDE-FT} & \textbf{TabPFN} & \textbf{KDE} & \textbf{GMM} \\
        \midrule

        \multirow{4}{*}{\textbf{Dim 1}}
        & Cosine & \pmv{0.35}{0.08} & \pmv{0.62}{0.15} & \pmv{0.45}{0.10} & \pmv{0.45}{0.10} & \pmv{0.70}{0.10} & \pmv{0.15}{0.02} & \pmv{0.51}{0.09} & \pmv{\textbf{0.76}}{0.06} & \pmv{\underline{0.73}}{0.08} \\
        & Rank & \pmv{6.88}{0.87} & \pmv{3.25}{1.14} & \pmv{6.25}{0.41} & \pmv{6.75}{0.65} & \pmv{3.00}{0.00} & \pmv{8.88}{0.11} & \pmv{5.50}{0.56} & \pmv{\textbf{2.00}}{0.61} & \pmv{\underline{2.50}}{0.43} \\
        \noalign{\smallskip}\cline{2-11}\noalign{\smallskip}
        & Time & \pmv{0.30}{0.00} & \pmv{1.16}{0.02} & \pmv{\textbf{0.01}}{0.00} & \pmv{0.56}{0.00} & \pmv{0.18}{0.01} & \pmv{1.16}{0.00} & \pmv{2.73}{1.14} & \pmv{\underline{0.14}}{0.00} & \pmv{8.30}{0.05} \\
        & Time Rank & \pmv{4.00}{0.00} & \pmv{7.00}{0.35} & \pmv{\textbf{1.00}}{0.00} & \pmv{5.00}{0.00} & \pmv{3.00}{0.00} & \pmv{6.50}{0.25} & \pmv{7.50}{0.43} & \pmv{\underline{2.00}}{0.00} & \pmv{9.00}{0.00} \\
        \midrule

        \multirow{4}{*}{\textbf{Dim 2}}
        & Cosine & \pmv{0.50}{0.16} & \pmv{0.87}{0.09} & \pmv{0.86}{0.05} & \pmv{0.81}{0.07} & \pmv{0.87}{0.02} & \pmv{0.26}{0.02} & \pmv{0.61}{0.07} & \pmv{\underline{0.90}}{0.01} & \pmv{\textbf{0.97}}{0.01} \\
        & Rank & \pmv{6.88}{1.56} & \pmv{\underline{3.38}}{1.08} & \pmv{4.50}{0.25} & \pmv{5.25}{0.65} & \pmv{4.50}{1.03} & \pmv{8.50}{0.25} & \pmv{6.50}{0.75} & \pmv{4.00}{0.61} & \pmv{\textbf{1.50}}{0.43} \\
        \noalign{\smallskip}\cline{2-11}\noalign{\smallskip}
        & Time & \pmv{0.29}{0.00} & \pmv{1.60}{0.05} & \pmv{\textbf{0.01}}{0.00} & \pmv{0.56}{0.00} & \pmv{\underline{0.17}}{0.01} & \pmv{1.16}{0.00} & \pmv{6.79}{0.15} & \pmv{0.43}{0.00} & \pmv{9.35}{0.39} \\
\bottomrule
    \end{tabular}%
    }
\end{table}
\begin{table}[t]
    \centering
 \caption{\textbf{Score estimation} performance via cosine similarity between score vector estimates and  ground-truth ($\uparrow$) and rank, as well as runtime in seconds ($\downarrow$) and runtime rank for varying data dimensions.
    Average rank per metric (lower is better) is across methods over test datasets from the SCM prior.}
    \vspace{-0.1in}
    \label{tab:cos_score_copula}
    \resizebox{\columnwidth}{!}{%
    \begin{tabular}{lc|cc|ccccccc}
        \toprule
        \textbf{Dimension} & \textbf{Metric} & \textbf{PRODiGI} & \textbf{PRODiGI-FT} & \textbf{DiscoFormer} & \textbf{DiscoFormer-FT}  &\textbf{NNKDE} & \textbf{NNKDE-FT} & \textbf{TabPFN} & \textbf{KDE} & \textbf{GMM} \\
        \midrule

        \multirow{4}{*}{\textbf{Dim 1}}
        & Cosine & \pmv{\underline{0.68}}{0.20} & \pmv{\textbf{0.73}}{0.20} & \pmv{-0.00}{0.00} & \pmv{-0.00}{0.00} & \pmv{0.29}{0.05} & \pmv{0.09}{0.02} & \pmv{0.12}{0.04} & \pmv{0.20}{0.05} & \pmv{0.13}{0.06} \\
        & Rank & \pmv{\underline{2.75}}{1.24} & \pmv{\textbf{2.50}}{1.03} & \pmv{8.25}{0.12} & \pmv{8.25}{0.12} & \pmv{3.00}{0.61} & \pmv{5.50}{0.56} & \pmv{5.12}{0.37} & \pmv{3.50}{0.25} & \pmv{6.12}{1.04} \\
        \noalign{\smallskip}\cline{2-11}\noalign{\smallskip}
        & Time & \pmv{0.33}{0.04} & \pmv{1.61}{0.05} & \pmv{\textbf{0.06}}{0.02} & \pmv{0.56}{0.00} & \pmv{0.17}{0.01} & \pmv{1.17}{0.02} & \pmv{1.43}{0.00} & \pmv{\underline{0.15}}{0.00} & \pmv{9.32}{0.35} \\
        & Time Rank & \pmv{4.00}{0.00} & \pmv{8.00}{0.00} & \pmv{\textbf{1.00}}{0.00} & \pmv{5.00}{0.00} & \pmv{2.75}{0.22} & \pmv{6.00}{0.00} & \pmv{7.00}{0.00} & \pmv{\underline{2.25}}{0.22} & \pmv{9.00}{0.00} \\
        \midrule

        \multirow{4}{*}{\textbf{Dim 2}}
        & Cosine & \pmv{\underline{0.78}}{0.13} & \pmv{\textbf{0.86}}{0.07} & \pmv{0.17}{0.14} & \pmv{0.22}{0.15} & \pmv{0.28}{0.14} & \pmv{0.04}{0.03} & \pmv{0.16}{0.09} & \pmv{0.33}{0.11} & \pmv{0.22}{0.16} \\
        & Rank & \pmv{\underline{3.25}}{1.39} & \pmv{\textbf{2.50}}{1.03} & \pmv{7.12}{0.67} & \pmv{5.38}{0.99} & \pmv{\underline{3.25}}{0.41} & \pmv{8.12}{0.37} & \pmv{6.25}{0.22} & \pmv{3.50}{0.25} & \pmv{5.62}{1.49} \\
        \noalign{\smallskip}\cline{2-11}\noalign{\smallskip}
        & Time & \pmv{1.00}{0.60} & \pmv{2.05}{0.14} & \pmv{\textbf{0.01}}{0.00} & \pmv{0.56}{0.00} & \pmv{\underline{0.20}}{0.01} & \pmv{1.15}{0.00} & \pmv{7.16}{0.06} & \pmv{0.55}{0.00} & \pmv{13.08}{0.48} \\
\bottomrule
    \end{tabular}%
    }
\end{table}

\begin{table}[t]
    \centering
 \caption{\textbf{Score estimation} performance via cosine similarity between score vector estimates and  ground-truth ($\uparrow$) and rank, as well as runtime in seconds ($\downarrow$) and runtime rank for varying data dimensions.
    Average rank per metric (lower is better) is across methods over test datasets from the Copula prior.}
    \vspace{-0.1in}
    \label{tab:cos_score_copula}
    \resizebox{\columnwidth}{!}{%
    \begin{tabular}{lc|cc|ccccccc}
        \toprule
        \textbf{Dimension} & \textbf{Metric} & \textbf{PRODiGI} & \textbf{PRODiGI-FT} & \textbf{DiscoFormer} & \textbf{DiscoFormer-FT}  &\textbf{NNKDE} & \textbf{NNKDE-FT} & \textbf{TabPFN} & \textbf{KDE} & \textbf{GMM} \\
        \midrule

        \multirow{4}{*}{\textbf{Dim 1}}
        & Cosine & \pmv{\textbf{0.73}}{0.21} & \pmv{\textbf{0.73}}{0.21} & \pmv{0.47}{0.20} & \pmv{0.54}{0.21} & \pmv{0.48}{0.15} & \pmv{0.14}{0.04} & \pmv{0.27}{0.08} & \pmv{0.53}{0.17} & \pmv{0.57}{0.21} \\
        & Rank & \pmv{\textbf{3.00}}{0.68} & \pmv{\textbf{3.00}}{0.68} & \pmv{6.25}{0.82} & \pmv{4.25}{1.08} & \pmv{5.75}{0.65} & \pmv{7.75}{0.82} & \pmv{6.75}{0.65} & \pmv{4.75}{0.54} & \pmv{3.50}{0.75} \\
        \noalign{\smallskip}\cline{2-11}\noalign{\smallskip}
        & Time & \pmv{0.29}{0.00} & \pmv{1.79}{0.45} & \pmv{\textbf{0.06}}{0.04} & \pmv{0.73}{0.10} & \pmv{0.17}{0.00} & \pmv{1.15}{0.00} & \pmv{1.38}{0.22} & \pmv{\underline{0.15}}{0.00} & \pmv{8.54}{0.10} \\
        & Time Rank & \pmv{4.00}{0.00} & \pmv{7.75}{0.22} & \pmv{\textbf{1.50}}{0.43} & \pmv{5.00}{0.00} & \pmv{2.50}{0.25} & \pmv{6.50}{0.25} & \pmv{6.75}{0.41} & \pmv{\underline{2.00}}{0.35} & \pmv{9.00}{0.00} \\
        \midrule

        \multirow{4}{*}{\textbf{Dim 2}}
        & Cosine & \pmv{\underline{0.69}}{0.13} & \pmv{\textbf{0.73}}{0.12} & \pmv{0.46}{0.10} & \pmv{0.45}{0.11} & \pmv{0.62}{0.07} & \pmv{0.18}{0.03} & \pmv{0.47}{0.06} & \pmv{0.62}{0.07} & \pmv{0.57}{0.09} \\
        & Rank & \pmv{4.50}{1.30} & \pmv{3.25}{1.24} & \pmv{6.25}{0.41} & \pmv{6.75}{0.65} & \pmv{\textbf{3.00}}{0.61} & \pmv{9.00}{0.00} & \pmv{5.25}{0.89} & \pmv{\textbf{3.00}}{0.35} & \pmv{4.00}{1.27} \\
        \noalign{\smallskip}\cline{2-11}\noalign{\smallskip}
        & Time & \pmv{0.30}{0.00} & \pmv{2.92}{0.43} & \pmv{\textbf{0.01}}{0.00} & \pmv{0.55}{0.00} & \pmv{\underline{0.19}}{0.01} & \pmv{1.16}{0.00} & \pmv{7.68}{0.43} & \pmv{0.45}{0.00} & \pmv{11.62}{0.65} \\
\bottomrule
    \end{tabular}%
    }
\end{table}

\begin{table}[t]
    \centering
 \caption{\textbf{Distribution match} performance via MMD ($\downarrow$) and rank, as well as runtime in seconds ($\downarrow$) and runtime rank for varying data dimensions.
    Average rank per metric (lower is better) across methods over real-world test datasets.}
    \vspace{-0.1in}
    \label{tab:real_mmd}
    \resizebox{\columnwidth}{!}{%
\begin{tabular}{c@{\hspace{2pt}}l|@{\hspace{2pt}}*{2}{c}|@{\hspace{2pt}}*{7}{c}}
        \toprule
        & \textbf{Metric}  &\textbf{PRODiGI} & \textbf{-FT} & \textbf{DiscoFormer} & \textbf{-FT} & \textbf{NNKDE} & \textbf{-FT} & \textbf{TabPFN} & \textbf{KDE} & \textbf{GMM} \\
        \midrule
\multirow{3}{*}{\rotatebox[origin=c]{90}{\textbf{Generation}}}
                & MMD ($\downarrow$) {\tiny$\times 10^3$} & \pmv{8.61}{1.65} & \pmv{\underline{1.80}}{0.44} & \pmv{35.64}{6.98} & \pmv{78.29}{20.89} & \pmv{15.72}{2.38} & \pmv{7.90}{1.36} & -\tnote{c} & \pmv{3.19}{0.74} & \pmv{\textbf{1.33}}{0.56} \\

                & Rank ($\downarrow$) & \pmv{4.22}{0.17} & \pmv{\underline{2.53}}{0.12} & \pmv{7.02}{0.10} & \pmv{7.03}{0.11} & \pmv{5.63}{0.14} & \pmv{4.66}{0.15} & -\tnote{c} & \pmv{3.16}{0.16} & \pmv{\textbf{1.75}}{0.06} \\

        \noalign{\smallskip}\cline{2-11}\noalign{\smallskip}
                & \% $p$$<$$0.05$ ($\downarrow$) & 64.6\% & \underline{34.4\%} & 100\% & 100\% & 90.6\% & 75.0\% & -\tnote{c} & 49.0\% & \textbf{16.7\%} \\

        \midrule
        \midrule
\multirow{6}{*}{\rotatebox[origin=c]{90}{\textbf{Density}}}
                & MAE ($\downarrow$) & \pmv{19.55}{2.92} & \pmv{7.58}{0.94} & \pmv{8.09}{0.90} & \pmv{\underline{6.68}}{0.80} & \pmv{8.44}{0.96} & \pmv{8.31}{0.97} & -\tnote{c} & \pmv{8.84}{1.17} & \pmv{\textbf{5.78}}{0.79} \\

                & Rank ($\downarrow$) & \pmv{7.03}{0.14} & \pmv{\underline{3.99}}{0.23} & \pmv{5.08}{0.20} & \pmv{4.25}{0.24} & \pmv{4.74}{0.14} & \pmv{4.50}{0.13} & -\tnote{c} & \pmv{4.46}{0.27} & \pmv{\textbf{1.95}}{0.12} \\

        \noalign{\smallskip}\cline{2-11}\noalign{\smallskip}
                & Spear. ($\uparrow$) & \pmv{0.55}{0.03} & \pmv{0.70}{0.03} & \pmv{0.69}{0.02} & \pmv{0.72}{0.02} & \pmv{\underline{0.81}}{0.01} & \pmv{0.79}{0.01} & -\tnote{c} & \pmv{0.81}{0.01} & \pmv{\textbf{0.83}}{0.01} \\
\bottomrule
    \end{tabular}%
    }
\end{table}

\begin{table}[t]
    \centering
    \caption{\textbf{Rejection rate of two-sample tests for distribution match} w.r.t. MMD  over test datasets over real-world test datasets..
    Tests under generated data that successfully match the input data distribution fail to reject the null (i.e., $p>0.05$) that the two samples come from the same distribution.}
    \vspace{-0.1in}
    \label{tab:pval_mmd_all_priors}
    \resizebox{\columnwidth}{!}{%
    %
%

    }
\end{table}

\begin{table}[t]
    \centering
    \caption{\textbf{Density estimation} performance via MAE ($\downarrow$) and rank, as well as runtime in seconds ($\downarrow$) and runtime rank for varying data dimensions.
    Average rank per metric (lower is better) across methods over real-world test datasets.}
    \vspace{-0.1in}
    \label{tab:mae_dense_all_priors}
    \resizebox{\columnwidth}{!}{%
    %
%
    }
\end{table}

\begin{table}[t]
    \centering
   \caption{\textbf{Density estimation} performance via Spearman correlation between the overall ranking by estimates and  ground-truth ($\uparrow$) and rank, as well as runtime in seconds ($\downarrow$) and runtime rank for varying data dimensions.
    Average rank per metric (lower is better) across methods over real-world test datasets.}
    \vspace{-0.1in}
    \label{tab:corr_dense_all_priors}
    \resizebox{\columnwidth}{!}{%
    %
%
    }
\end{table}

\begin{table}[t]
    \centering
    \caption{\textbf{Density estimation} performance via NLL ($\downarrow$) and rank, as well as runtime in seconds ($\downarrow$) and runtime rank for varying data dimensions.
    Average rank per metric (lower is better) across methods over real-world test datasets. We do not report NLL's for TabPFN and Disco variants, as their predicted densities do not integrate to one.}
    \vspace{-0.1in}
    \label{tab:mae_dense_all_priors}
    \resizebox{\columnwidth}{!}{%
    %
%
    }
\end{table}

\begin{table}[t]
    \centering
  \caption{\textbf{Score estimation}    performance via MAE ($\downarrow$) and rank, as well as runtime in seconds ($\downarrow$) and runtime rank for varying data dimensions.
    Average rank per metric (lower is better) across methods over real-world test datasets.}
    \vspace{-0.1in}
    \label{tab:mae_score_all_priors}
    \resizebox{\columnwidth}{!}{%
    %
%
    }
\end{table}

\begin{table}[t]
    \centering
 \caption{\textbf{Score estimation} performance via cosine similarity between score vector estimates and  ground-truth ($\uparrow$) and rank, as well as runtime in seconds ($\downarrow$) and runtime rank for varying data dimensions.
    Average rank per metric (lower is better) across methods over real-world test datasets.}
    \vspace{-0.1in}
    \label{tab:cos_score_all_priors}
    \resizebox{\columnwidth}{!}{%
    %
%
    }
\end{table}

\end{document}